\documentclass[a4paper,fleqn]{cas-sc}

\usepackage[numbers]{natbib}

\usepackage{booktabs}
\usepackage{mathtools}
\usepackage{multirow}
\usepackage{soul}
\usepackage{xcolor}
\usepackage{changepage}
\usepackage{threeparttable}
\usepackage{caption}
\usepackage{subcaption}
\usepackage{pifont}
\usepackage{float}
\usepackage{microtype}

\usepackage[figuresright]{rotating}

\usepackage[linesnumbered,ruled,vlined]{algorithm2e}
\usepackage{amssymb}
\usepackage{algpseudocode}

\begin{document}
\let\WriteBookmarks\relax
\def\floatpagepagefraction{1}
\def\textpagefraction{.001}

\shorttitle{CLAMP}

\shortauthors{W. Weng et~al.}

\title [mode = title]{Cross-Domain Continual Learning via CLAMP}                      



%
\author[1]{Weiwei Weng}
\ead{weiwei002@e.ntu.edu.sg}
\fnmark[1]
\cormark[1]

\affiliation[1]{organization={SCSE, Nanyang Technological University},
    addressline={Nanyang Avenue}, 
    city={Singapore},
    country={Singapore}}

\author[2]{Mahardhika Pratama}
\ead{dhika.pratama@unisa.edu.au}
\fnmark[1]
\cormark[1]

\author[1]{Jie Zhang}

\affiliation[2]{organization={STEM, University of South Australia},
    addressline={Mawson Lakes Boulevard}, 
    city={Adelaide},
    country={Australia}}

\author[1]{Chen Chen}
\author[3]{Edward Yapp Kien Yie}
\author[4]{Ramasamy Savitha}

\affiliation[3]{organization={SIMTech, A*Star},
    addressline={Fusionopolis Way}, 
    city={Singapore},
    country={Singapore}}

\affiliation[4]{organization={I2R, A*Star},
    addressline={Fusionopolis Way}, 
    city={Singapore},
    country={Singapore}}
    
\cortext[cor1]{Corresponding author.}



\begin{abstract}
Artificial neural networks, celebrated for their human-like cognitive learning abilities, often encounter the well-known catastrophic forgetting (CF) problem, where the neural networks lose the proficiency in previously acquired knowledge. Despite numerous efforts to mitigate CF, it remains the significant challenge particularly in complex changing environments. This challenge is even more pronounced in cross-domain adaptation following the continual learning (CL) setting, which is a more challenging and realistic scenario that is under-explored. To this end, this article proposes a cross-domain CL approach making possible to deploy a single model in such environments without additional labelling costs. Our approach, namely continual learning approach for many processes (CLAMP), integrates a class-aware adversarial domain adaptation strategy to align a source domain and a target domain. An assessor-guided learning process is put forward to navigate the learning process of a base model assigning a set of weights to every sample controlling the influence of every sample and the interactions of each loss function in such a way to balance the stability and plasticity dilemma thus preventing the CF problem. The first assessor focuses on the negative transfer problem rejecting irrelevant samples of the source domain while the second assessor prevents noisy pseudo labels of the target domain. Both assessors are trained in the meta-learning approach using random transformation techniques and similar samples of the source domain.
Theoretical analysis and extensive numerical validations demonstrate that CLAMP significantly outperforms established baseline algorithms across all experiments by at least $10\%$ margin.


\end{abstract}


\begin{highlights}
\item This paper presents a scarcely addressed problem, cross-domain continual learning where a model needs to handle the catastrophic forgetting problem and the domain shifts problem simultaneously. 
\item This paper proposes a new algorithm, continual learning approach for many processes (CLAMP), to cope with the cross-domain continual learning problem. CLAMP is built upon the concept of an assessor-guided learning process under the meta-learning approach coupled with a class-aware adversarial domain adaptation. 
\item Extensive numerical validations and theoretical studies are performed to guarantee the advantage of CLAMP where it outperforms prior arts with significant margins.  
\end{highlights}

\begin{keywords}
Continual Learning \sep Domain Adaptation \sep Lifelong Learning \sep Incremental Learning \sep Cross Domain Adaptation \sep Unsupervised Domain Adaptation
\end{keywords}

\maketitle

\section{Introduction}
The underlying goal of continual learning (CL) is to design a learning algorithm handing never-ending environments efficiently. Unlike conventional learning algorithms dealing with a single task, the CL model is supposed to cope with a sequence of different tasks $T_1,T_2,T_3,...,T_K$ where $K$ denotes the number of tasks which might be unknown prior to process runs \textcolor{black}{Chen et al.} \cite{chen2018lifelong}. Note that the continual learning problem distinguishes itself from the multi-task learning problem because the learning tasks are received sequentially and not available at once. Each task possesses non-stationary characteristics, i.e., different data distributions, different class labels, or combinations between the two types. The key challenge of continual learning lies in the fast adaptation to new environments while preventing \textcolor{black}{the catastrophic forgetting (CF) problem} with the absence of data samples of old tasks $T_{k-1}$, i.e., previously relevant parameters are overwritten when learning new tasks.  

Numerous works have been devoted in the past few years where the main focus is to combat the issue of CF paving a way for knowledge accumulation across streaming tasks \textcolor{black}{Parisi et al.} \cite{Parisi2018ContinualLL}. The regularization-based approaches \textcolor{black}{Kirpatrick et al.} \cite{Kirkpatrick2016OvercomingCF} addresses the catastrophic forgetting problem where the regularization term protects important parameters of old tasks from changing. Although this approach is computationally efficient and easy to implement, it does not scale well for large-scale problems. \textcolor{black}{Mao et al.} \cite{Mao2021ContinualLV} aims to address this problem by considering mutual and shareable information across tasks while \textcolor{black}{Cha et al.} \cite{Cha2020CPRCR} proposes a projection concept actualizing a wide local optimum region. Nonetheless, the regularization-based approach is over-dependent on the task IDs for inference process. The structure-based approach adds new network components to embrace new tasks while isolating previous network parameters \textcolor{black}{Rusu et al.}\cite{Rusu2016ProgressiveNN}. Some approaches propose network evolution criteria for growing/pruning whereas other works put forward search-based approaches to find near-optimal network structures to accommodate new tasks \textcolor{black}{Pratama et al.}\cite{Pratama2021UnsupervisedCL}, \textcolor{black}{Yoon et al.}\cite{Yoon2017LifelongLW}. The first approach is fast to compute but does not assure optimal solutions whereas the second approach is computationally prohibitive. As with the regularization-based approach, the structure-based approach rely on the task IDs imposing extra domain knowledge. The memory-based approach \textcolor{black}{LopezPaz et al.} \cite{lopez2017gradient} is applicable without the task IDs where the experience replay mechanism is carried out using old samples stored in the memory. However, it requires hundreds of samples per class to be stored in the memory. This mechanism is tackled in \textcolor{black}{Shin et al.} \cite{Shin2017ContinualLW} using the pseudo-rehearsal approach via generative models such as generative adversarial network (GAN) \textcolor{black}{Goodfellow et al.} \cite{Goodfellow2014GenerativeAN} or variational auto-encoder (VAE) \textcolor{black}{Kingma et al.} \cite{Kingma2013AutoEncodingVB}. The pseudo-rehearsal mechanism is inferior compared to the conventional approaches because synthetic samples of generative models have lower qualities than original samples. 

The continual learning problem still requires in-depth study because the vast majority of existing works are limited to only a single domain and the cross-domain CL problem remains a relatively uncharted territory. This problem differs from multistream classification problems in \textcolor{black}{Chandra et al.} \cite{Chandra2016AnAF} because they still considers a single task. To the best of our knowledge, only \textcolor{black}{Lao et al.} \cite{Lao2021ATC} has explored the cross-domain continual learning problem. Our approach differs from this work where a meta-learning strategy is proposed to induce two assessors guiding the continual learning processes. In addition, the class-aware adversarial domain adaptation is put forward for domain alignments to address the domain shift problems.

Continual learning approach for many processes (CLAMP) is proposed as a solution of the cross-domain continual learning problems where there exist a fully labelled source domain and an unlabelled target domain. The underlying goal is to craft a predictive model of the target domain without any labels using discriminative information of the source domain. The source domain and the target domain are different but related where they share the same labelling function. The key difference with the multistream classification problem in \textcolor{black}{Chandra et al.} \cite{Chandra2016AnAF} lies in the cross-domain nature handling not only a single time-varying task but also a sequence of different tasks across different domains. Our problem also differs from \textcolor{black}{Lin et al.} \cite{Lin2022PrototypeGuidedCA} because the source domain is incremental in nature and does not carry complete class information. This problem extends the conventional continual learning problem where a continual learner not only handles a non-stationary process without the CF problem but also addresses a domain shift between source domain and target domain. 

CLAMP is designed to cope with the cross-domain CL problems where a class-aware adversarial domain adaptation approach \textcolor{black}{Ganin et al. and Li et al.} \cite{Ganin2015DomainAdversarialTO,Li2023MetaReweightedRF} is integrated to produce a domain-invariant network. The domain alignment strategy makes use of a min-max game between a feature extractor and a domain classifier. The feature extractor is trained in an adversarial manner to generate indistinguishable features of source domain and target domain while the domain classifier predicts sample's origins. Our approach goes one step ahead of the vanilla adversarial domain adaptation performing the domain alignment only and ignoring the class alignment of the target domain and the source domain. The class alignment is achieved by performing a self-training of the target domain where a pseudo-labelling process is committed to highly confident samples of the target domain. As with \textcolor{black}{Li et al. and Zheng et al.} \cite{Li2023MetaReweightedRF,Zheng2020DeepML}, the meta-learning-based regularization approach is applied to combat the noisy pseudo label problem. The key difference of our approach lies in dual meta-training loops also weighting every sample of the source domain to overcome the negative transfer issue in addition to the target domain to address the noisy pseudo-label problem. Our approach also relies on sequence-aware assessors producing a set of weights for every sample rather than a single weight. This approach not only controls the sample's influences addressing the issues of negative transfers and noisy pseudo labels but also the interactions of the three loss functions to achieve a proper tradeoff between the stability and the plasticity, a key issue of continual learning.

\textcolor{black}{This paper at least conveys five contributions: 
\begin{itemize}
    \item we propose a highly challenging but rarely discussed problem, namely the cross-domain continual learning problem. This problem features two different but related continual learning problems drawn from two different domains while sharing the same label space such that the CF problem and the domain shift problem must be tackled simultaneously. Source domain is fully labelled but the target domain suffers from the absence of any labels. The goal is to craft a continual learning model that performs well for both domains benefiting from label information of the source domain to the target domain. Such problem differs from Chandra et al. \cite{Chandra2016AnAF} because both source and target domains characterizes the continual learning problem and Lin et al. \cite{Lin2022PrototypeGuidedCA} because the source domain also features the continual learning process. Our problem is similar to those of De Carvalho \cite{VinciusdeCarvalho2024TowardsCC} and Lao et al \cite{Lao2021ATC} but we offer different approaches here.  \item CLAMP is proposed to resolve the cross-domain continual learning problem. This approach is inspired by the concept of meta weighted adversarial domain adaptation Li et al. \cite{Li2023MetaReweightedRF} where we expand this approach for the continual learning problem involving a multi-objective loss function. Our approach not only includes a meta-training loop in the target domain to reject noisy pseudo label but also in the source domain to avoid the issue of negative transfer. That is, a source-domain sample can be down-weighted if it poses a risk of negative transfer. Pseudo labels can be rejected by assigning low weights to minimize their effects;
    \item The concept of assessor-guided learning is put forward to improve the learning process where dual assessors are put forward to address the issue of negative transfer and noisy pseudo label as well as to attain proper trade off between the plasticity and the stability. This strategy innovates \textcolor{black}{Zheng et al and Masum et al} \cite{Zheng2020DeepML,Masum2023AssessorGuidedLF} where the assessor is not yet applied in such configurations and limited to a static setting. Note that Masum et al. \cite{Masum2023AssessorGuidedLF} applies the concept of assessor-guided learning to the ordinary continual learning problems while Zheng et al. \cite{Zheng2020DeepML} develops such concept only to a single-task metric learning problem;
    \item The theoretical analysis analyzing the generalization bound of the target domain is provided;
    \item the source codes of CLAMP are made publicly available in \url{https://github.com/wengweng001/CLAMP_torch.git} for reproducibility and convenient further study.
\end{itemize}}
\textcolor{black}{The remainder of this paper is structured as follows: Section 2 discusses the related works; Section 3 outlines the problem formulation; Section 4 explains the methodology of our paper; Section 5 is devoted to the theoretical analysis; Section 6 describes the experiments, discussions and limitations; Some concluding remarks are drawn in Section 7.}


\section{Related Works}
\textcolor{black}{This section discusses the related works which comprise three sub-sections: continual learning, meta-learning and multi-stream mining.}
\subsection{Continual Learning}
\noindent\textbf{Regularization-based approach} 
\cite{Kirkpatrick2016OvercomingCF,aljundi2018memory,Li2016LearningWF,Schwarz2018ProgressC,Paik2019OvercomingCF,Mao2021ContinualLV,Cha2020CPRCR} addresses the catastrophic forgetting problem in the continual learning domain with the application of additional regularization term protecting important parameters of old tasks from changing. Nevertheless, the regularization-based approach requires the task IDs and the task boundaries to be known hindering its feasibility in practise. \textbf{Structure-based approach} \cite{Rusu2016ProgressiveNN,Yoon2017LifelongLW,Li2019LearnTG,Xu2021AdaptivePC,Pratama2021UnsupervisedCL,Ashfahani2021UnsupervisedCL,Rakaraddi2022ReinforcedCL} expands network structures to deal with new tasks leaving old network parameters unchanged. As with regularization-based approach, this approach is sensitive to the task IDs. \textbf{Memory-based Approach} \cite{Rebuffi2016iCaRLIC,lopez2017gradient,Chaudhry2019OnTE,Chaudhry2018EfficientLL,Chaudhry2019UsingHT,Shin2017ContinualLW,VinciusdeCarvalho2022ClassIncrementalLV,Dam2022ScalableAO} takes different strategies where it stores a subset of old samples in the episodic memory. These samples are interleaved with new samples when learning a new task to reduce the CF problem. \textcolor{black}{Wang et al. \cite{Wang2021LearningTP,Wang2022DualPromptCP} recently offers a rehearsal-free approach using the concept of prompts. That is, learnable prompts are inserted to every task while fixing the backbone ViT network. These works have been successful without the use of any memories in the challenging class-incremental learning problems.} All these works still considers a single domain CL problem where we aim to extend here into cross-domain CL problems where the goal is to generalize over an unlabelled target domain from a fully labelled source domain. To the best of our knowledge, this problem is only addressed in \textcolor{black}{Lao et al.} \cite{Lao2021ATC} using the pseudo-rehearsal mechanism to address the catastrophic forgetting problem. Our approach differs from this work where the assessor-guided learning strategy and the class-aware adversarial domain adaptation approach are put forward. This problem also goes one step ahead of the multistream mining topic \textcolor{black}{Chandra et al.} \cite{Chandra2016AnAF} yet considering a single non-stationary task and of the class-incremental unsupervised domain adaptation \textcolor{black}{Lin et al.} \cite{Lin2022PrototypeGuidedCA} possessing all classes in the source domain. \textcolor{black}{Recent work by De Carvalho et al. \cite{VinciusdeCarvalho2024TowardsCC} addresses the same problem, the cross-domain continual learning problem. However, our approach is significantly different from this work where the assessor-guided learning technique containing the dual meta-training loop is proposed. Our numerical study also finds that our approach, CLAMP, outperforms this work.} 

\subsection{Meta-learning}
Meta-learning \textcolor{black}{Finn et al. and Schmidhuber et al.} \cite{Finn2017ModelAgnosticMF,Schmidhuber1987EvolutionaryPI} constitutes a learning-to-learn technique where the goal is to improve the learning performance of another learning algorithm. Such approach has been explored for CL \cite{Javed2019MetaLearningRF,Gupta2020LaMAMLLM,Dam2022ScalableAO} where \textcolor{black}{Javed et al.} \cite{Javed2019MetaLearningRF} introduces two networks, representation network and prediction network, trained in the meta-learning manner \textcolor{black}{and Gupta et al. \cite{Gupta2020LaMAMLLM} extends the concept of model agnostic meta-learning (MAML) to continual learning problems and puts forward the concept of learnable per variable learning rates}  while \textcolor{black}{Pham et al.} \cite{Pham2021ContextualTN} and \textcolor{black}{Dam et al.} \cite{Dam2022ScalableAO} adopt this concept for online task-incremental learning problems. We go one step ahead of these works where the meta-learning strategy is applied for cross-domain continual learning problem. That is, it aims to generalize well for a sequence of unlabelled target learning tasks provided with a sequence of labelled source learning tasks. The notion of assessor-guided learning is put forward in \textcolor{black}{Zheng et al.} \cite{Zheng2020DeepML} for a single-task metric learning. We innovate over this concept where it is applied in the continual multi-task environments. In addition, it not only performs a soft sample selection mechanism but also undertakes selections of learning strategies, i.e., which losses to be favoured in the multi-objective optimization problem. In \textcolor{black}{Masum et al.} \cite{Masum2023AssessorGuidedLF}, the notion of assessor-guided learning is adopted to cope with the continual learning problems. Our work distinguishes itself where such idea is extended for the cross-domain continual learning problems. Such problem requires the catastrophic forgetting problem and the domain shift problem to be overcome simultaneously.
\subsection{Multistream Mining}
Multistream mining is seen as an extension of the unsupervised domain adaptation problem where each domain features a streaming process rather than a fixed process \textcolor{black}{Chandra et al.} \cite{MSC}. As with the unsupervised domain adaptation problem, it aims to develop a stream-invariant model to handle an unlabelled target stream by transferring relevant knowledge of a source stream in which true class labels are available. The underlying challenge lies in the presence of asynchronous drift because the source stream and the target stream independently runs. A pioneering study in this area is proposed in \textcolor{black}{Chandra et al.} \cite{MSC} where it integrates the kernel mean matching (KMM) as a domain adaptation method and a drift detection approach to detect concept drifts in each stream. FUSION is proposed in \textcolor{black}{Haque et al.} \cite{FUSION} using KLIEP as the domain adaptation method and a density ratio to signal the asynchronous drifts to reduce computational burden of MSC. MSCRDR is put forward in \textcolor{black}{Dong et al.} \cite{MSCRDR} where the notion of Pearson's divergence is incorporated to address the domain shift problem. A deep learning solution is offered in \textcolor{black}{Pratama et al.} \cite{ATL}, where the multistream classification problem is handled using the encoder-decoder structure having shared parameters as well as the KL divergence approach.

The multistream classification problem is expanded to consider the presence of multi-source stream in \textcolor{black}{Du et al.} \cite{MELANIE}. This work is extended in \textcolor{black}{Du et al.} \cite{MARLINE}. Recently, the same problem is considered in \textcolor{black}{Xie et al.} \cite{AOMSDA} where the CMD-based regularization approach is proposed. The aforementioned works rely on a single domain assumption where the source stream and the target stream are drawn from the same feature space but different distributions. The problem of cross-domain multistream classification is attacked in \textcolor{black}{Tao et al.} \cite{COMC} and \textcolor{black}{Carvalho et al.} \cite{ACDC} where each stream features a unique feature space but shares the same target attributes. In \textcolor{black}{Tao et al.} \cite{COMC}, the empirical maximum mean discrepancy method is proposed whereas the adversarial domain adaptation technique is utilized in \textcolor{black}{Carvalho et al.} \cite{ACDC}. \textcolor{black}{Weng et al.} \cite{Weng2022AutonomousCD} considers the problem of label scarcity where no labels are provided in the source stream and target stream for model updates. This paper distinguishes itself from these works because the source domain and the target domain present a continual learning problem, i.e., a model is exposed with a sequence of different tasks in both the source domain and the target domain. In other words, we go beyond a single non-stationary task where a model not only handles the domain shift problem and the asynchronous drift problem but also addresses the task shift problem and the catastrophic forgetting problem. To the best of our knowledge, this problem is only addressed in \textcolor{black}{Lao et al.} \cite{Lao2021ATC} using the pseudo-rehearsal mechanism to address the catastrophic forgetting problem. Our approach differs from this work where the assessor-guided learning strategy and the class-aware adversarial domain adaptation approach are put forward.

\section{Problem Formulation}
The cross-domain continual learning problem is defined here as a learning problem to handle two different but related continual learning problems: $\mathcal{T}_1^S,\mathcal{T}_2^S,...,\mathcal{T}_K^S$ a source domain and $\mathcal{T}_1^{T},\mathcal{T}_2^{T},...,\mathcal{T}_K^T$ a target domain $k\in\{1,...,K\}$ where $K$ stands for the number of tasks. Labelled samples are only provided to the source domain $\mathcal{T}_k^{S}=\{(x_i^{S},y_i^{S})\}_{i=1}^{N_k^S}$ leaving the target domain completely unlabelled $\mathcal{T}_k^{T}=\{(x_i^{T})\}_{i=1}^{N_k^T}$. The source domain and the target domain are different but related because they possess different feature space $x_k^{S}\in\mathcal{X}_k^S,x_k^{T}\in\mathcal{X}_k^T,\mathcal{X}_k^S\neq\mathcal{X}_k^T$ but share the same target variables $y_k^{S},y_k^{T}\in\mathcal{Y}_k, y_i=[l_1,l_2,...,l_m]$ where $\mathcal{X}_k^{S}\times\mathcal{Y}_{k}\in\mathcal{D}_k^{S}$ and $\mathcal{X}_k^{T}\times\mathcal{Y}_{k}\in\mathcal{D}_k^{T}$. $m$ is the number of target classes. The source domain and the target domain run independently and follow different distributions $P(x_S)\neq P(x_T)$ such that $\mathcal{D}_{k}^{S}\neq\mathcal{D}_{k}^{T}$. $N_k^S,N_k^T$ denote the size of the $k-th$ task of the source domain and the target domain respectively where $N_k^S\neq N_k^T$.  

\begin{figure}[h]
\centering
  \centering
  \includegraphics[width=0.8\linewidth]{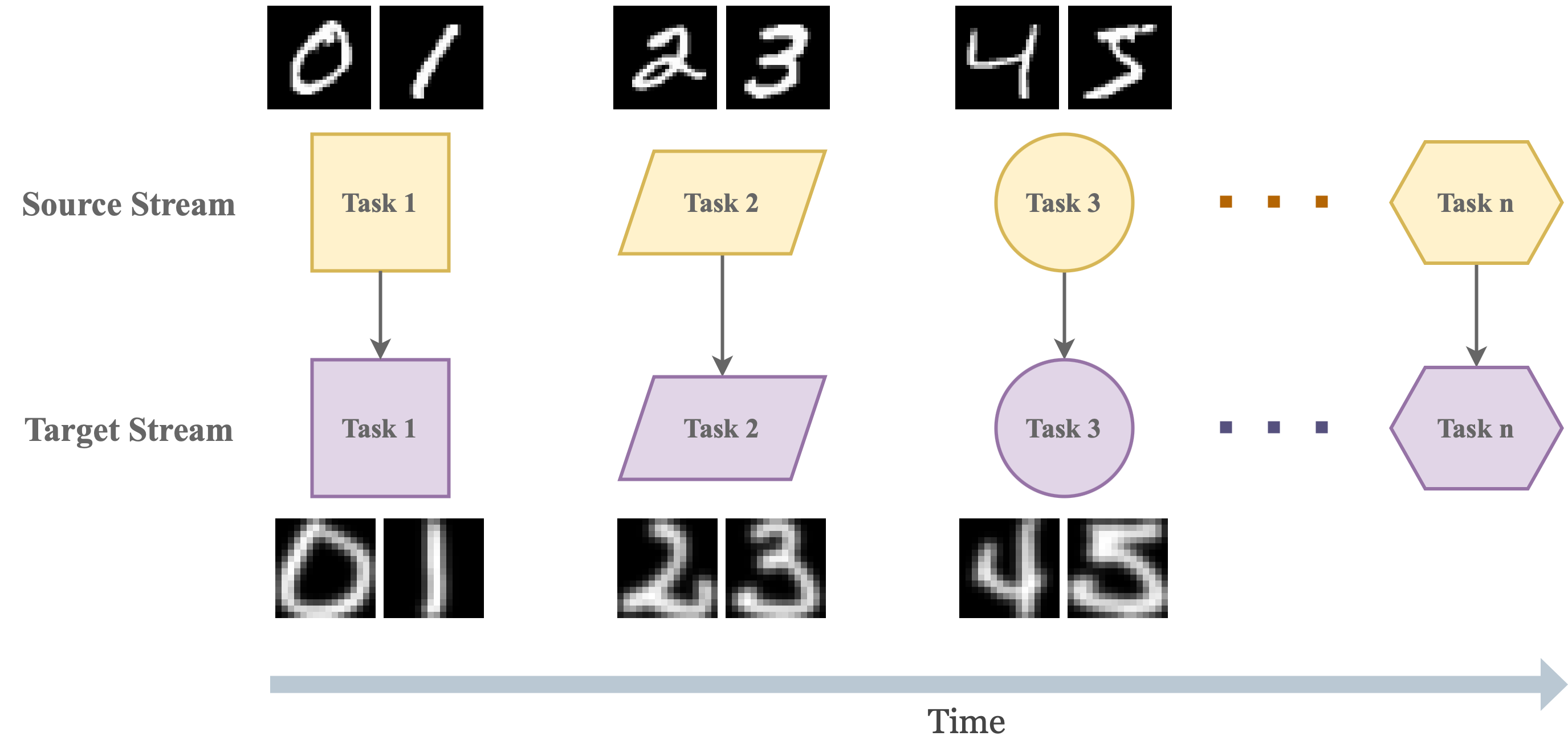}
  \caption{Cross-Domain Continual Learning Problems}
  \label{fig:scenario}
\end{figure}

Every task of the source domain and the target domain features non-stationary characteristics: domain-incremental learning, task-incremental learning and class-incremental learning \textcolor{black}{Van De Ven et al.} \cite{vandeVen2019ThreeSF}. The domain-incremental learning problem presents changing data distributions or concept drifts  of each task $P(x,y)_k^{S}\neq P(x,y)_{k+1}^{S}, P(x,y)_k^{T}\neq P(x,y)_{k+1}^{T}$. The task-incremental learning problem and the class-incremental learning problem feature different label sets of each task. Suppose that $L_k^{S,T}$ denotes a label set of the $k-th$ task of the source domain and the target domain, i.e, both domains carry the same label set, $L_k^{S,T}\cap L_{k'}^{S,T}=\emptyset, k,k'\in\{1,...,K\}$, i.e, every task carry unique label sets. The key difference of the task-incremental learning problem and the class-incremental learning problem lies in the presence of a task IDs for the task-incremental learning problem making possible to deploy different classifiers with a shared feature extractor. Since the class-incremental learning problem is more challenging than the task-incremental learning problem, we only focus on the class-incremental learning here. A model is only offered by data samples of the current task $(x^{S},y^{S})_k\backsim\mathcal{D}_k^{S}$ and $(x^{T})_k\backsim\mathcal{D}_k^{T}$ whereas old samples are discarded for the sake of memory and computational efficiencies. This strategy leads to the catastrophic forgetting problem. This problem is visualized in Fig. \ref{fig:scenario}.

The goal of this problem is to develop a predictive model $g_{\phi}(f_{\theta}(.))$ where $g_{\phi}(.)$ is a classifier parameterized by $\phi$ and $f_{\theta}(.)$ is a feature extractor parameterized by $\theta$ to perform well on all tasks of the unlabelled target domain by transferring knowledge of the labelled source domain. This is achieved by learning the source domain and the difference of the source domain and target domain $\frac{1}{K}\sum_{k=1}^{K}\mathcal{L}_k,\mathcal{L}_k\triangleq\mathbb{E}_{(x,y)_k\backsim\mathcal{D}_k^S}[l(g_{\phi}(f_{\theta}(x)),y)]+d(\mathcal{D}_k^S,\mathcal{D}_k^T)$. 

\section{Learning Procedure of CLAMP}
\begin{figure}[h]
    \centering
    \includegraphics[scale=0.35]{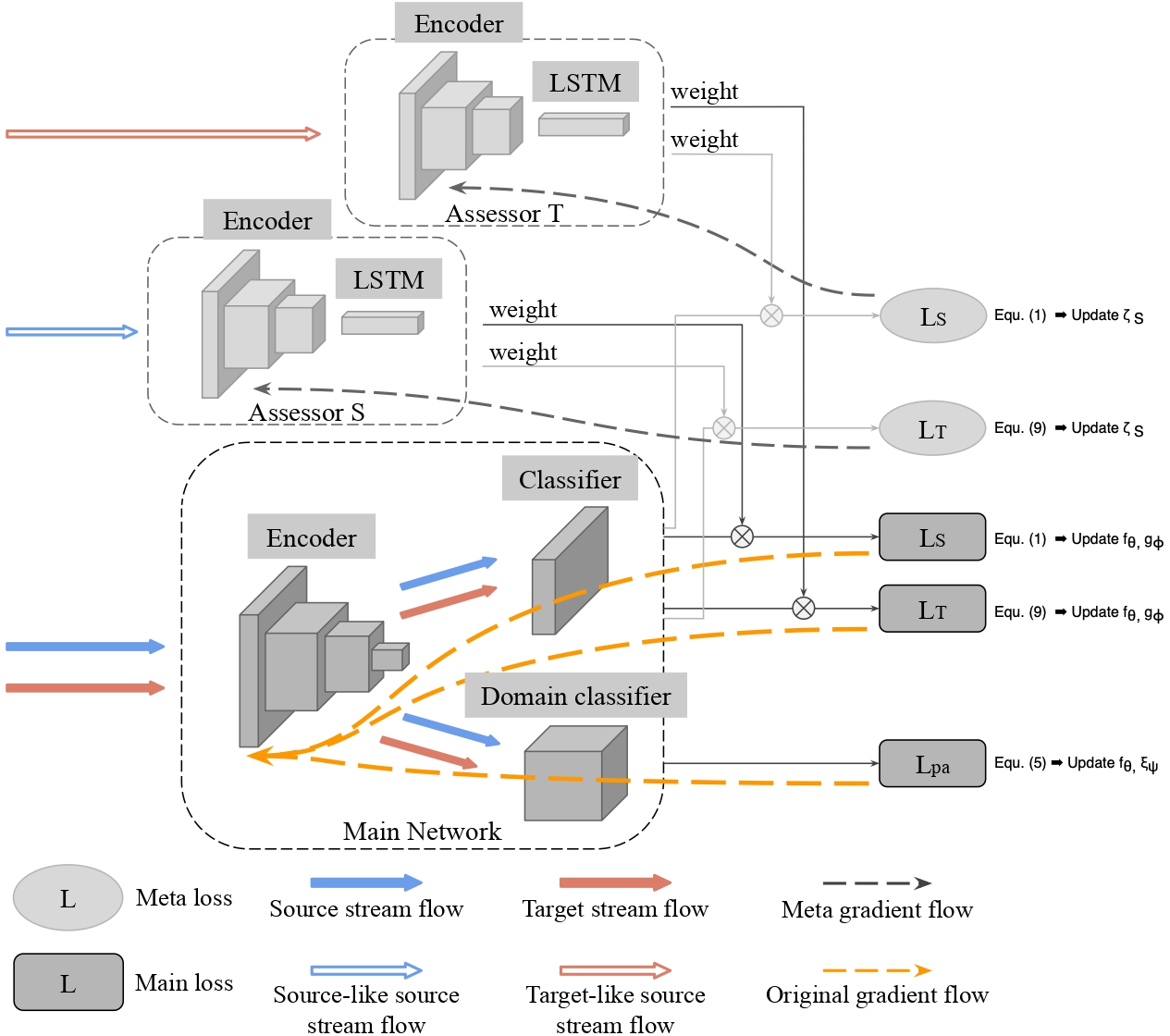}
    \caption{CLAMP architecture includes the main module (encoder $f_{\theta}$, classifier $g_{\phi}$ and domain classifier ${\xi}_{\psi}$) and two assessors ($\zeta_S$ and $\zeta_T$). In cross-domain continual learning scenario, the adversarial learning process of the base learner, encoder and classifier is adjusted by two assessors trained in the meta-learning method. Domain classifier of main network aims to solve unsupervised domain adaptation.}
    \label{fig:network structure}
\end{figure}

CLAMP relies on the class-aware adversarial training strategy \cite{Ganin2015DomainAdversarialTO,Li2023MetaReweightedRF} comprising the feature extractor $f_{\theta}(.)$, the classifier $g_{\phi}(.)$ and the domain classifier $\xi_{\psi}(.)$. The feature extractor $f_{\theta}(.)$ is trained in an adversarial manner to fool the domain classifier $\xi_{\psi}(.)$ classifying whether a data sample belongs from the target domain or the source domain. A domain-invariant network is produced if the feature extractor generates indistinguishable samples. In other words, the feature extractor and the domain classifier play a minimax game with a gradient reversal strategy to update the feature extractor converting the minimization problem into the maximization problem. The classifier $g_{\phi}(.)$ delivers the final prediction of the network. The continual learning strategy across the source domain and the target domain is formulated as a meta-weighted combination of the three loss functions, the cross entropy loss function, the dark experience replay loss function \textcolor{black}{Buzzega et al.} \cite{Buzzega2020DarkEF} and the knowledge distillation loss function \textcolor{black}{Rebuffi et al.} \cite{Rebuffi2016iCaRLIC}. That is, two assessors trained in the meta-learning manner are deployed to guide the learning process of the base learner, i.e., they control the sample influences making sure only positive samples to obtain high weights as well as the interaction of the three loss functions determining the best combination in learning a sample - proper trade off between the plasticity and the stability. This approach makes use of an episodic memory $\mathcal{M}$ based on reservoir sampling where two episodic memories, namely the source memory $\mathcal{M}_S$ and the target memory $\mathcal{M}_T$, are used here. Fig. \ref{fig:network structure} shows the network structure of CLAMP.


\subsection{Training Strategy of Source Domain}
Referring to \textcolor{black}{Ben David et al.} \cite{BenDavid2006AnalysisOR,BenDavid2010ATO}, the generalization power of the target domain is upper bounded by the empirical error of the source domain. This requires a model to be trained in the source domain which can be done simply by minimizing the cross entropy loss function because of the presence of true class labels. Nonetheless, this process risks on the negative transfer problem. Our approach is inspired by the assessor-guided training process \textcolor{black}{Zheng et al. and Masum et al.} \cite{Zheng2020DeepML,Masum2023AssessorGuidedLF} performing the soft-weighting mechanism for every sample of the source domain. We adapt this approach to the cross-domain continual learning problem where the assessor generates a set of weights corresponding to the three loss functions, the cross entropy loss function, the DER loss function \textcolor{black}{Buzzega et al.} \cite{Buzzega2020DarkEF} and the distillation loss function \textcolor{black}{Rebuffi et al.} \cite{Rebuffi2016iCaRLIC} thereby leading to a suitable weighted linear combination in learning a data sample. That is, a sequence-aware assessor produces a set of weights $\{\alpha_S,\beta_S,\gamma_S\}=\kappa_{\zeta_{S}}^{S}(.)\in[0,1]$ where $\zeta_S$ is the parameters of the assessor. $\alpha_S,\beta_S,\gamma_S$ respectively stand for the cross entropy weight, the DER weight and the distillation weight. A high weight, close to one, is provided to good samples whereas a low weight is produced to poor samples minimizing its effects. The loss function of the source domain is set as follows:
\begin{equation}\label{loss_source}
    \begin{split}
        \mathcal{L}_{S}=\underbrace{\mathbb{E}_{(x,y)\backsim\mathcal{D}_k^S\cup\mathcal{M}_{k-1}^S}[\alpha_S l(s_{W}(o),y)]}_{L_{CE}}+\underbrace{\mathbb{E}_{(x,y)\backsim\mathcal{M}_{k-1}^S}\beta_S[||o-h||+l(s_{W}(o),y)]}_{L_{DER}}+\underbrace{\mathbb{E}_{(x,y)\backsim\mathcal{M}_{k-1}^S}[\gamma_S l(o,h)]}_{L_{distill}}
    \end{split}
\end{equation}
where $o=g_{\phi}(f_{\theta}(x))_{k},h=g_{\phi}(f_{\theta}(x))_{k-1}$ denote the current output logit, i.e. the pre-softmax response, and the previous output logit before learning the current task. $l(.)$ stands for the cross entropy loss function while $s_{W}(.)$ labels the softmax layer parameterized by $W$. The assessor $\kappa_{\zeta_{S}}^{S}(.)$ is implemented as an LSTM with a convolutional feature extractor taking into account a sequence of past weights in generating a current weight. A fully connected layer with a sigmoid activation function is inserted at a last layer assuring a bounded output in between 0 and 1. It is seen that the soft-weighting mechanism is carried out here rather than the binary hard-weighting approach to induce a smooth learning process. The cross entropy loss function focuses on the current and past concepts, the plasticity to new knowledge, whereas the DER loss function and the distillation loss function mitigates the catastrophic forgetting effect, the stability of old concepts. The meta-weighting approach to the three loss functions is applied here to achieve proper tradeoff between the plasticity and the stability. 

\textcolor{black}{The cross entropy (CE) loss function learns both current samples of the current task and past samples of the past tasks and is weighted by a meta-weight generated by the assessor. That is, it learns the current concept while retaining the previously learned knowledge to prevent the CF problem. The use of CE loss function alone does not suffice to avoid the CF problem because the number of old tasks is much more than the current task and restricted to a tiny memory storing past samples. To this end, the DER loss function by Buzzega et al. \cite{Buzzega2020DarkEF} is integrated and considers both the soft label and the hard label. The hard label is incorporated to prevent the output bias problem. As with the CE loss function, the DER loss function is meta-weighted by the assessor controlling its contributions to the training process. Since the DER loss function utilizes the L2 distance function to compare the current and previous output logits, the distillation loss function Rebuffi et al \cite{Rebuffi2016iCaRLIC} is integrated to further combat the CF problem and uses the cross entropy loss function to match the current and previous output logits. The influence of the distillation loss function is governed by the assessor. That is, the assessor controls how much a model should learn from the current and past concepts or balances the stability and the plasticity, i.e., the CE loss function learns both the current and past concepts while the DER loss function and the distillation loss function focus on the past concept in the memory.}

The meta-learning approach is adopted to train both the base learner and the assessor. Two data partitions, namely a training set $\mathcal{T}_{train}^{k}$ and a validation set $\mathcal{T}_{val}^{k}$, are created for this purpose. The training set is nothing but data samples of the current task and data samples of the episodic memory $\mathcal{T}_{train}^{k}=\mathcal{T}_{k}^{S}\cup\mathcal{M}_{k-1}^{S}$ whereas the validation set applies the random transformation approach to the training set \textcolor{black}{Volpi et al.} \cite{Volpi2020ContinualAO}. Specifically, three random transformations, namely imageinvert,  Gaussian noise perturbation,  RGB-rand perturbation, are applied to the original images of the training set $\mathcal{T}_{val}^{k}=\{T(x_i),y_i\}_{i=1}^{N_{train}}\backsim\mathcal{T}_{train}^k$ where $T(.)$ is an image transformation operation. The training process of the base learner and the assessor is formulated as a bi-level optimization problem where the base learner is updated using the training set in the outer loop while the assessor utilizes the validation set in the inner loop. 
\begin{equation}\label{bilevel1}
    \begin{split}
    \{\theta^*,\phi^*\}=\arg\min_{\theta,\phi}\mathbb{E}_{(x,y)\backsim\mathcal{T}_{train}^{k}}\mathcal{L}_S\\
        s.t \quad \min_{\zeta_S}\mathbb{E}_{(x,y)\backsim\mathcal{T}_{val}^{k}}\mathcal{L}_{S}
     \end{split}
\end{equation}
where $\{\theta^*,\phi^*\}$ are optimal parameters of the base learner in respect to the current assessor $\zeta_{S}$. Because of the absence of ground truth, the meta-training strategy is carried out where the assessor is fine-tuned to minimize the validation loss of the base learner. That is, the base learner is evaluated using the validation set $\mathcal{T}_{val}^k$ returning the validation loss. This implies the use of a meta-objective in adapting the base learner. The validation loss is differentiable in respect to the assessor because it is induced by the cross entropy weight $\alpha$, the DER weight $\beta$ and the distillation weight $\gamma$. The assessor is updated first here because it determines the optimal base network:
\begin{equation}\label{ml1}
  \zeta_{S}'=\zeta_{S}-\beta\sum_{\mathclap{(x,y)\in\mathcal{T}_{val}^{k}}}\nabla_{\zeta_S}\mathcal{L}_S
\end{equation}
where $\beta$ is the learning rate of the inner loop. The next step is to train the base network making use of the three meta-weights of the assessor as follows:
\begin{equation}\label{ml2}
      \{\phi,\theta\}=\{\phi,\theta\}-\mu\sum_{\mathclap{(x,y)\in\mathcal{T}_{train}^{k}}}\nabla_{\{\phi,\theta\}}\mathcal{L}_S
\end{equation}
where $\mu$ is the learning rate of the outer loop. This strategy implies a joint update of the base network and the assessor where they work in the collaborative manner, i.e., the adjustment of the base learner involves the tuning process of the assessor. It also simulates the training and testing phases of the base learner where the training process is done using the training set with the help of the assessor. Its generalization power is evaluated using the validation set where the validation loss controls the training process of the assessor. It leads to an improved performance of the base learner because the base learner is adjusted using the three meta-weights generated by an updated assessor $\{\alpha_S,\beta_S,\gamma_S\}=\kappa_{\zeta_{S}'}^{S}(x)$. 
\subsection{Unsupervised Domain Adaptation Strategy}
\textcolor{black}{A domain adaptation step is required here to achieve a domain-invariant network because the source domain and the target domain are different but related. That is, they follow different distributions $P(x_s)\neq P(x_t)$ and are drawn from distinct feature spaces $\mathcal{X}_S\neq\mathcal{X}_T$ but are related because of the same label spaces. This process has to be carried out in an unsupervised way due to the absence of labelled samples of the target domain. Our approach makes use of the class-aware adversarial domain adaptation strategy \textcolor{black}{Ganin et al. and Li et al.} \cite{Ganin2015DomainAdversarialTO,Li2023MetaReweightedRF} deploying the domain classifier $\hat{d}_{i}=\xi_{\psi}(f_{\theta}(x_i))$ formed as a single hidden layer network to classify the origin of features generated by the feature extractor $f_{\theta}(.)$, i.e., the domain classifier is tasked to perform a binary classification problem whether a data sample belongs to the source domain or the target domain. The domain adaptation loss is formulated as the domain classifier loss:
\begin{equation}
\begin{split}
      \mathcal{L}_{pa}=\mathbb{E}_{(x,d_n)\backsim D_k^S\cup \mathcal{M}_{k-1}^S}[\log(\xi_{\psi}(f_{\theta}(x)))]+\mathbb{E}_{(x,d_n)\backsim D_k^T\cup \mathcal{M}_{k-1}^T}[\log(1-\xi_{\psi}(f_{\theta}(x)))]\\= \frac{1}{N_s}\sum_{i=1}^{N_S}\mathcal{L}_{d}(d_{i},\hat{d}_{i})+\frac{1}{N_T}\sum_{i=1}^{N_t}\mathcal{L}_{d}(d_{i},\hat{d}_{i})\\\mathcal{L}_{d}(d_i,\hat{d}_{i})=-[d_i\log{(\hat{d}_i)}+(1-d_i)\log(1-\hat{d}_{i})]  
\end{split}
\end{equation}
where $d_n\in\{0,1\}$ is the origin of data samples, i.e., $1$ for the source domain and $0$ for the target domain. In a nutshell, the loss function $\mathcal{L}$ is optimized by searching for a saddle point solution $\phi,\psi,\theta$ of the minimax problem.
\begin{equation}
    \begin{split}
    (\hat{\phi},\hat{\theta})=\arg\min_{\phi,\theta}\mathcal{L}(\phi,\theta,\hat{\psi})\\\hat{\psi}=\arg\max_{\psi}\mathcal{L}(\hat{\phi},\hat{\theta},\psi)
    \end{split}
\end{equation}
The gradient reversal layer is applied when adjusting the feature extractor thus converting the minimization problem into the maximization problem. This strategy minimizes the divergence of the source domain and the target domain because the feature extractor is trained to produce indistinguishable feature representations, i.e., maximizing similarities across the two domain thereby fooling the domain classifier. The gradient reversal layer has no parameters and simply reverses the update directions, i.e, multiplying the gradient with $-1$. We apply the stochastic gradient descent method for model updates.
\begin{equation}
    \begin{split}
        \theta=\theta+\lambda\frac{\partial\mathcal{L}_{d}^{i}}{\partial\theta}\\\phi=\phi-\lambda\frac{\partial\mathcal{L}_{d}^{i}}{\partial\phi}\\ \psi=\psi-\lambda\frac{\partial\mathcal{L}_{d}^{i}}{\partial\psi}
    \end{split}
\end{equation}
where the positive sign of the feature extractor update rule is as a result of the gradient reversal layer. Although this approach is effective in closing the gap of the source domain and the target domain, it only executes the global alignment approach excluding the class-wise discriminability on representations. CLAMP also involves the training process of the target domain based on the pseudo labelling approach to enhance the generalization performance. }
\subsection{Training Strategy of Target Domain}
The vanilla adversarial domain adaptation approach only performs a global domain alignment strategy ignoring the class alignment of the two domains \textcolor{black}{Li et al.} \cite{Li2023MetaReweightedRF}. This problem is addressed here using the pseudo labelling strategy of the target domain. Pseudo labelled samples of the target domain are learned in the discriminative fashion using the same loss as the source domain \eqref{loss_source}. In other words, the self-learning mechanism is carried out for the target domain where a pseudo-label is derived from a predictive output of a model. The pseudo-labelling approach is undertaken:
\begin{equation}\label{pseudo}
\begin{split}
    \hat{y}^T=\arg\max_{o=1,...,m}P(y|x),
    P(y|x)=s_{W}(g_{\phi}(f_{\theta}(x^T)))
\end{split}    
\end{equation}
The pseudo labelling strategy returns the training set of the target domain $\mathcal{T}_{ps}^{k}=\{(x_i^T,\hat{y}_i^{T})\}_{i=1}^{N_{k}^{ps}}$ where $N_{k}^{ps}$ denotes the number of pseudo-labelled samples of the target domain of the $k-th$ task. Note that the pseudo-labelling mechanism is also performed to those of the target episodic memory $\mathcal{M}_{k-1}^{T}$. Nevertheless, this mechanism risks on noisy pseudo labels undermining model's generalization performance. The assessor-guided learning mechanism using the sequence-aware assessor is implemented here to reject noisy pseudo-labels. We adopt similar strategy as \textcolor{black}{Li et al.} \cite{Li2023MetaReweightedRF} but the key difference lies in the use of the sequence-aware target assessor $\kappa_{\zeta_T}^{T}(.)\in[0,1]$. As with the source domain, the target assessor delivers a set of weights $\alpha_T,\beta_T,\gamma_T$ rather than a single weight as per \textcolor{black}{Li et al.} \cite{Li2023MetaReweightedRF} to achieve a proper tradeoff between plasticity and stability in the continual learning setting. The loss function of the target domain is formulated:
\begin{equation}\label{loss_target}
    \begin{split}
        \mathcal{L}_{T}=\underbrace{\mathbb{E}_{(x,y)\backsim\mathcal{D}_k^T\cup\mathcal{M}_{k-1}^T}[\alpha_T l(s_{W}(o),y)]}_{L_{CE}}+\underbrace{\mathbb{E}_{(x,y)\backsim\mathcal{M}_{k-1}^T}\beta_T[||o-h||+l(s_{W}(o),y)]}_{L_{DER}}+\underbrace{\mathbb{E}_{(x,y)\backsim\mathcal{M}_{k-1}^T}[\gamma_T l(o,h)]}_{L_{distill}}
    \end{split}
\end{equation}
 As with the source domain, the assessor of the target domain performs the soft-weighting mechanism rejecting noisy pseudo-labels by allocating small weights. The assessor of the target domain is trained with a meta-objective using the validation loss of the base network. 

The validation set $\mathcal{T}_{vl}^{k}=\{(x_i^{S},y_i^{S})\}_{i=1}^{N_k^{val}}$is constructed from highly similar samples of the source domain \textcolor{black}{Li et al.} \cite{Li2023MetaReweightedRF}. Such samples are detectable from the output of the domain classifier $\xi_{\psi}(f_{\theta}(x))$ where highly similar samples must be close to $0.5$. That is, we measure the absolute differences of the domain classifier and 0.5 as $|\xi_{\psi}(f_{\theta}(x))-0.5|$. The differences are sorted in the descending order and the top $p=2$ samples are taken for each class making sure a balanced class proportion $N_k^{val}=p*m$. As with the training set, the validation set includes data samples of the source episodic memory $\mathcal{M}_{k-1}^{S}$. The meta training process is carried out similarly as the training strategy of the source domain where the bi-level optimization strategy is formulated:
\begin{equation}\label{bilevel2}
    \begin{split}
        \{\theta^*,\phi^*\}=\arg\min_{\theta,\phi}\mathbb{E}_{(x,y)\backsim\mathcal{T}_{vl}^{k}}\mathcal{L}_T\\
        s.t \quad
        \min_{\zeta_T}\mathbb{E}_{(x,y)\backsim\mathcal{T}_{ps}^{k}}\mathcal{L}_T
    \end{split}
\end{equation}
where $\{\theta^*,\phi^*\}$ are the optimal parameters of the base learner with respect to the current assessor of the target domain $\kappa_{\zeta_T}^T(.)$. The inner loop focuses on the training strategy of the assessor using the training set while the outer loop is directed to the training process of the base learner using the validation set. Both the inner and outer loops are solved with the SGD approach since they only call for the first order optimization process. The base learner is evaluated with the validation set resulting in the validation loss utilized to update the target assessor.
\begin{equation}\label{inner1}
    \zeta_{T}'=\zeta_{T}-\beta\sum_{\mathclap{(x,y)\in\mathcal{T}_{vl}^{k}}}\nabla_{\zeta_T}\mathcal{L}_T
\end{equation}
where $\beta$ is the learning rate of the inner loop. The updated target assessor produces the three weights $\{\alpha_T,\beta_T,\gamma_T\}=\kappa_{\zeta_{T}'}^{T}(x)$ navigating the learning process of the base learner. 
\begin{equation}\label{ml5}
    \{\phi,\theta\}=\{\phi,\theta\}-\mu\sum_{\mathclap{(x,y)\in\mathcal{T}_{ps}^{k}}}\nabla_{\{\phi,\theta\}}\mathcal{L}_{T}
\end{equation}
where $\mu$ is the learning rate of the outer loop. A joint update of the base learner and the assessor is presented here where every outer loop contains the inner loop. Both assessor and base learner work in collaborative manner. 

\subsection{Loss Function}
The overall loss function of CLAMP for the $k-th$ task is written:
\begin{equation}
    \mathcal{L}_{all}=\mathcal{L}_S-\mathcal{L}_{pa}+\mathcal{L}_T
\end{equation}
where a negative sign is inserted to realize the gradient reversal strategy applied to the feature extractor $f_{\theta}(.)$. The alternate optimization approach is implemented here where at first CLAMP is trained to minimize $\mathcal{L}_{S}$ affecting the classifier and the feature extractor $g_{\phi}(f_{\theta}(.))$. It is done in the meta-learning manner with the inner loop to update the source assessor $\kappa_{\zeta_{S}^{S}}(.)$ and the outer loop to tune the base network $g_{\phi}(f_{\theta}(.))$. The domain adaptation step is carried out afterward where the domain classifier $\xi_{\psi}(.)$ and the feature extractor $f_{\theta}(.)$ are updated in respect to $\mathcal{L}_{pa}$ where the gradient reversal layer is applied to the feature extractor $-\frac{\partial\mathcal{L}_k^{ps}}{\partial\theta}$ whereas a normal gradient update is taken for the domain classifier $\frac{\partial\mathcal{L}_k^{ps}}{\partial\psi}$. The next stage addresses the target domain minimizing $\mathcal{L}_{T}$. As with the source domain, it is carried out with the meta learning approach where the target assessor $\kappa_{\zeta_{T}^{T}}(.)$, inner loop, and the base network $g_{\phi}(f_{\theta}(.))$, outer loop, work collaboratively.
\section{Theoretical Analysis}
\noindent\textbf{Theorem 1 Ben david et al and Lao et al. \cite{BenDavid2010ATO,Lao2021ATC}}: under a non-continual learning framework, the generalization error of the target domain $\mathcal{D}^T$ is upper bounded by the empirical error of a model on the source domain $\mathcal{D}^S$ plus discrepancy between two domains:
\begin{equation}
    \epsilon_{T}\leq\epsilon_{S}+d(\mathcal{D}^S,\mathcal{D}^T)+g*
\end{equation}
where $d(\mathcal{D}^S,\mathcal{D}^T)$ is the discrepancy of the two domains which can be modeled by the $\mathcal{H}$ divergence while $g*$ is the error of an optimal classifier for both source and target domains. This is expandable for the continual learning setting where sequences of different tasks are observed from the the unlabelled target domain and the labelled source domain $\mathcal{T}_k^S,\mathcal{T}_k^T,k\in[1,...,K]$. Let $\lambda_k=d(\mathcal{D}_k^S,\mathcal{D}_k^T)$ be the discrepancy of the source domain and the target domain in the embedding space $f_{\theta}(x^S)$ and $f_{\theta}(x^T)$. The error of the $k-th$ task of the target domain is formalized:
\begin{equation}
    \epsilon_{T_k}\leq\epsilon_{S_k}+\lambda_k+g*
\end{equation}
where $g*=\arg\min_{\phi,\theta}(\epsilon_{T_k}+\epsilon_{S_k})$ is the error of an optimal classifier for both source domain and target domain. 

\noindent\textbf{Theorem 2 Lao et al. \cite{Lao2021ATC}}: the total error of the target domain after executing $K$ tasks is bounded as follows:
\begin{equation}
    \epsilon_{T}\leq\sum_{k=1}^{K}(\epsilon_{S_k}+\lambda_{k})+\sum_{k=1}^{K-1}(\epsilon_{\mathcal{M}_{k}^{S}}+\epsilon_{\mathcal{M}_{k}^{T}})+g*
\end{equation}
where $\epsilon_{\mathcal{M}_k}^{S},\epsilon_{\mathcal{M}_k}^{T}$ are the compression errors due to the episodic memory for experience replay.

\noindent\textit{Proof}: we use $\hat{\epsilon}_k^{T}$ to estimate $\epsilon_k^T$ because the episodic memory is used to generate previous experiences $k\in[1,K-1]$. The total error of the target domain is thus formalised as follows: 
\begin{equation}
    \begin{split}
        \epsilon_{T}&=\epsilon_{T_k}+\sum_{k=1}^{K-1}(\hat{\epsilon}_{T_k}+\epsilon_{\mathcal{M}_{k}^{T}})\\
        &=\epsilon_{S_k}+\lambda_k+\sum_{k=1}^{K-1}(\hat{\epsilon}_{S_k}+\lambda_k+\epsilon_{\mathcal{M}_{k}^{T}})+g*
    \end{split}
\end{equation}
As with the target domain, the episodic memory $\mathcal{M}_{k}^{S}$ is applied in the source domain to perform the experience replay mechanism. This issue results in the compression error. Hence, the error of the source domain at the $k-th$ task is defined as follows:
\begin{equation}
    \hat{\epsilon}_{S_k}=\epsilon_{S_k}+\epsilon_{\mathcal{M}_{k}^{S}}
\end{equation}
The total error of the target domain is defined as follows:
\begin{equation}
    \begin{split}
        \epsilon_{T}&\leq\epsilon_{S_k}+\lambda_k+\sum_{k=1}^{K-1}(\epsilon_{S_k}+\lambda_k+\epsilon_{\mathcal{M}_{k}^{T}}+\epsilon_{\mathcal{M}_{k}^{S}})+g*\\
        &=\sum_{k=1}^{K}\epsilon_{S_k}+\lambda_k+\sum_{k=1}^{K-1}(\epsilon_{\mathcal{M}_{k}^{T}}+\epsilon_{\mathcal{M}_{k}^{S}})+g*
    \end{split}
\end{equation}
The error bound of the target domain implies the importance of CLAMP's learning modules. The training strategy of the source domain $\mathcal{L}_S$ contributes to reduction of $\epsilon_{S_k}$ and $\epsilon_{M_k^S}$ while the discrepancy between the source domain and the target domain $\lambda_k$ is minimized by the unsupervised domain adaptation strategy $\mathcal{L}_{pa}$. Since the training process of the target domain models the target domain directly, it has an impact on $\epsilon_T$, $\lambda_k$, and $\epsilon_{M_k^T}$. It should be carefully committed due to the risk of noisy pseudo labels.

\section{Experiments}
This section presents our numerical validations of CLAMP including benchmark problems, comparisons against prior arts, ablation study, memory analysis, sensitivity analysis and the t-sne analysis. 

\subsection{Setup}
\noindent\textbf{Datasets}: 
we evaluate our method, CLAMP, on four problems to simulate up to 21 cross-domain CL problems.
1) Digit Recognition is constructed by two domains, MNIST(MN)$\leftrightarrow$USPS(US). 
2) Office-31 \textcolor{black}{Saenko et al.} \cite{Saenko2010AdaptingVC} has 4652 colored images from three domains: Amazon (A), DSLR (D) and Webcam (W). Each domain consists of 31 classes. 
3) Office-Home \textcolor{black}{Ventakesware et al.} \cite{Venkateswara2017DeepHN} contains four distinct domains: Artistic images (Ar), Clip art (Cl), Product images (Pr), and Real-World images (Rw) where each domain has 65 classes and 12 problems are set. 
4) VisDA 2017 \textcolor{black}{Peng et al.} \cite{Peng2017VisDATV} consists of a source dataset of synthetic 2D rendering of 3D models generated from different angles and lightning conditions while a target dataset contains  photo-realistic images.

\noindent\textbf{Task Division}: We detail the task division for each dataset. 1) for the digit recognition case (MNIST and USPS), we divide the 10 classes in each domain into 5 tasks, where each task encompasses 2 classes. 2) for Office-31,  we sort the classes in alphabetically ascending order and divide each domain into 5 tasks. The first four tasks include 6 classes each, and the last task includes 7 classes. 3) for Office-Home, we split the 65 classes into 13 subsets of 5 classes per task. 4) for VisDA problem, there exist 12 classes split into 4 tasks of 3 classes each.

\noindent\textbf{Benchmark Algorithms}:
\textcolor{black}{CLAMP is compared against ten prior arts: EWC Kirkpatrick et al. \cite{Kirkpatrick2016OvercomingCF}, LwF Li et al. \cite{Li2016LearningWF}, SI Zenke et al. \cite{Zenke2017ContinualLT}, MAS Aljundi et al. \cite{aljundi2018memory}, RWalk Chaudhry et al. \cite{Chaudhry2018RiemannianWF}, iCARL Rebuffi et al. \cite{Rebuffi2016iCaRLIC}, IL2M Belouadah et al.\cite{Belouadah2019IL2MCI}, EEIL Castro et al. \cite{Castro2018EndtoEndIL}, DANN Ganin et al. \cite{Ganin2015DomainAdversarialTO}, HAL Chaudhry et al. \cite{Chaudhry2019UsingHT}, AGLA Masum et al. \cite{Masum2023AssessorGuidedLF} and CDCL De Carvalho et al. \cite{VinciusdeCarvalho2024TowardsCC}. No comparisons are done against Lao et al. \cite{Lao2021ATC} due to the absence of its public source codes. \textit{The source codes for CLAMP, including benchmarks and task split details, are publicly available at \url{https://github.com/wengweng001/CLAMP_torch.git}}.}

\noindent\textbf{Evaluation Metric}: \textcolor{black}{the performance of consolidated algorithms are evaluated using the average classification accuracy of unlabelled target domain assessing their performances for all already seen tasks, thus reflecting the existence of CF problems Chaudhry et al. \cite{Chaudhry2018EfficientLL}. The target domain is chosen to measure their performances against the domain shift problem whether or not the domain adaptation mechanism succeeds. The average accuracy $\mathcal{A}\in[0,1]$ is mathematically expressed as follows:
\begin{equation}
    \mathcal{A}_{\uparrow}=\frac{1}{k}\sum_{j=1}^{k}a_{k,B_{k,j}}
\end{equation}
where $a_{k,B_{k,j}}\in[0,1]$ denotes the accuracy evaluated on the test set of task $j$ after the model has been trained with the task $k$. In a nutshell, the average accuracy $\mathcal{A}_{T}$ evaluates the accuracy of all tasks after learning the last task. 
}

\noindent\textbf{Implementation Details}:
We report the average accuracy over five different runs under the same computational resources to ensure fair comparisons; however, for VisDA, the average accuracy is reported from three runs. Our numerical study is carried out under the FACIL framework \textcolor{black}{Masana et al.} \cite{Masana2020ClassincrementalLS} to ensure fair comparisons where the hyper-parameters of all algorithms are selected using the grid search approach and reported in the appendix. 
LeNet is used as a backbone network for the digit recognition problem, while for Office-31 and VisDA, we use ResNet-34 \textcolor{black}{He et al.} \cite{He2015DeepRL} pretrained on ImageNet \textcolor{black}{Deng et al} \cite{5206848}, and for the Office-Home, we apply pretrained ResNet-50 \textcolor{black}{Deng et al.} \cite{5206848}. We train our model using SGD optimizer. Batch size is 128 for digit recognition and 32 for Office-31, Office-Home, and VisDA. 

\begin{table*}[!htp]\centering
\caption{Average query accuracy (\%) of all learned tasks on digit recognition, MNIST(MN) \& USPS(US), and VisDA.}\label{tab: results-mn-us}
\scriptsize
\begin{tabular}{l|cccccc}\toprule
& MN$\rightarrow$US & US$\rightarrow$MN & VisDA\\
\midrule
Source Only & 17.80 $\pm$ 0.2 & 12.55 $\pm$ 0.9 & 12.85 $\pm$ 1.2 \\
Joint Training & 77.81 $\pm$ 1.9 & 49.81 $\pm$ 0.7 & 44.76 $\pm$ 0.9 \\
\midrule
EWC   & 17.88 $\pm$ 0.3 & 13.30 $\pm$ 1.8 & 12.65 $\pm$ 3.5 \\
LwF   & 17.74 $\pm$ 0.4 & 12.94 $\pm$ 1.5 & 13.10 $\pm$ 1.2 \\
SI    & 17.80 $\pm$ 0.2 & 12.54 $\pm$ 1.0 & 13.40 $\pm$ 2.1 \\
MAS   & 17.92 $\pm$ 0.3 & 13.18 $\pm$ 2.0 & 13.45 $\pm$ 1.0 \\
RWalk & 17.92 $\pm$ 0.3 & 13.39 $\pm$ 1.7 & 13.91 $\pm$ 1.7 \\
iCaRL & 42.11 $\pm$ 2.5 & 24.04 $\pm$ 0.9 & 15.67 $\pm$ 1.7 \\
IL2M  & 43.67 $\pm$ 3.3 & 23.16 $\pm$ 1.9 & 19.58 $\pm$ 1.4 \\
EEIL  & 38.75 $\pm$ 3.0 & 22.46 $\pm$ 1.4 & 15.99 $\pm$ 1.8 \\
HAL  & 72.31 $\pm$ 1.4 & 47.55 $\pm$ 2.3 & 16.87 $\pm$ 1.6\\
\textcolor{black}{AGLA}  & 62.75 $\pm$ 4.2 & 47.98 $\pm$ 1.1 & 25.8 $\pm$ 2.7 \\
\textcolor{black}{CDCL}  & 66.73 & 52.50 & 10.16 \\
DANN  & 17.89 $\pm$ 0.9 & 16.90 $\pm$ 3.1 & 16.16 $\pm$ 0.1 \\
\textbf{CLAMP} & \textbf{84.98 $\pm$ 1.3} & \textbf{89.63 $\pm$ 1.0} & \textbf{30.71 $\pm$ 1.8} \\
\bottomrule
\end{tabular}
\\[1pt]
\textit{The results for CDCL are taken from the original paper.}
\end{table*}

\begin{table*}[!htp]\centering
\caption{Average query accuracy (\%) of all learned tasks on Office-31.}\label{tab: results-office31}
\resizebox{\linewidth}{!}{
\begin{tabular}{l|ccccccc}\toprule
Source &A &A &D &D &W &W &\multirow{2}{*}{\textit{Office-31 Avg.}} \\
Target &D &W &A &W &A &D & \\
\midrule
Source Only & 14.96 $\pm$ 3.0 & 14.90 $\pm$ 1.6 & 8.44 $\pm$ 2.5 & 20.07 $\pm$ 2.8 & 14.91 $\pm$ 3.0 & 24.29 $\pm$ 3.0 & \textit{16.26 $\pm$ 5.4} \\
Joint Training & 68.59 $\pm$ 2.6 & 59.27 $\pm$ 1.3 & 42.67 $\pm$ 2.8 & 80.88 $\pm$ 2.9 & 49.24 $\pm$ 2.0 & 90.43 $\pm$ 2.3 & \textit{65.18 $\pm$ 18.4} \\
\midrule
EWC & 15.05 $\pm$ 2.8 & 15.09 $\pm$ 1.5 & 8.67 $\pm$ 2.7 & 20.25 $\pm$ 2.3 & 14.64 $\pm$ 3.4 & 25.65 $\pm$ 3.0 & \textit{16.56 $\pm$ 5.8} \\
LwF & 19.85 $\pm$ 2.1 & 18.30 $\pm$ 1.9 & 7.67 $\pm$ 2.4 & 21.18 $\pm$ 4.8 & 16.23 $\pm$ 2.3 & 27.20 $\pm$ 2.4 & \textit{18.41 $\pm$ 6.4} \\
SI & 15.59 $\pm$ 2.3 & 14.67 $\pm$ 1.9 & 8.38 $\pm$ 2.7 & 19.86 $\pm$ 2.9 & 14.91 $\pm$ 3.0 & 24.29 $\pm$ 2.3 & \textit{16.28 $\pm$ 5.4} \\
MAS & 15.95 $\pm$ 3.9 & 14.90 $\pm$ 1.5 & 8.67 $\pm$ 2.7 & 20.25 $\pm$ 2.3 & 14.52 $\pm$ 3.4 & 25.65 $\pm$ 3.0 & \textit{16.66 $\pm$ 5.8} \\
RWalk & 15.95 $\pm$ 3.0 & 15.09 $\pm$ 1.5 & 8.61 $\pm$ 2.7 & 20.25 $\pm$ 2.3 & 14.64 $\pm$ 3.4 & 25.28 $\pm$ 3.0 & \textit{16.64 $\pm$ 5.6} \\
iCaRL & 47.68 $\pm$ 3.8 & 37.87 $\pm$ 1.8 & 38.77 $\pm$ 5.2 & 79.84 $\pm$ 3.6 & 42.34 $\pm$ 1.8 & 88.60 $\pm$ 1.8 & \textit{55.85 $\pm$ 22.4} \\
IL2M & 48.12 $\pm$ 2.0 & 43.18 $\pm$ 3.3 & 39.95 $\pm$ 5.4 & 79.61 $\pm$ 2.2 & 42.84 $\pm$ 1.7 & 86.46 $\pm$ 1.6 & \textit{56.69 $\pm$ 20.7} \\
EEIL & 45.51 $\pm$ 2.3 & 41.02 $\pm$ 3.9 & 40.23 $\pm$ 5.0 & 79.21 $\pm$ 2.4 & 42.76 $\pm$ 1.9 & 86.72 $\pm$ 1.9 & \textit{55.91 $\pm$ 21.2} \\
HAL   & 33.51 $\pm$ 1.4 & 32.55 $\pm$ 3.9 & 25.62 $\pm$ 1.7 & 62.81 $\pm$ 2.8 & 29.42 $\pm$ 3.3 & 53.52 $\pm$ 2.4 & \textit{39.57 $\pm$ 2.58} \\
\textcolor{black}{AGLA}  & 55.83 $\pm$ 5.98 & 49.81 $\pm$ 3.77 & 29.61 $\pm$ 4.08 & 62.50 $\pm$ 6.02 & 31.26 $\pm$ 1.20 & 56.85 $\pm$ 5.59 & 47.64 $\pm$ 13.93 \\
\textcolor{black}{CDCL}  & 9.68 & 10.98 & 7.02 & 27.97 & 6.73 & 28.98 & \textit{15.23} \\
DANN & 9.03 $\pm$ 1.2 & 17.77 $\pm$ 2.8 & 2.07 $\pm$ 1.1 & 4.73 $\pm$ 3.9 & 17.09 $\pm$ 2.0 & 17.88 $\pm$ 8.2 & \textit{11.43 $\pm$ 7.1} \\
\textbf{CLAMP} & \textbf{84.29 $\pm$ 2.9} & \textbf{80.39 $\pm$ 2.0} & \textbf{61.37 $\pm$ 1.3} & \textbf{94.85 $\pm$ 0.3} & \textbf{59.08 $\pm$ 4.2} & \textbf{98.47 $\pm$ 2.2} & \textbf{\textit{79.74 $\pm$ 16.5}} \\
\bottomrule
\end{tabular}}
\\[1pt]
\textit{The results for CDCL are taken from the original paper.}
\end{table*}

\begin{sidewaystable}[htp]
\centering
\caption{Average query accuracy (\%) of all learned tasks on Office-Home.}\label{tab: results-home}
\resizebox{\linewidth}{!}{
\begin{tabular}{l|ccccccccccccc}\toprule
Method &Ar$\rightarrow$Cl &Ar$\rightarrow$Pr &Ar$\rightarrow$Ew &Cl$\rightarrow$Ar &Cl$\rightarrow$Pr &Cl$\rightarrow$Rw &Pr$\rightarrow$Ar &Pr$\rightarrow$Cl &Pr$\rightarrow$Rw &Rw$\rightarrow$Ar &Rw$\rightarrow$Cl &Rw$\rightarrow$Pr & \textit{Avg. }\\\midrule
Source Only &7.75 $\pm$ 1.7 &11.14 $\pm$ 1.1 &15.47 $\pm$ 1.3 &5.11 $\pm$ 2.6 &7.89 $\pm$ 0.8 &7.32 $\pm$ 0.7 &7.09 $\pm$ 1.5 &6.62 $\pm$ 1.1 &10.38 $\pm$ 0.3 &13.70 $\pm$ 0.9 &8.28 $\pm$ 1.7 &13.89 $\pm$ 0.1 &\textit{9.55 $\pm$ 3.3} \\
Joint Training &37.17 $\pm$ 1.3 &57.04 $\pm$ 1.1 &67.62 $\pm$ 1.9 &40.96 $\pm$ 2.2 &48.90 $\pm$ 1.5 &54.86 $\pm$ 1.0 &43.77 $\pm$ 1.6 &34.25 $\pm$ 2.5 &64.62 $\pm$ 1.2 &60.06 $\pm$ 1.7 &40.72 $\pm$ 1.2 &72.53 $\pm$ 1.1 &\textit{51.88 $\pm$ 12.7} \\\midrule
EWC &7.34 $\pm$ 1.6 &11.68 $\pm$ 1.5 &15.38 $\pm$ 1.4 &5.06 $\pm$ 1.6 &7.70 $\pm$ 0.6 &7.51 $\pm$ 0.7 &7.68 $\pm$ 1.4 &6.42 $\pm$ 0.7 &10.35 $\pm$ 0.6 &13.78 $\pm$ 1.5 &7.75 $\pm$ 1.8 &13.43 $\pm$ 0.2 &\textit{9.51 $\pm$3.3} \\
LwF &8.11 $\pm$ 2.1 &11.91 $\pm$ 1.4 &16.20 $\pm$ 1.3 &5.09 $\pm$ 2.6 &8.68 $\pm$ 0.8 &9.40 $\pm$ 1.1 &8.30 $\pm$ 1.6 &7.35 $\pm$ 1.2 &12.65 $\pm$ 1.7 &13.56 $\pm$ 1.3 &8.51 $\pm$ 1.7 &15.02 $\pm$ 0.7 &\textit{10.40 $\pm$ 3.4} \\
SI &7.44 $\pm$ 1.7 &11.15 $\pm$ 1.0 &15.50 $\pm$ 1.4 &4.28 $\pm$ 1.9 &7.59 $\pm$ 0.8 &7.05 $\pm$ 1.0 &7.32 $\pm$ 1.2 &6.35 $\pm$ 1.3 &9.92 $\pm$ 1.0 &12.85 $\pm$ 1.4 &7.87 $\pm$ 1.8 &13.73 $\pm$ 0.4 &\textit{9.25 $\pm$ 3.4} \\
MAS &7.38 $\pm$ 2.1 &11.59 $\pm$ 1.0 &15.51 $\pm$ 2.0 &4.69 $\pm$ 1.8 &7.42 $\pm$ 1.1 &7.12 $\pm$ 0.7 &7.81 $\pm$ 1.9 &6.44 $\pm$ 1.0 &10.87 $\pm$ 0.7 &14.09 $\pm$ 2.1 &7.46 $\pm$ 1.7 &13.37 $\pm$ 0.4 &\textit{9.48 $\pm$ 3.5} \\
RWalk &7.18 $\pm$ 1.8 &11.53 $\pm$ 1.4 &15.25 $\pm$ 1.5 &5.15 $\pm$ 2.2 &7.67 $\pm$ 0.6 &7.45 $\pm$ 0.4 &7.53 $\pm$ 1.7 &6.54 $\pm$ 1.0 &10.91 $\pm$ 0.6 &13.56 $\pm$ 1.8 &7.75 $\pm$ 1.6 &13.53 $\pm$ 0.4 &\textit{9.5 $\pm$3.3} \\
iCaRL &28.03 $\pm$ 1.0 &44.93 $\pm$ 1.3 &53.91 $\pm$ 2.1 &19.81 $\pm$ 1.0 &30.43 $\pm$ 2.0 &33.11 $\pm$ 1.3 &30.28 $\pm$ 2.1 &23.39 $\pm$ 1.5 &46.09 $\pm$ 2.1 &38.72 $\pm$ 1.4 &25.93 $\pm$ 1.4 &52.40 $\pm$ 0.9 &\textit{35.59 $\pm$ 11.4} \\
IL2M &26.60 $\pm$ 1.0 &43.82 $\pm$ 1.8 &52.72 $\pm$ 1.1 &18.03 $\pm$ 0.8 &28.14 $\pm$ 1.2 &30.85 $\pm$ 1.5 &28.17 $\pm$ 2.5 &20.24 $\pm$ 0.8 &46.12 $\pm$ 1.5 &37.52 $\pm$ 1.5 &25.85 $\pm$ 1.8 &52.08 $\pm$ 1.0 &\textit{34.18 $\pm$12.0} \\
EEIL &26.70 $\pm$ 1.2 &44.10 $\pm$ 1.9 &52.80 $\pm$ 1.2 &17.21 $\pm$ 1.0 &28.64 $\pm$ 0.8 &31.37 $\pm$ 1.2 &28.35 $\pm$ 1.6 &20.41 $\pm$ 0.9 &45.52 $\pm$ 1.1 &36.26 $\pm$ 1.6 &25.83 $\pm$ 1.4 &51.74 $\pm$ 0.9 &\textit{34.08 $\pm$ 11.9} \\
HAL & 10.90 $\pm$ 1.4 & 20.34 $\pm$ 1.4 & 29.51 $\pm$ 1.3 & 10.17 $\pm$ 1.7 & 13.04 $\pm$ 1.9 & 15.72 $\pm$ 1.9 & 14.77 $\pm$ 2.3 & 12.38 $\pm$ 1.6 & 26.32 $\pm$ 1.4 & 20.50 $\pm$ 2.4 & 13.34 $\pm$ 1.3 & 31.13 $\pm$ 1.0 & \textit{18.18 $\pm$ 1.6} \\
\textcolor{black}{AGLA}  & 18.58 $\pm$ 1.5 & 29.89 $\pm$ 2.1 & 34.20 $\pm$ 4.0 & 17.84 $\pm$ 1.5 & 23.39 $\pm$ 1.3 & 28.57 $\pm$ 2.9 & 20.02 $\pm$ 1.0 & 19.19 $\pm$ 2.3 & 32.89 $\pm$ 1.9 & 28.67 $\pm$ 5.4 & 20.61 $\pm$ 3.0 & 41.17 $\pm$ 2.6 & \textit{25.25 $\pm$ 7.5}\\
\textcolor{black}{CDCL}  & 4.15 & 4.77 & 6.01 & 5.56 & 8.32 & 6.35 & 3.92 & 5.02 & 6.29 & 6.45 & 6.26 & 8.35 & \textit{5.95} \\
DANN &4.03 $\pm$ 0.8 &4.40 $\pm$ 1.1 &5.82 $\pm$ 0.6 &2.62 $\pm$ 3.0 &5.00 $\pm$ 1.7 &5.01 $\pm$ 1.8 &4.71 $\pm$ 1.2 &3.48 $\pm$ 1.7 &5.82 $\pm$ 1.0 &6.06 $\pm$ 1.4 &4.21 $\pm$ 0.6 &7.59 $\pm$ 0.7 &\textit{4.90 $\pm$ 1.3} \\\midrule
\textbf{CLAMP} &\textbf{42.57 $\pm$ 0.9} &\textbf{62.20 $\pm$ 1.0} &\textbf{63.63 $\pm$ 1.7} &\textbf{38.50 $\pm$ 2.0} &\textbf{49.47 $\pm$ 1.5} &\textbf{49.92 $\pm$ 0.9} &\textbf{40.63 $\pm$ 3.1} &\textbf{40.45 $\pm$ 1.0} &\textbf{55.29 $\pm$ 1.9} &\textbf{54.88 $\pm$ 1.8} &\textbf{45.21 $\pm$ 1.3} &\textbf{63.18 $\pm$ 1.8} &\textit{\textbf{50.49 $\pm$ 9.3}}  \\
\bottomrule
\\[1pt]
\textit{The results for CDCL are taken from the original paper.}
\end{tabular}}

\vspace{10mm}

\caption{Ablation studies of different learning modules for CLAMP on MNIST(MN)$\leftrightarrow$USPS(US) and Office-31.}\label{tab: ablation}
\resizebox{\linewidth}{!}{
\begin{tabular}{l|ccccc|cccccccccc}\toprule
Method &PA &PL &Meta &R$_{1}$ &R$_{2}$ &MU $\rightarrow$ US &US $\rightarrow$ MN &A $\rightarrow$ D &A $\rightarrow$ W &D $\rightarrow$ A &D $\rightarrow$ W &W $\rightarrow$ A &W$\rightarrow$D & \textit{Office-31 avg.} \\\midrule
Naive &\ding{55} &\ding{55} &\ding{55} &\ding{55} &\ding{55} &17.49 $\pm$ 2.2 &17.15 $\pm$ 1.2 &13.60 $\pm$ 1.1 &15.28 $\pm$ 3.6 &12.36 $\pm$ 2.0 &29.90 $\pm$ 1.8 &16.63 $\pm$ 1.3 &33.97 $\pm$ 1.7 &\textit{20.29 $\pm$ 9.2} \\
PA &\ding{51} &\ding{55} &\ding{55} &\ding{55} &\ding{55} &18.90 $\pm$ 0.4 &16.45 $\pm$ 6.4 &13.60 $\pm$ 3.1 &13.95 $\pm$ 1.5 &14.04 $\pm$ 1.7 &28.72 $\pm$ 1.6 &14.15 $\pm$ 1.6 &35.82 $\pm$ 1.4 &\textit{20.05 $\pm$ 9.7} \\
Baseline 1 &\ding{51} &\ding{51} &\ding{55} &\ding{55} &\ding{55} &19.29 $\pm$ 0.6 &19.47 $\pm$ 0.7 &14.33 $\pm$ 2.5 &15.05 $\pm$ 2.8 &15.49 $\pm$ 2.1 &35.91 $\pm$ 1.6 &14.53 $\pm$ 2.2 &35.72 $\pm$ 2.3 &\textit{21.84 $\pm$ 10.8} \\
Baseline 2 &\ding{51} &\ding{51} &\ding{51} &\ding{55} &\ding{55} &19.21 $\pm$ 0.6 &19.46 $\pm$ 0.7 &12.10 $\pm$ 2.0 &16.81 $\pm$ 3.5 &16.01 $\pm$ 3.1 &34.64 $\pm$ 2.0 &14.92 $\pm$ 1.9 &40.16 $\pm$ 2.7 &\textit{22.44 $\pm$ 11.8} \\
Baseline 3 &\ding{51} &\ding{55} &\ding{55} &\ding{51} &\ding{55} &53.27 $\pm$ 5.4 &65.43 $\pm$ 2.3 &66.02 $\pm$ 1.6 &55.82 $\pm$ 2.8 &49.52 $\pm$ 2.6 &92.20 $\pm$ 1.6 &51.11 $\pm$ 1.4 &95.35 $\pm$ 1.8 &\textit{68.34 $\pm$ 20.1} \\
Baseline 4 &\ding{51} &\ding{51} &\ding{55} &\ding{51} &\ding{55} &58.31 $\pm$ 4.3 &58.31 $\pm$ 4.3 &74.29 $\pm$ 3.6 &67.53 $\pm$ 2.1 &55.23 $\pm$ 2.5 &92.63 $\pm$ 1.7 &55.56 $\pm$ 1.7 &96.97 $\pm$ 1.2 &\textit{73.70 $\pm$ 17.9} \\
Baseline 5 &\ding{51} &\ding{51} &\ding{55} &\ding{51} &\ding{51} &79.50 $\pm$ 2.0 &89.61 $\pm$ 0.8 &79.90 $\pm$ 2.1 &77.54 $\pm$ 2.8 &62.91 $\pm$ 1.5 &94.13 $\pm$ 0.9 &62.58 $\pm$ 3.8 &99.20 $\pm$ 1.3 &\textit{79.38 $\pm$15.3} \\
\textbf{CLAMP} &\ding{51} &\ding{51} &\ding{51} &\ding{51} &\ding{51} &\textbf{84.98 $\pm$ 1.3} &\textbf{89.63 $\pm$ 1.0} &\textbf{84.29 $\pm$ 2.9} &\textbf{80.39 $\pm$ 2.0} &\textbf{61.37 $\pm$ 1.3} &\textbf{94.85 $\pm$ 0.3} &\textbf{59.08 $\pm$ 4.2} &\textbf{98.47 $\pm$ 2.2} &\textit{\textbf{79.74 $\pm$ 16.5}} \\
\bottomrule
\end{tabular}}

    \begin{tablenotes}
      \scriptsize
      \item PA: process adaptation, PL: pesudo label, Meta: assessor, R$_{1}$: previous source tasks memory replay, R$_{2}$: previous target tasks memory replay.
    \end{tablenotes}
\end{sidewaystable}

\subsection{Numerical Results}
Table \ref{tab: results-mn-us} \& \ref{tab: results-office31} report numerical results of consolidated algorithms in the digit recognition problem, the VisDA problem and the office-31 problem separately. 
CLAMP's superiority is evident in the digit recognition problem, where it beats other algorithms with large margins, at least $41\%$. 
This finding demonstrates that the problem of cross-domain continual learning is extremely challenging and unsolvable with classic single-domain CL algorithms. Note that the digit recognition problem is considered as a toys CL problem under a single domain configuration. This fact also substantiates the importance of domain alignment step to dampen the gap across two domains while protecting against the CF problem as exemplified by CLAMP. 
CLAMP's performance, surpassing even joint training, which was jointly trained on all tasks of the source process, attests to its efficacy in domain alignment. On the other hand, CLAMP demonstrates its advantage in the VisDA problem outperforming all other methods with significant margin except the joint training serving as the upper bound. The case of office 31 presents similar finding where CLAMP beats other methods with significant margins.

\begin{figure}
    \centering
    \includegraphics[width=\textwidth]{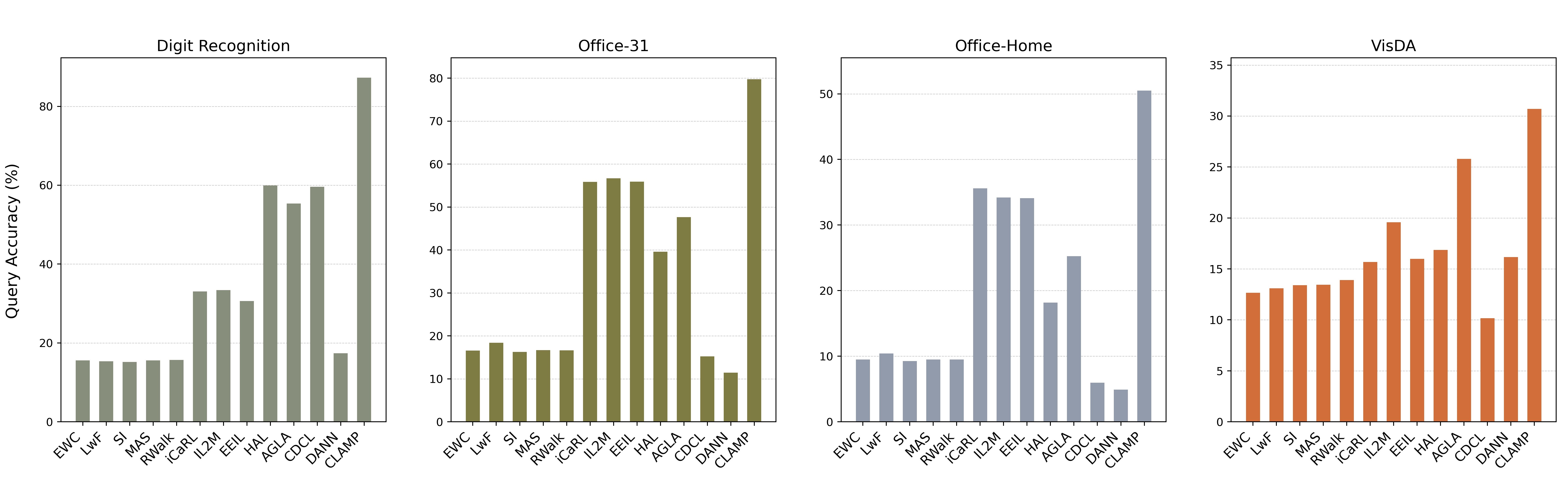}
    \caption{\textcolor{black}{Average query accuracy comparison across different baselines for four benchmarks: digit recognition, Office-31, Office-Home and VisDA}}
    \label{fig:overall-compare}
\end{figure}

A similar pattern is observed in the Office-Home problem, where CLAMP consistently outperforms the performance of other algorithms by at least $10\%$ for each experiment and approximately $22\%$ across different tasks on average. Moreover, in the more challenging Office-Home dataset, CLAMP is also superior to other algorithms by a minimum of $14\%$ across 13 tasks, as exhibited in Table \ref{tab: results-home}. Analyzing the Office-31 and Office-Home experiments, we hypothesize that significant domain discrepancies in two specific pairings of Office-31, A$\leftrightarrow$D and A$\leftrightarrow$W, and across most tasks in Office-Home, given the overall poor performance of other methods. In such challenging conditions, CLAMP demonstrates its robustness by achieving a notable margin of superiority. The problem of cross-domain continual learning is difficult to tackle because of the presence of large domain discrepancies and the CF issue. This issue cannot be sufficiently overcome with single-process continual learning algorithms or traditional domain adaptation algorithms alone. \textcolor{black}{Fig. \ref{fig:overall-compare} depicts the difference across various methods for four benchmarks: digit recognition, Office-31, Office-Home and VisDA. CLAMP surpasses competing methods in all evaluation scenarios with notable gaps.}

\subsection{Ablation Studies}
Ablation study is performed to understand the impact of each learning component where numerical results are reported in Table \ref{tab: ablation}. It is perceived that the naive training on the source domain leads to the poorest results since the catastrophic forgetting problem and the domain discrepancies between the source and target domains occur. An additional adversarial domain adaptation step as shown in the PA does not help to improve numerical results. Baseline 1 integrating the pseudo labelling strategy of the target domain and the adversarial domain adaptation method improves numerical results of the naive approach. The deployment of assessors to produce meta-weights as shown in Baseline 2 also boosts numerical results compared to Baseline 1. The use of episodic memory addresses the catastrophic forgetting problem as shown in Baseline 3 but still does not deliver good performances because of the absence of domain adaptations. Baseline 4 improves numerical results of baseline 3 but suffers from the catastrophic forgetting problem of the target domain because of the absence of episodic memory of the target domain. The vital role of assessors is confirmed in Baseline 5 where their absences deteriorate numerical results. These findings substantiate positive impacts of CLAMP's learning modules. 

\begin{table}[!htp]\centering
\caption{Accuracy (\%) of CLAMP with different exemplar sizes per class on Office-31}\label{tab: abl-buffer}
\scriptsize
\begin{tabular}{ccccc}\toprule
Exemplar per class &1 &3 &5 &\textbf{10(ours)} \\
\midrule
A$\rightarrow$D &62.97 &79.64 &81.3 &\textbf{84.29} \\
A$\rightarrow$W &61.85 &72.73 &75.29 &\textbf{80.39} \\
D$\rightarrow$A &46.53 &58.85 &60.6 &\textbf{61.37} \\
D$\rightarrow$W &88.74 &92.19 &93.74 &\textbf{94.85} \\
W$\rightarrow$A &45.41 &57.42 &60.78 &\textbf{59.08} \\
W$\rightarrow$D &92.96 &97.12 &98.21 &\textbf{98.47} \\
\textit{Avg.} &\textit{66.41} &\textit{76.33} &\textit{78.32} &\textit{\textbf{79.74}} \\
\bottomrule
\end{tabular}
\end{table}



\subsection{Memory Analysis}
\textcolor{black}{Table \ref{tab: abl-buffer} tabulates numerical results of CLAMP under different memory sizes per class for both target domain and source domain. It is perceived that the performance of CLAMP degrades when decreasing the number of exemplar per class in the memory. This situation occurs because reduction of memory sizes leads to the extreme class imbalanced problem between the current task and the episodic memory leading to the catastrophic forgetting problem. However, the performance gaps gradually decreases with the increase of the memory size. Note that CLAMP utilizes 10 exemplars per class in the main numerical results showing modest memory usages.}

\subsection{Effect of Inner Loop Steps in Assessor Network Update}
To validate the approximation of the assessor's single-step update, we investigate the influence of different step sizes on the final average accuracy and corresponding training time. The results are summarized in Table \ref{tab: meta-inner epoch}, where steps indicate how many times the inner iteration updates the assessor network parameters before determining the three meta-weights. It turns out that larger inner iteration steps do not significantly improve performance and impose more processing time, which is similar to \textcolor{black}{Li et al.} \cite{Li2023MetaReweightedRF}. The increase of inner-iteration of the assessor network update results in the unstable impact on the optimization of the main network parameters. Based on the results, the single meta update step approaches satisfactory performance and incurs much less execution time. Thus, we apply the single-step update in the inner loop.

\begin{table}[!]\centering
\caption{Results of using different inner-iteration steps to update assessor network parameters on MNIST(MN)$\leftrightarrow$USPS(US).}\label{tab: meta-inner epoch}
\scriptsize
\resizebox{\linewidth}{!}{
\begin{tabular}{c|r|ccccc}\toprule
& Steps &1(ours) &2 &3 &4 &5 \\\cmidrule{2-7}
\multirow{1}{*}{MN$\rightarrow$US} & Accuracy (\%) &84.98 $\pm$ 1.3 &84.88 $\pm$ 2.1 &84.26 $\pm$ 2.1 &83.83 $\pm$ 1.3 &82.28 $\pm$ 2.3 \\\cmidrule{2-7}
&Training Time (sec.) &2267 $\pm$ 413.8 &3802 $\pm$ 110.3 &4920 $\pm$ 81.2 &6132 $\pm$ 89.8 &7404 $\pm$ 161.4 \\\midrule
& Steps &1(ours) &2 &3 &4 &5 \\\cmidrule{2-7}
\multirow{1}{*}{US$\rightarrow$MN} & Accuracy (\%) &89.63 $\pm$ 1.0 &89.40 $\pm$ 1.0 &88.84 $\pm$ 1.9 &89.49 $\pm$ 0.9 &89.57 $\pm$ 0.7 \\\cmidrule{2-7}
&Training Time (sec.) &597 $\pm$ 2.4 &1030 $\pm$ 23.3 &1153 $\pm$ 15.4 &1270 $\pm$ 25.9 &1381 $\pm$ 20.5 \\
\bottomrule
\end{tabular}}
\end{table}

\begin{figure*}
     \centering
     \begin{subfigure}[t]{0.16\textwidth}
         \centering
         \includegraphics[width=\textwidth]{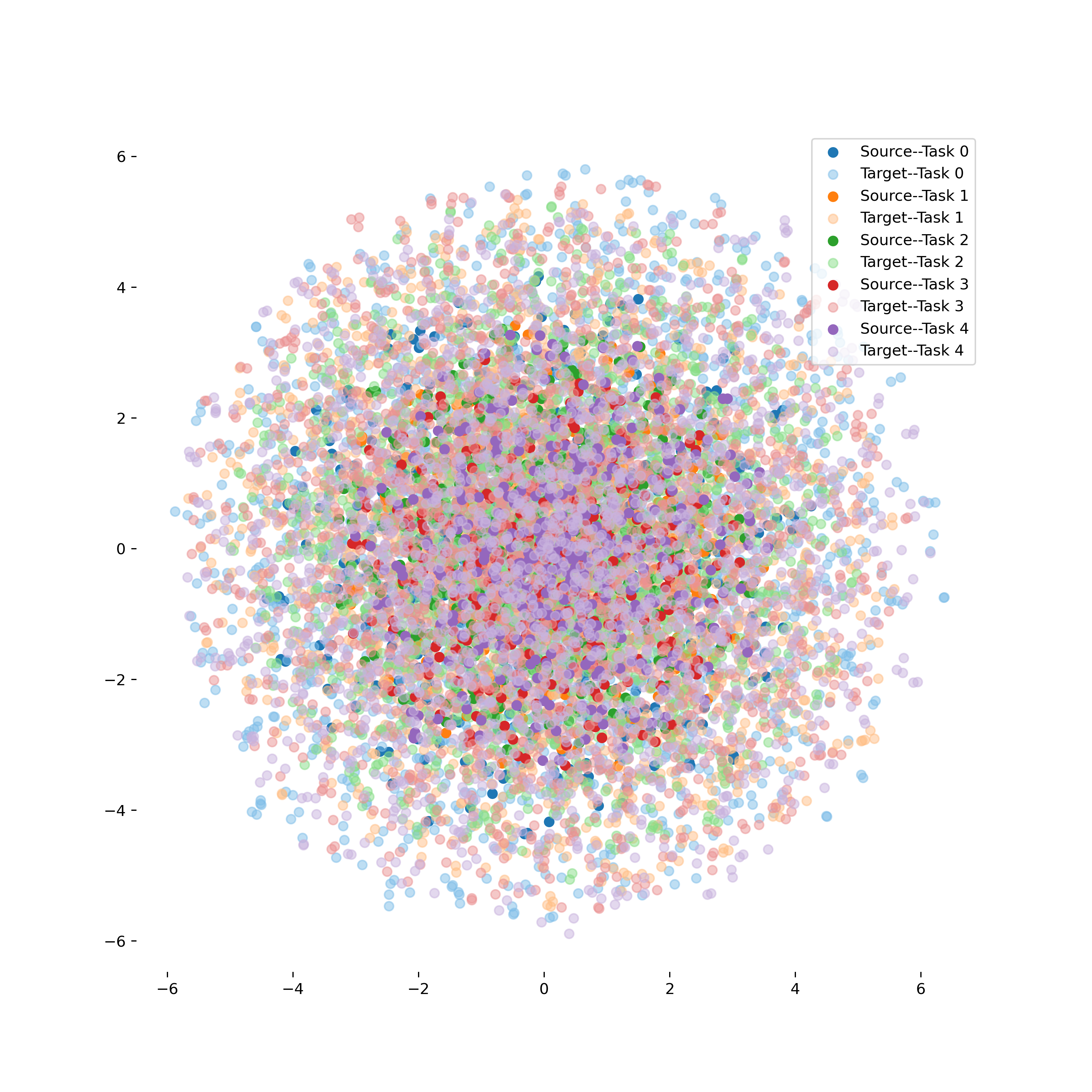}
         \caption{Before alignment}
         \label{fig:tsne0}
     \end{subfigure}
     \hfill
     \vrule
     \begin{subfigure}[t]{0.16\textwidth}
         \centering
         \includegraphics[width=\textwidth]{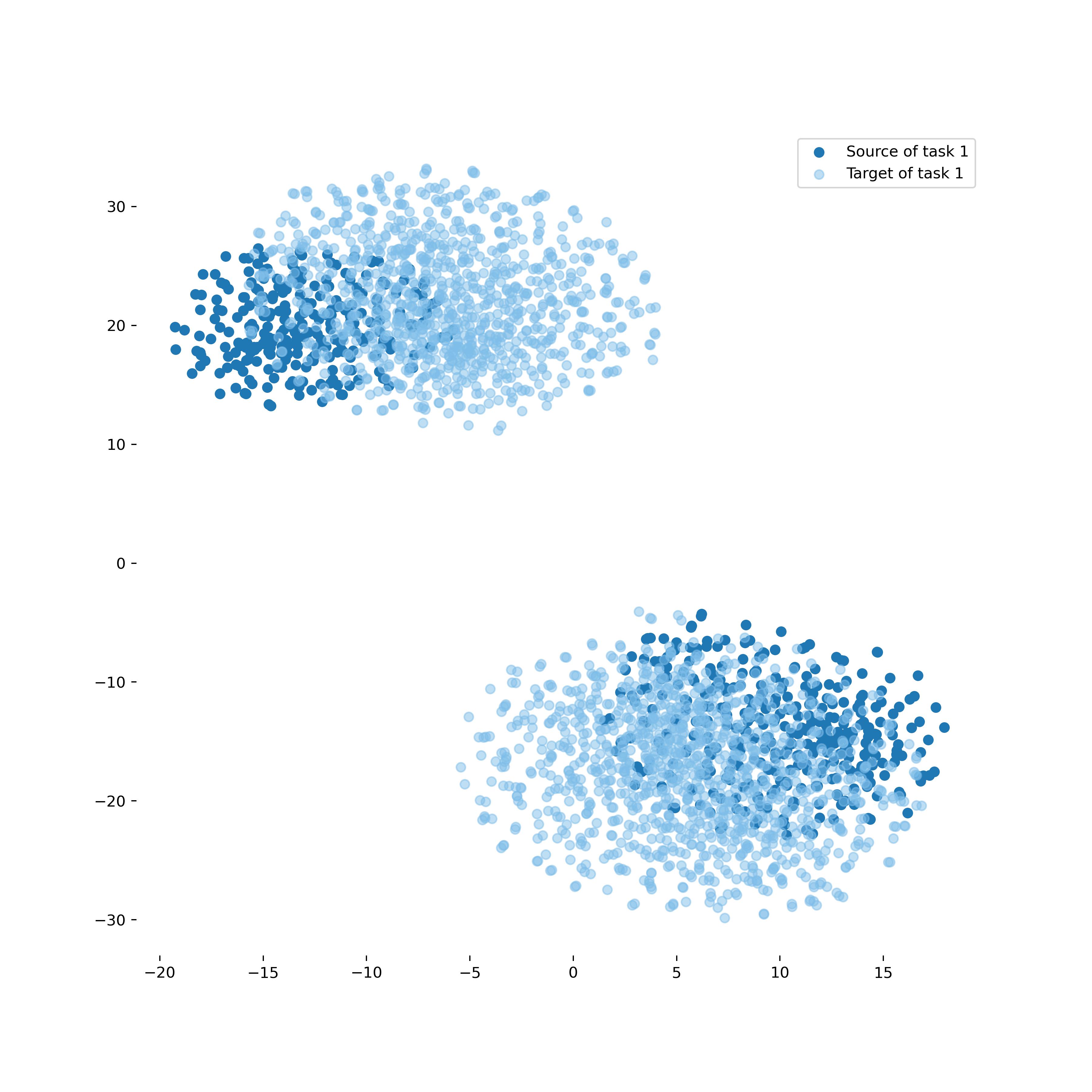}
         \caption{Task \textbf{\textcolor{blue}{1}}}
         \label{fig:tsne1}
     \end{subfigure}
     \begin{subfigure}[t]{0.16\textwidth}
         \centering
         \includegraphics[width=\textwidth]{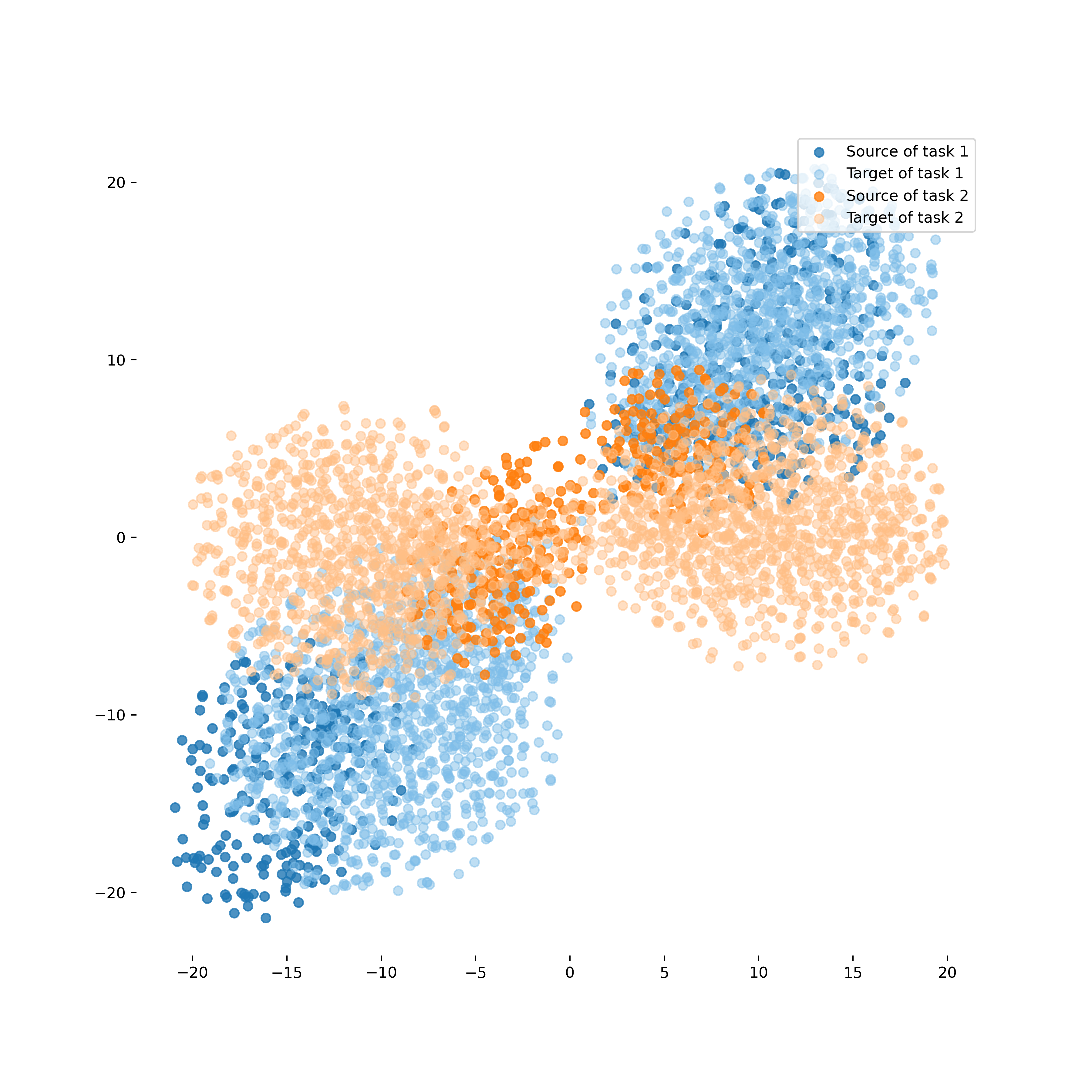}
         \caption{Task 
         \textbf{\textcolor{orange}{2}}}
         \label{fig:tsne2}
     \end{subfigure}
     \hfill
     \begin{subfigure}[t]{0.16\textwidth}
         \centering
         \includegraphics[width=\textwidth]{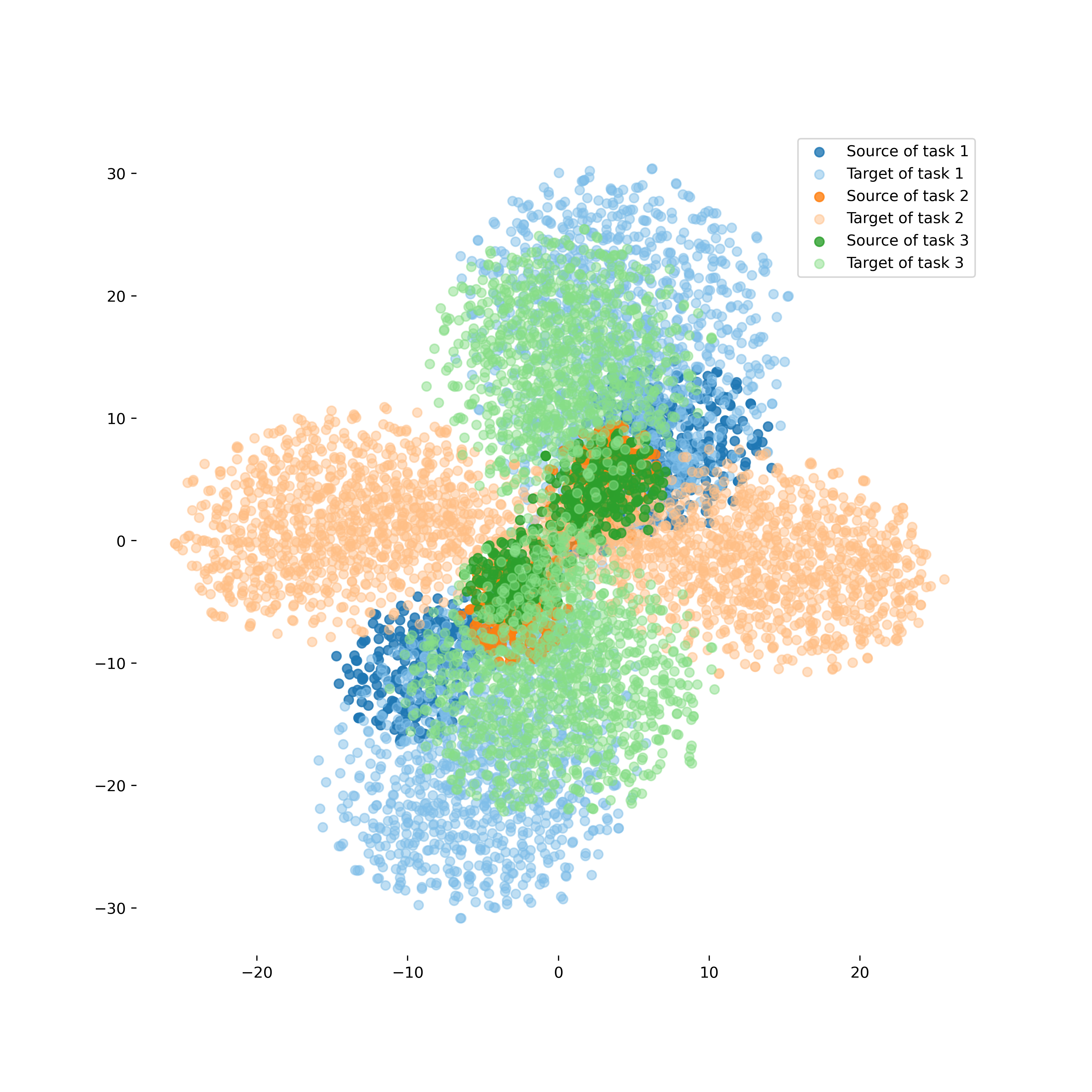}
         \caption{Task 
         \textbf{\textcolor{green}{3}}}
         \label{fig:tsne3}
     \end{subfigure}
     \hfill
     \begin{subfigure}[t]{0.16\textwidth}
         \centering
         \includegraphics[width=\textwidth]{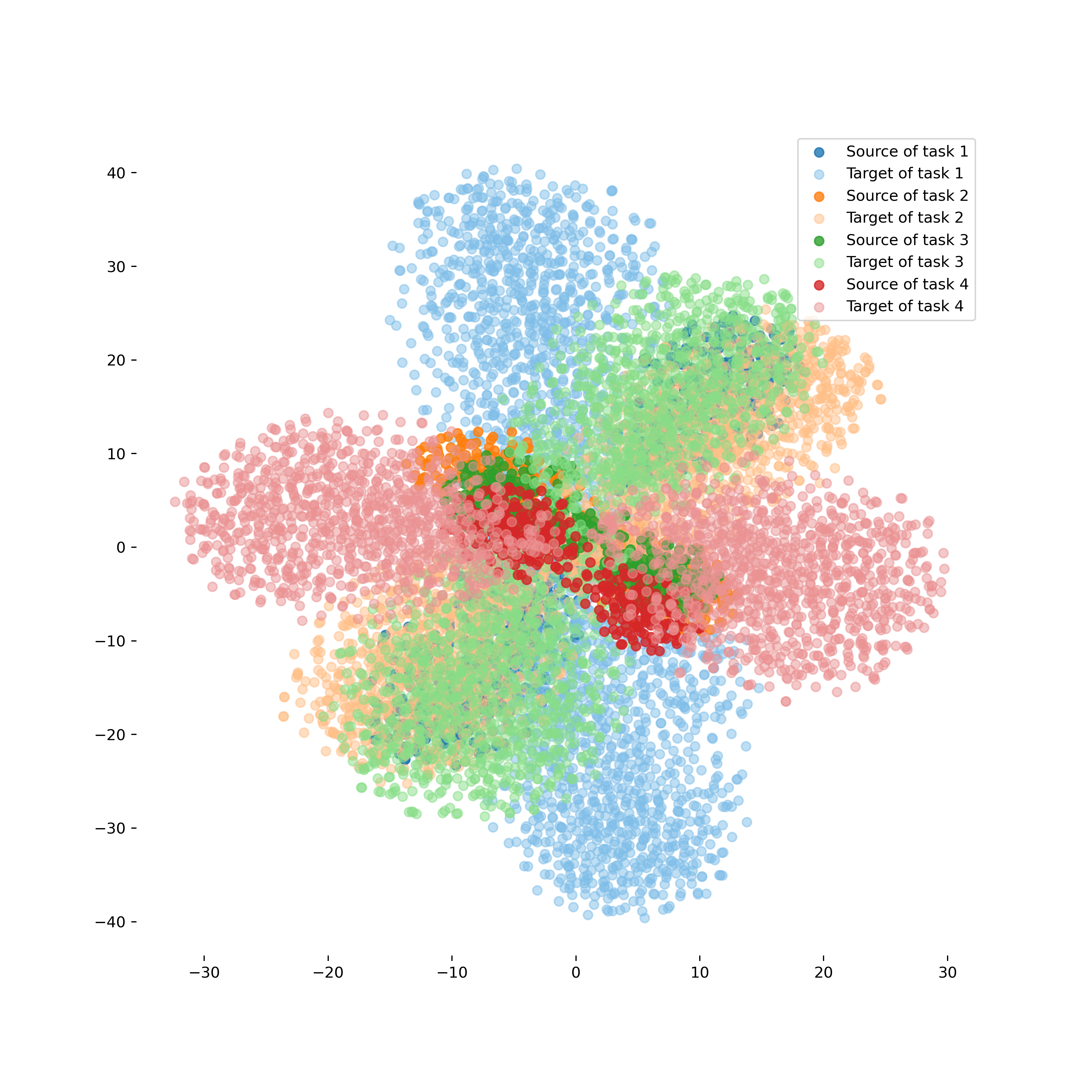}
         \caption{Task 
         \textbf{\textcolor{red}{4}}}
         \label{fig:tsne4}
     \end{subfigure}
     \hfill
     \begin{subfigure}[t]{0.16\textwidth}
         \centering
         \includegraphics[width=\textwidth]{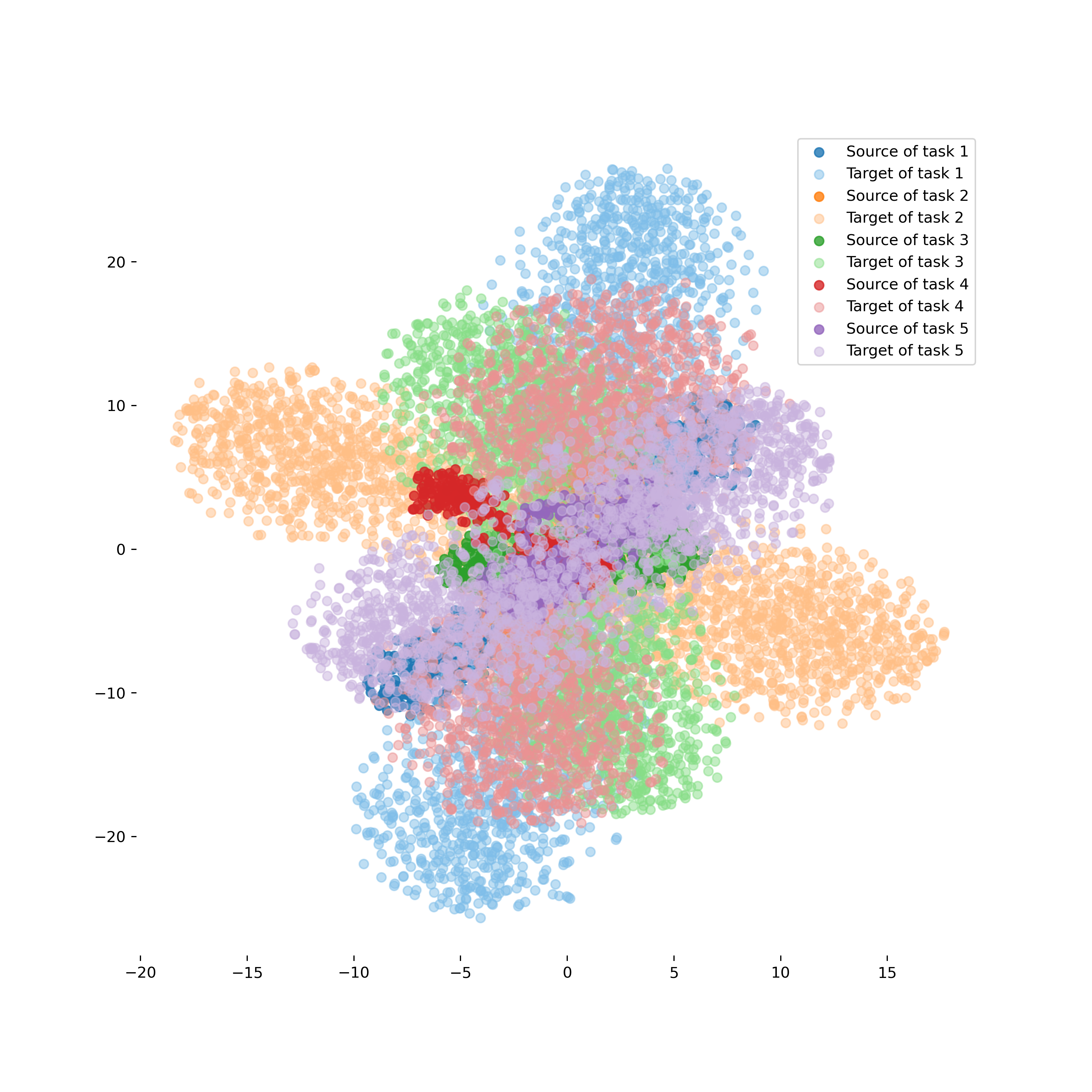}
         \caption{Task 
         \textbf{\textcolor{violet}{5}}}
         \label{fig:tsne5}
     \end{subfigure}
        \caption{t-SNE plots of the USPS (in light colors) $\rightarrow$ MNIST (in dark colors) case}
        \label{fig:tsne}
\end{figure*}

\subsection{T-SNE Analysis}
The t-sne analysis is performed to evaluate the embedding qualities of CLAMP using the USPS$\rightarrow$MNIST case. Fig. \ref{fig:tsne} visualizes the t-SNE plot across each task. Before the training process begins, there do not exist any clear patterns indicating confused predictions. The advantage of CLAMP is clearly seen after the training process where the source domain and target domain are successfully aligned and show remarkable cluster structures showing confident predictions. In addition, different classes are projected differently in the feature space signifying the robustness against catastrophic forgetting problem. 

\begin{table}[htbp]
  \small
  \centering
  \caption{Average query accuracy comparison of learning rate impact on base learner (\(l_b\)) and assessors (\(l_a\)) performance in MNIST$\rightarrow$USPS experiment, highlighting the optimal method results in bold.}
  \begin{tabular}{cllll}
    \hline
    \(l_a\) & 0.1 & 0.01 & 0.001 & 0.0001 \\
    \(l_b = 0.001\) & 38.13 \(\pm\) 2.9 & 78.05 \(\pm\) 8.0 & 82.34 \(\pm\) 2.5 & \textbf{84.98 \(\pm\) 1.3} \\
    \hline
    & & \\ 
    \hline
    \(l_b\) & 0.1 & 0.01 & 0.001 & 0.0001 \\
    \(l_a = 0.0001\) & 63.32 \(\pm\) 5.8 & 79.88 \(\pm\) 1.9 & \textbf{84.98 \(\pm\) 1.3} & 53.98 \(\pm\) 4.4 \\
    \hline
  \end{tabular}%
  \label{tab: lr-digits}%
\end{table}%

\subsection{Sensitivity Analysis}
\textcolor{black}{The influence of learning rate for base learner (\(l_b\)) and assessors (\(l_a\)) on the performance of CLAMP is elaborated in this section. Table \ref{tab: lr-digits} shows learning rate analysis results on the digit recognition experiment. A careful examination of Table \ref{tab: lr-digits} reveals significant insights into the optimal learning rate settings for both base learners and assessors in the context of MNIST$\rightarrow$USPS. It is evident that the base learner's performance improves with a decrease in the learning rate, with the most notable improvement observed when \(l_b\) is set to 0.001. The accuracy peaks at 84.98\% (\(\pm\) 1.3), indicating a higher stability and efficiency at this learning rate. To strive a balance between performance and training time, we select 0.0001 as the assessor's learning rate. Conversely, for assessors, the learning rate of 0.0001 with \(l_b\) fixed at 0.001 demonstrates the best performance, achieving an accuracy of 84.98\% (\(\pm\) 1.3). This suggests that a lower learning rate for assessors is conducive to achieving optimal performance.
Additionally, the results indicate a clear trend of diminishing returns at extremely low or high learning rates, as seen with the other combinations of \(l_a\) and \(l_b\), which yield lesser accuracy and higher standard deviations. The highlighted results in bold in the table further emphasize the optimal learning rate combinations for maximizing accuracy in digit recognition tasks within the CLAMP framework. This analysis aligns well with the fact that CLAMP forms a bi-level optimization approach between the assessor in the inner loop and the base learner in the outer loop. That is, the assessor should be updated in the slower pace than that the base learner to assure model's convergences.
 Table \ref{tab:pseudolabel} reports our numerical results when varying the pseudo label thresholds. It is seen that a too low threshold leads to performance degradation by about $1\%$ because it leads to too many noisy pseudo labels to be involved in the training process. On the other hands, a too high threshold also results in performance deterioration because too few pseudo labels are induced. Generally speaking, CLAMP is not too sensitive to the pseudo label threshold because of the presence of a meta-training loop in the target domain contributing to reject noisy pseudo labels. The pseudo label threshold is fixed at 0.85 in all our experiments. 
}

\begin{figure}[ht]
    \centering
    
    \begin{subfigure}[b]{0.9\textwidth}
        \centering
        \includegraphics[width=\textwidth]{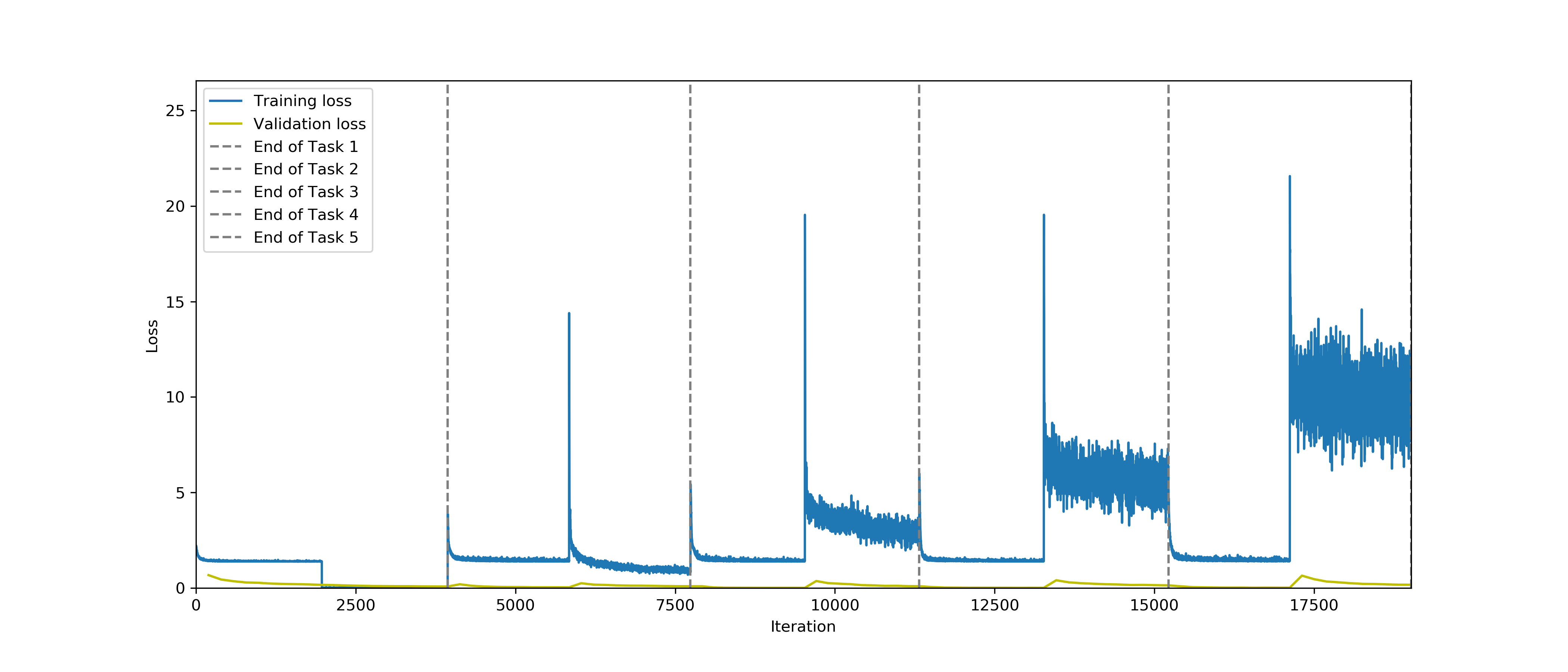}
        \caption{MNIST$\rightarrow$USPS}
        \label{fig:sub1}
    \end{subfigure}
    
    \begin{subfigure}[b]{0.9\textwidth}
        \centering
        \includegraphics[width=\textwidth]{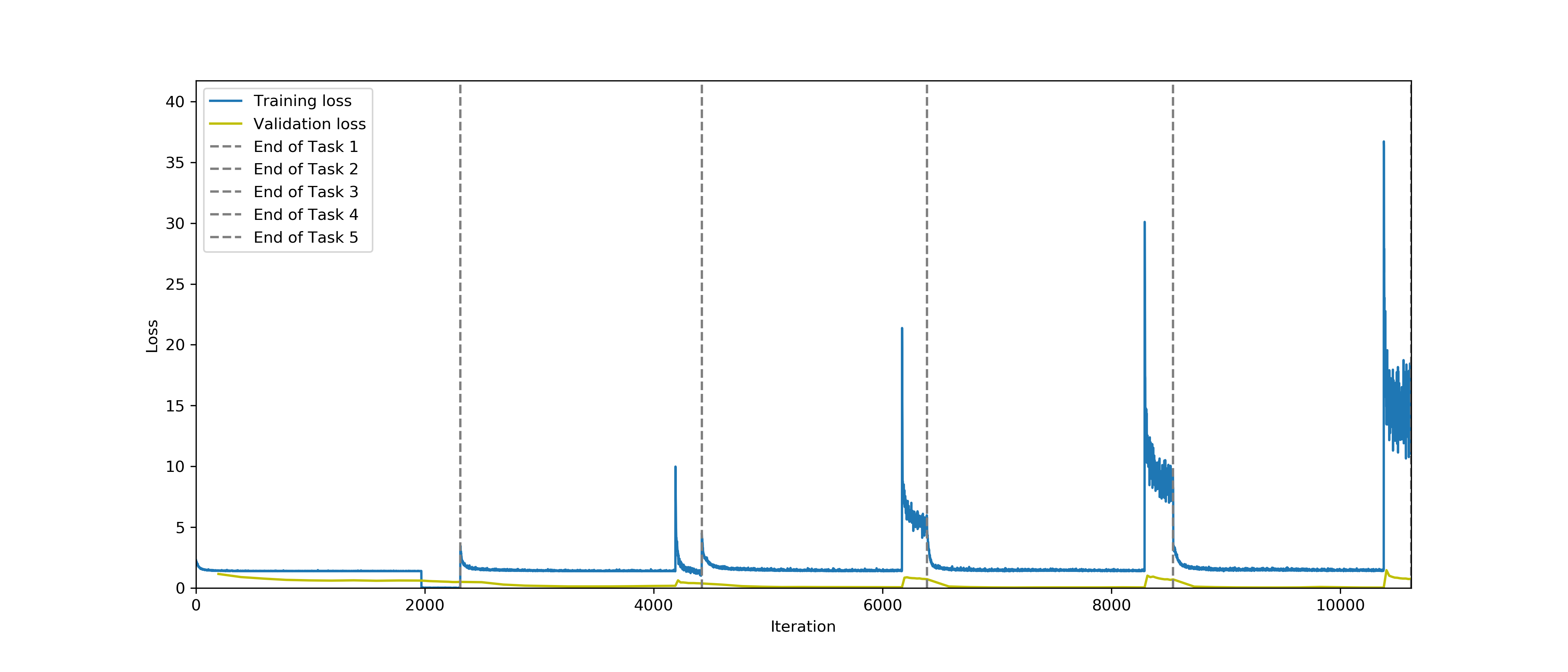}
        \caption{USPS$\rightarrow$MNIST}
        \label{fig:sub2}
    \end{subfigure}

    \caption{CLAMP loss trajectories in digits recognition: (a) MNIST$\rightarrow$USPS, (b) USPS$\rightarrow$MNIST. It illustrates the trajectories of training loss and validation loss over the entire training process in the cross-domain continual learning scenario. }
    \label{fig:loss_digit}
\end{figure}

\begin{table}[ht]
  \centering
  \caption{Sensitivity analysis with varying threshold ($\gamma$) for MNIST$\leftrightarrow$USPS.}
    \begin{tabular}{ccc}
    \toprule
    $\gamma$ & MNIST$\rightarrow$USPS & USPS$\rightarrow$MNIST \\
    \midrule
    0.7   & 83.24 \(\pm\) 1.57 & 88.80 \(\pm\) 0.94 \\
    0.75  & 83.58 \(\pm\) 0.97 & 89.65 \(\pm\) 0.91 \\
    0.8   & 84.56 \(\pm\) 1.21 & 89.31 \(\pm\) 1.00 \\
    0.85(ours)  & 84.98 $\pm$ 1.3 & 89.63 $\pm$ 1.0 \\
    0.9   & 84.07 \(\pm\) 1.26 & 89.29 \(\pm\) 1.15 \\
    \bottomrule
    \end{tabular}%
  \label{tab:pseudolabel}%
\end{table}%

\subsection{Analysis of Model Generalization}
This section is meant to analyze the generalization performance of CLAMP where the trace of training losses and validation losses are portrayed in Fig. \ref{fig:loss_digit} for the digit recognition problem. It is evident that CLAMP does not exhibit any over-fitting signs where the validation losses are well below the training losses during the cross-domain continual learning process. That is, CLAMP generalizes well to unseen samples of the validation set.

\begin{figure}
    \centering
    \includegraphics[width=0.95\textwidth]{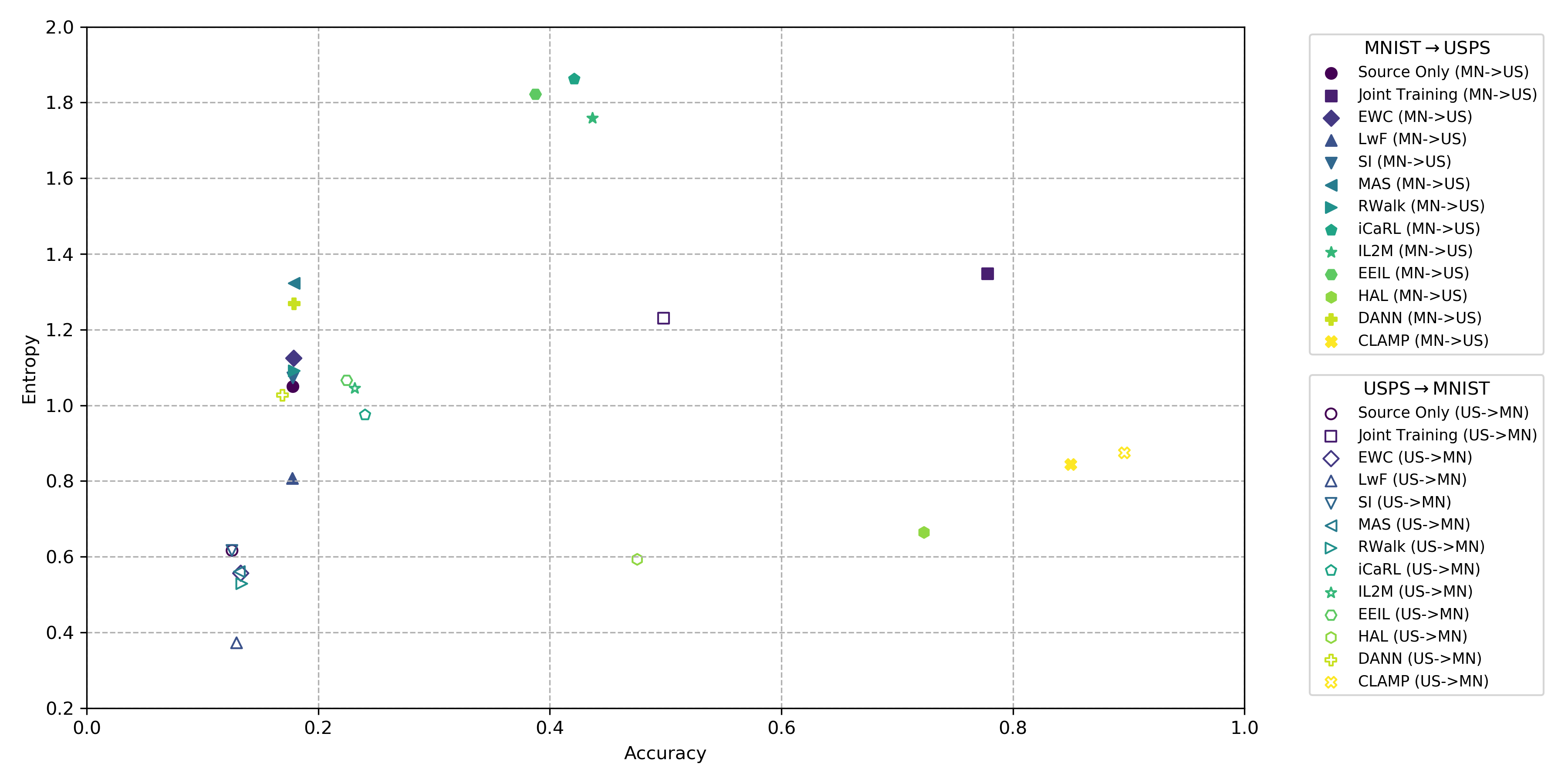}
    \caption{Comparative Analysis of Model Performance on digit recognition tasks: Accuracy vs. Entropy (predictive uncertainty) }
    \label{fig:entropy-acc}
\end{figure}

\subsection{Analysis of Model's Uncertainties}
Using the entropy formulation $H(X) = -\sum_{i=1}^{n} P(x_i) \log_b P(x_i)$, we analyze the model's performance in relation to entropy. Fig. \ref{fig:entropy-acc} demonstrates that for digit recognition tasks on MNIST and USPS datasets, CLAMP achieves the highest average query accuracy among all compared methods. This plot shows the correlation between model uncertainty and entropy, in which lower entropy indicate greater certainty in model predictions. Specifically, CLAMP achieves an entropy of 0.8445 for MNIST$\rightarrow$USPS and 0.8745 for USPS$\rightarrow$MNIST. While these values are not the lowest, they are comparatively low, denoting a respectable level of certainty in its predictions. Note that HAL displays the lowest entropy, suggesting it is more certain in its predictions than CLAMP for the MNIST$\rightarrow$USPS task. The figure clearly shows that CLAMP maintains high accuracy while achieving relatively lower entropy compared to other methods. This visually confirms that CLAMP is generally more accurate and certain in its predictions than most other approaches.

\begin{table}[htbp]
\small
  \centering
  \caption{T-test results (significance level: 0.05) comparing CLAMP with benchmark algorithms on digit recognition and VisDA.}
    \begin{tabular}{crrr}
    \toprule
    Source & Split MNIST & Split USPS & VisDA \\
    Target & Split USPS & Split MNIST &  \\
    \midrule
    SourceOnly & 2288.18 & 5729.31 & 230.28 \\
    Joint Training & 139.53 & 3262.18 & \textcolor[rgb]{ 1,  0,  0}{-194.73} \\
    EWC   & 2253.13 & 3706.91 & 127.99 \\
    LwF   & 2214.7 & 4254  & 227.05 \\
    SI    & 2288.18 & 5451.09 & 174.56 \\
    MAS   & 2251.79 & 3418.95 & 233.8 \\
    RWalk & 2251.79 & 3865.52 & 189.26 \\
    iCaRL & 681.58 & 4875.27 & 169.44 \\
    IL2M  & 521.78 & 3095.82 & 136.14 \\
    EEIL  & 633.44 & 3904.18 & 161.29 \\
    HAL   & 297.1 & 1677.84 & 160.29 \\
    DANN  & 1900.91 & 2232.83 & 225.12 \\
    \bottomrule
    \end{tabular}%
  \label{tab:t-test-1}%
\end{table}%

\begin{table}[htbp]
  \centering
  \caption{T-test results (significance level: 0.05) comparing CLAMP with benchmark algorithms on the Office-31.}
    \begin{tabular}{crrrrrr}
    \hline
          & A$\rightarrow$D & A$\rightarrow$W & D$\rightarrow$A & D$\rightarrow$W & W$\rightarrow$A & W$\rightarrow$D \\
    \hline
    Source Only & 370.8 & 720.95 & 996.98 & 748.74 & 454.21 & 444.97 \\
    Joint Training & 26.46 & 249.64 & 321.51 & 135.1 & 112.27 & 56.37 \\
    EWC   & 118.65 & 736.47 & 933.4 & 906.84 & 436.49 & 436.81 \\
    LwF   & 90.99 & 634.62 & 1044.21 & 431.9 & 474.94 & 488.5 \\
    SI    & 115.26 & 671.72 & 938.53 & 725.23 & 454.21 & 520.11 \\
    MAS   & 110.62 & 738.62 & 933.4 & 906.84 & 437.67 & 436.81 \\
    RWalk & 111.91 & 736.47 & 934.46 & 906.84 & 436.49 & 439.03 \\
    iCaRL & 37.97 & 445.56 & 223.79 & 117.15 & 194.44 & 77.49 \\
    IL2M  & 37.71 & 271.89 & 204.68 & 193.53 & 190.23 & 98.52 \\
    EEIL  & 39.86 & 253.27 & 217.18 & 182.32 & 187.9 & 90.2 \\
    HAL   & 54.15 & 307.76 & 886.62 & 320.8 & 294.72 & 308.1 \\
    DANN  & 199.39 & 513.12 & 1848.2 & 649.62 & 479.08 & 211.83 \\

    \hline
    \end{tabular}%
  \label{tab:t-test-office31}%
\end{table}%

\begin{table}[htbp]
  \centering
  \caption{T-test results (significance level: 0.05) comparing CLAMP with benchmark algorithms on the Office-Home.}
\resizebox{\columnwidth}{!}{
    \begin{tabular}{lrrrrrrrrrrrr}
    \toprule
    Source & Art & Art & Art & Clipart & Clipart & Clipart & Product & Product & Product & Real World & Real World & Real World \\
    Target & Clipart & Product & Real World & Art & Product & Real World & Art & Clipart & Real World & Art & Clipart & Product \\
    \hline
    Source Only & 386.55 & 744.62 & 485.27 & 165.7 & 530.25 & 805.68 & 158.54 & 485.95 & 503.46 & 333.11 & 368.49 & 592.74 \\
    Joint Training & 72.93 & 75.25 & \textcolor[rgb]{ 1,  0,  0}{-33.75} & \textcolor[rgb]{ 1,  0,  0}{-17.67} & 5.83  & \textcolor[rgb]{ 1,  0,  0}{-79.18} & \textcolor[rgb]{ 1,  0,  0}{-14.65} & 49.17 & -89.53 & \textcolor[rgb]{ 1,  0,  0}{-34.06} & 54.19 & \textcolor[rgb]{ 1,  0,  0}{-96.09} \\
    EWC   & 409.81 & 607.53 & 472.45 & 278.8 & 560.52 & 802.09 & 157.69 & 595.32 & 486.37 & 285.55 & 360.27 & 595.53 \\
    LwF   & 322.08 & 633.7 & 477.91 & 217.5 & 520.18 & 614.78 & 150.86 & 452.5 & 360.65 & 302.94 & 366.2 & 540.61 \\
    SI    & 390   & 782.58 & 471.27 & 264.89 & 534.08 & 687.13 & 163.12 & 443.98 & 455.66 & 300.04 & 359.11 & 581.4 \\
    MAS   & 328.9 & 775.84 & 395.31 & 268.32 & 490.09 & 809.47 & 146.94 & 513.54 & 473.06 & 240.07 & 376.67 & 585.63 \\
    RWalk & 375.52 & 638.49 & 460.16 & 239.53 & 560.93 & 929.87 & 152.4 & 512.03 & 480.31 & 264.24 & 388.02 & 583.75 \\
    iCaRL & 230.78 & 228.28 & 77.58 & 178.49 & 165.11 & 229.26 & 45    & 202.08 & 70.05 & 115.36 & 215.5 & 116.13 \\
    IL2M  & 253.48 & 193.51 & 116.19 & 202.93 & 240.73 & 235.08 & 50.93 & 337   & 81.69 & 120.61 & 186.19 & 116.87 \\
    EEIL  & 225.93 & 182.76 & 112.23 & 203.32 & 265.64 & 266.67 & 57.3  & 318.08 & 95.96 & 125.86 & 216.62 & 123.24 \\
    HAL   & 406.34 & 527.48 & 343.8 & 230.47 & 326.26 & 350.79 & 109.06 & 317.69 & 264.7 & 186.56 & 370.17 & 337.44 \\
    DANN  & 683.46 & 842.91 & 691.49 & 212.5 & 425.24 & 481.22 & 175.9 & 400.27 & 496.84 & 348.51 & 611.49 & 624.01 \\
    \bottomrule
    \end{tabular}%
    }
  \label{tab:t-test-home}%
\end{table}%

\subsection{Statistical Test}
Table \ref{tab:t-test-1}, Table \ref{tab:t-test-office31} and Table \ref{tab:t-test-home} present the results of T-tests comparing CLAMP with benchmark algorithms on all experiments. The T-values in the table provide statistical evidence of significant differences in performance between CLAMP and the other algorithms, with notably high t-values. It is important to note that the p-values, although they are not presented in the table, are extremely small, all well below 0.0001, indicating highly significant differences. Overall, the T-values provide strong statistical evidence that CLAMP performs significantly better than the benchmark algorithms in most cross-domain continual learning tasks. This superiority is evident in various cases, except when compared with Joint Training on the VisDA dataset and in six experiments of Office-Home, as indicated by red markings. This exception may be attributed to the elevated discrepancies between source and target domains in these specific experiments. Note that the joint training presents the case of upper bound for continual learning. These findings underscore the effectiveness of CLAMP in addressing cross-domain continual learning problem.

\begin{table}[htp]
    \centering
    \begin{tabular}{c|cc}
    \toprule
    Method & clipart $\rightarrow$ sketch & sketch $\rightarrow$ clipart \\
    \midrule
    CDCL   & 3.49   & 2.17 \\
    \textbf{CLAMP}  & \textbf{12.87}  & \textbf{13.67} \\
    \bottomrule
    \end{tabular}
    \\[1pt]
    \textit{The results for CDCL are taken from the original paper.}
    \caption{Average query accuracy (\%) of all learned tasks on DomainNet (clipart and sketch).}\label{tab: results-domainnet}
    \label{tab: DomainNet}
\end{table}

\subsection{Experiments with DomainNet}
\textcolor{black}{Additional experiments are performed using a highly challenging problem for domain adaptation, DomainNet \cite{Peng2018MomentMF}. The DomainNet problem presents 345 classes and 6 domains where the continual learning problem is constructed by splitting 345 classes into 15 tasks with 23 classes each. CLAMP is compared with CDCL \cite{VinciusdeCarvalho2024TowardsCC} representing prior arts in the cross-domain continual learning. 2 cross domain experiments, namely $sketch\leftrightarrow clipart$, are executed. Table \ref{tab: DomainNet} reports our numerical results. It is perceived that CLAMP outperforms CDCL with over $10\%$ margins across 2 cross-domain cases, $sketch\leftrightarrow clipart$. Note that the DomainNet problem is highly challenging because of the large number of classes and significant discrepancies across the two domains, i.e., CDCL only obtains $~2-3\%$ accuracy. These results further confirm the robustness of our finding in respect to the advantage of CLAMP for cross-domain continual learning problems.}

\subsection{Complexity Analysis}
\textcolor{black}{In this part, we analyze the complexity of CLAMP. Let $N_k^S$ and $N_k^T$ denote the number of samples for the $k$-th task in the source and target domains, respectively, where $k \in \{1,\ldots,K\}$. Let $M_{k-1}^S$ and $M_{k-1}^T$ represent the episodic memory for the source and target domains of the previous task, starting from $k=2$. $E$ is the number of epochs, and $E_{inner}$ and $E_{outer}$ are the number of inner and outer epochs in the meta learning process, respectively. At $k$-th task, the complexity of unsupervised process adaption (Stage 1 in Algorithm \ref{alg:clamp}) is $O\left (\min(N_k^S+M_{k-1}^S,N_k^T+M_{k-1}^T)\right)$, the complexity of Stage 2 is $O\left(E_{outer} \times \left(N_{trans} \times (N_k^S+M_{k-1}^S)+E_{inner} \times N_{trans} \times (N_k^S+M_{k-1}^S)\right)\right)$ and that of Stage 3 is $O\left(E_{outer} \times \left(N_{trans} \times (N_k^T+M_{k-1}^T)+E_{inner} \times N_{trans} \times (N_k^T+M_{k-1}^T)\right)\right)$. $N_{trans}$ represents the number of transformations, which is three in CLAMP. Given $E_{inner}$ and $E_{outer}$ are both 1 (Table \ref{hyper-params}) and assuming $min(N_k^S+M_{k-1}^S,N_k^T+M_{k-1}^T)$ is represented as $N_k+M_k$. Thus, the complexity of each stage simplifies to $O\left((N_k+M_k)\right)$ for Stage 1, $O(6(N_k^S+M_{k-1}^S))$ for Stage 2 and $O(6(N_k^T+M_{k-1}^T))$ for Stage 3. The overall complexity of CLAMP can be expressed as:
\begin{align}
    O_{\text{overall}} &= E \times \sum_{k=1}^{K}\left(O(\text{Stage 1}) + O(\text{Stage 2}) + O(\text{Stage 3})\right) \\
    &= E \times \sum_{k=1}^{K}\left(O\left(\min(N_k^S + M_{k-1}^S, N_k^T + M_{k-1}^T)\right)\right. \notag \\
    &\quad + O\left(E_{outer} \times \left(3 \times (N_k^S + M_{k-1}^S) + E_{inner} \times 3 \times (N_k^S + M_{k-1}^S)\right)\right) \notag \\
    &\quad \left. + O\left(E_{outer} \times \left(3 \times (N_k^T + M_{k-1}^T) + E_{inner} \times 3 \times (N_k^T + M_{k-1}^T)\right)\right) \right) \\
    &= E \times \sum_{k=1}^{K}\left(O\left((N_k + M_k)\right) + O(6(N_k^S + M_{k-1}^S)) + O(6(N_k^T + M_{k-1}^T))\right) \\
    &\leq \sum_{k=1}^{K}O(7E \times (N_k + M_k))
\end{align}
}
\subsection{Discussion}
\textcolor{black}{Five aspects are observed from our experiments:
\begin{itemize}
    \item The cross-domain continual learning problem is challenging where it suffers from both the CF issue and the domain shift issue. In addition, because both the source domain and the target domain characterize sequences of tasks, the CF problem occurs in both domain. 
    \item The performance of other algorithms are poor because they only handle the CF problem in the source domain ignoring the CF problem in the target domain as well as the issue of domain shift due to the absence of domain alignment procedures.
    \item CLAMP performs satisfactorily compared to other methods because it addresses the CF problem in both domain as well as the domain shift problem. The presence of adversarial domain adaptation not only dampens the gap between the source domain and the target domain, i.e., the domain alignment but also attains the class alignment because of the pseudo-labelling mechanism. In addition, the dual assessors contribute positively to balance the stability and the plasticity with generations of meta-weights steering the roles of loss functions. 
    \item The learning rates play vital roles in the meta-training configuration. We find that the inner loop should be set lower than the outer loop to assure model's convergence. 
    \item Joint training usually serving as an upper bound in the continual learning area does not perform well in the cross-domain continual learning problems because it does not have any strategies to overcome the issue of domain shift. 
    \item \textcolor{black}{We are certain that the problem of shortcut learning \cite{Geirhos2020ShortcutLI} where a model only focuses on simple features does not exist in our method because these models are given with sufficient samples in the source domain and the target domain.}
\end{itemize}}

\subsection{Limitation}
Although CLAMP delivers commendable performances in the cross-domain continual learning problems compared to the prior arts, several issues remain open:
\begin{itemize}
    \item CLAMP still utilize the episodic memories storing old data samples for the source domain and the target domain. Notwithstanding that the episodic memory is common in the continual learning, it imposes additional memory expenditures and the size of episodic memory grows as the number of tasks.
    \item CLAMP imposes high complexity because two assessor networks need to be trained simultaneously aside from the main networks. 
    \item Our setting assumes that the source domain and the target domain share the same label space which may not hold in several practical applications. This setting can be extended to the partial domain adaption, the open-set domain adaptation or the universal domain adaptation. 
    \item CLAMP requires an access to the source domain samples, which might not be available due to the issue of privacy. In other words, our setting can be extended to the source-free case where no source domain samples are offered while performing domain adaptations. 
    \item \textcolor{black}{CLAMP is not applicable yet for the control system context. Recently, three exist some interesting works in this area. Song et al. \cite{Song2023SwitchingLikeES} proposes an event-triggered state estimation for reaction–diffusion neural networks (RDNNs) subject to Denial-of-Service (DoS) attacks. Zhuang et al. \cite{Zhuang2023AnOI} puts forward an optimal iterative learning control (ILC) algorithm for linear time-invariant multiple-input–multiple output (MIMO) systems with nonuniform trial lengths under input constraints. Wang et al. \cite{Wang2023QlearningBF} proposes a Q-learning based fault estimation (FE) and fault tolerant control (FTC) scheme under ILC framework.} 
\end{itemize}

\section{Conclusion}
\textcolor{black}{This paper presents the cross-domain continual learning problem calling for a model to overcome the issues of domain shift and catastrophic forgetting simultaneously. To this end, continual learning approach for many processes (CLAMP) is proposed where it is built upon the concept of dual assessors meta-weighting the multi-objective function such that the trade-off between the stability and the plasticity is achieved and the concept of class-aware adversarial domain adaptation dampening the gaps between the source domain and the target domain. Our experimental results confirm the advantage of CLAMP beating prior arts with significant margins. Moreover, our ablation study demonstrates the positive contributions of each learning module while our analysis of generalization show no over-fitting signals of CLAMP and CLAMP utilizes low memory budgets. Our future study is devoted to reduce the complexity of CLAMP where three networks have to be trained at once.} 

\appendix

\section{Pseudo Code}
We provide the details of CLAMP in Algorithm \ref{alg:clamp}.

\begin{algorithm}
\small
    \SetAlgoLined
	\caption{CLAMP}\label{alg:clamp}
	\KwIn{current task $k$, labeled source data $\mathcal{T}_k^{S}=\{x_i^{S},y_i^{S}\}_{i=1}^{N_k^S}$, unlabelled target data $\mathcal{T}_k^{T}=\{x_i^{T}\}_{i=1}^{N_k^T}$, iteration numbers $n_{inner}, n_{outer}$, training epoch $E$, pseudo label selection threshold $\gamma$ for target data, learning rate $\eta$ and $\mu$}
	\KwOut{feature extractor $f_{\theta}$, the classifier $g_{\phi}$ and the domain classifier $\xi_{\psi}$, source assessor network $\kappa_{\zeta_S}^S$, target assessor network $\kappa_{\zeta_T}^T$}
    \textbf{Init}: $M_S\leftarrow\emptyset$, $M_T\leftarrow\emptyset$ when k=1 \tcp*{\scriptsize Initialize episodic memories}
    $\mathcal{T}_{train_S}^k\leftarrow \mathcal{T}_k^S \cup M_S$ \;
    \small
	\For{$e = 1, ..., E$}{
		\small{\textbf{Stage 1} Unsupervised Process Adaptation} \tcp*{\scriptsize $1 \leq e \leq E_{pa}$}
        Update $f_{\theta}$, $\xi_{\psi}$ by $\mathop{\mathcal{L}_{pa}}_{(x,y)\in \small{T}_{train_S}^k\&(\mathcal{T}_k^{T} \cup M_T)}$ \;
		\small{\textbf{Stage 2} Source Domain} \tcp*{\scriptsize $e \textgreater E_{pa}$}
        \For{$iter_o = 1, ..., n_{outer}$ }{
            Form validation set $\mathcal{T}_{val_S}^k$ by applying three random transformations to $\mathcal{T}_{train_S}^k$ \;
            \For{$iter_i = 1, ..., n_{inner}$}{
            Update $\kappa_{\zeta_S}^S$ based on $\zeta_{S} \leftarrow \zeta_{S}-\mu\nabla_{\zeta_S}{\mathcal{L}_S}_{(x,y)\in \mathcal{T}_{val_S}^k}$ \tcp*{\scriptsize $\mu$: inner iteration learning rate}
            }
            Update $f_{\theta}$, $g_{\phi}$ by ${\mathcal{L}_{CE}}_{(x,y)\in \mathcal{T}_{train_S}^k}$, ${\mathcal{L}_{DER}}_{(x,y)\in M_S}$, ${\mathcal{L}_{distill}}_{(x,y)\in M_S}$
        }
		\small{\textbf{Stage 3} Target Domain} \tcp*{\scriptsize $e \textgreater E_{pa}$}
        Select pseudo-labeled target samples set $\mathcal{T}_k^{ps}=\{x^{T}, \widetilde y\}$\; 
        $\mathcal{T}_{train_{ps}}^k\leftarrow \mathcal{T}_k^{ps} \cup M_T$ \;
        \For{$iter_o = 1, ..., n_{outer}$ }{
            Select validation set $\mathcal{T}_{val_T}^k$ based on process similar samples from $\mathcal{T}_{train_S}^k$ \;
            \For{$iter_i = 1, ..., n_{inner}$}{
            Update $\kappa_{\zeta_T}^T$ based on $\zeta_{T} \leftarrow \zeta_{T}-\mu\nabla_{\zeta_T}{\mathcal{L}_T}_{(x,y)\in \mathcal{T}_{val_T}^k}$ 
            }
            Update $f_{\theta}$, $g_{\phi}$ by ${\mathcal{L}_{CE}}_{(x,y)\in \mathcal{T}_{train_{ps}}^k}$, ${\mathcal{L}_{DER}}_{(x,y)\in M_T}$, ${\mathcal{L}_{distill}}_{(x,y)\in M_T}$
        }
	}
    $M_S \leftarrow M_S \cup Sample(\mathcal{T}_k^S)$, $M_T\leftarrow M_T \cup Sample(\mathcal{T}_k^{ps})$ \tcp*{\scriptsize Update episodic memory}
\end{algorithm}

\section{Experimental Task Division}
In this section, we provide details of the division for each subset in all benchmark datasets, namely, MNIST \textcolor{black}{LeCun et al.} \cite{LeCun1998GradientbasedLA}, USPS \textcolor{black}{Hull et al.} \cite{Hull1994ADF}, Office-31 \textcolor{black}{Saenko et al.} \cite{Saenko2010AdaptingVC} and Office-Home \textcolor{black}{Venkateswara et al.} \cite{Venkateswara2017DeepHN}. The division details are presented in Table \ref{tab: task split}. For the digit recognition case, which includes MNIST and USPS, the split step is 2. For Office-31,  we sort the classes in alphabetic order and divide each domain into 5 tasks. The first four tasks include 6 classes each, and the last task includes 7 classes. Office-Home consists of 65 classes, and we divide them into 13 subsets of 5 classes per task. For VisDA, it is divided into 4 tasks of 12 classes in total.

\begin{table*}[ht]\centering
\caption{Class names in each task on all benchmarks.}\label{tab: task split}
\scriptsize
\resizebox{\columnwidth}{!}{
\begin{tabular}{l|lll}\toprule
Dataset &Task &Class Index &Class Name \\
\midrule
\multirow{5}{*}{MNIST \& USPS} &1 &[0,1] &zero, one \\
&2 &[2,3] &two, three \\
&3 &[4,5] &four, five \\
&4 &[6,7] &six, seven \\
&5 &[8,9] &eight, nine \\\midrule
\multirow{5}{*}{Office-31} &1 &[0,1,2,3,4,5] &back pack, bike, bike helmet, bookcase, bottle, calculator \\
&2 &[6,7,8,9,10,11] &desk chair, desk lamp, desktop computer, file cabinet, headphones, keyboard \\
&3 &[12,13,14,15,16,17] &laptop computer, letter tray, mobile phone, monitor, mouse, mug \\
&4 &[18,19,20,21,22,23] &paper notebook, pen, phone, printer, projector, punchers \\
&5 &[24,25,26,27,28,29,30] &ring binder, ruler, scissors, speaker, stapler, tape dispenser, trash can \\\midrule
\multirow{13}{*}{Office-Home} &1 &[0,1,2,3,4] &Alarm Clock, Backpack, Batteries, Bed, Bike \\
&2 &[5,6,7,8,9] &Bottle, Bucket, Calculator, Calendar, Candles \\
&3 &[10,11,12,13,14] &Chair, Clipboards, Computer, Couch, Curtains \\
&4 &[15,16,17,18,19] &Desk Lamp, Drill, Eraser, Exit Sign, Fan \\
&5 &[20,21,22,23,24] &File Cabinet, Flipflops, Flowers, Folder, Fork \\
&6 &[25,26,27,28,29] &Glasses, Hammer, Helmet, Kettle, Keyboard \\
&7 &[30,31,32,33,34] &Knives, Lamp Shade, Laptop, Marker, Monitor \\
&8 &[35,36,37,38,39] &Mop, Mouse, Mug, Notebook, Oven \\
&9 &[40,41,42,43,44] &Pan, Paper Clip, Pen, Pencil, Postit Notes \\
&10 &[45,46,47,48,49] &Printer, Push Pin, Radio, Refrigerator, Ruler \\
&11 &[50,51,52,53,54] &Scissors, Screwdriver, Shelf, Sink, Sneakers \\
&12 &[55,56,57,58,59] &Soda, Speaker, Spoon, TV, Table \\
&13 &[60,61,62,63,64] &Telephone, ToothBrush, Toys, Trash Can, Webcam \\
\midrule
\multirow{4}{*}{VisDA} &1 &[0,1,2] &aeroplane, bicycle, bus \\
&2 &[3,4,5] &car, horse, knife \\
&3 &[6,7,8] &motorcycle, person, plant \\
&4 &[9,10,11] &skaeboard, train, truck \\
\bottomrule
\end{tabular}}
\end{table*}

\section{Model Architecture}
In CLAMP, we use LeNet with one additional linear layer of 100 units for MNIST$\leftrightarrow$USPS, ResNet-34 with one additional linear layer of 512 units for Office-31 and VisDA, and ResNet-50 with one additional linear layer of 1000 units for Office-Home as the backbone network of the main network. For the assessor network, we use a multi-layer perceptron with 2 hidden-layers of 256 nodes, followed by a two-layer LSTM with 64 units, and then connect to 2 linear-layers with 64 and 3 nodes each for MNIST$\leftrightarrow$USPS. As for Office-31, Office-Home and VisDA, we adopt ResNet-18 as part of the assessor network, followed by a two-layer LSTM with 128 units, and then 2 linear-layers with 128 and 3 nodes each. All ResNet models used in this work are pre-trained on ImageNet \textcolor{black}{Deng et al.} \cite{5206848}.

\section{Hyper-parameter Selection}\label{hyper-params}
\textcolor{black}{This section outlines the hyper-parameters choices for the various experiments conducted in this work. To enhance clarity, we use the same hyper-parameters descriptions and notation as in the corresponding original papers, and similarly to our proposed method. Below, we provide the hyper-parameters values or grid, and mark the best results we adopt in bold. Table \ref{tab: shared param} presents the shared hyper-parameters across different methods that are not specifically mentioned below. We use the SGD optimizer with weight decay and momentum set to 0.9 and 0.0005, respectively, for all methods discussed in this work. For CLAMP, we select pseudo target labels that meet $\hat{y}^T\geq0.85$ to train the target domain.}
\textcolor{black}{
\begin{itemize}
    \scriptsize
    \item CLAMP
        \begin{itemize}
            \item learning rate: \{ 0.01, \textbf{0.001}, 0.002, 0.0001 \} (MNIST$\leftrightarrow$ USPS), \{ 0.00001, \textbf{0.0001}, 0.0002, 0.001, 0.002, 0.01, 0.1\} (Office-31), \{ 0.01, 0.001, \textbf{0.0001}, 0.00001 \} (Office-Home), \textbf{0.001} (VisDA)
            \item epoch: \{5, 10, 15, \textbf{20}, 25, 50\} (MNIST$\leftrightarrow$ USPS), \{2, 3, 5, \textbf{10}, 11, 13, 15, 20, 25, 30, 40, 50\} (Office-31), \{ 5, \textbf{10}, 15, 20, 25 \} (Office-Home)
            \item process adaptation epoch: \{ 3, 5, \textbf{10} \} (MNIST$\leftrightarrow$ USPS), \{ 1, 3, \textbf{5}, 7, 8, 10, 15, 20, 25 \} (Office-31), \{ 1, 3, \textbf{5}, 10, 15 \} (Office-Home)
            \item memory: \{ 5, 10, 25, \textbf{50}, 100, 200, 250 \} (MNIST$\leftrightarrow$ USPS), \{ 1, 3, 5, \textbf{10}, 15 \} (Office-31), \{ 1, 3, \textbf{5}, 10, 15 \}
            \item pseudo label selection threshold $\gamma$: 0.85
        \end{itemize}
    \item EWC
        \begin{itemize}
            \item $\lambda$: 5000
        \end{itemize}
    \item LwF
        \begin{itemize}
            \item $T$: 2.0
            \item regularization $R$: 1.0
        \end{itemize}
    \item SI
        \begin{itemize}
            \item strength parameter $c$: 0.1
            \item damping parameter $\xi$: 0.1
        \end{itemize}
    \item MAS
        \begin{itemize}
            \item regularization $\lambda$: 1.0
        \end{itemize}
    \item RWalk
        \begin{itemize}
            \item regularization $\lambda$: 1.0
        \end{itemize}
    \item EEIL
        \begin{itemize}
            \item epochs for balanced fine-tuning: 30
            \item distillation parameter $T$: 2.0
        \end{itemize}
    \item HAL
        \begin{itemize}
            \item regularization $\lambda$: 0.1
            \item decay rate $\beta$: 0.5
            \item mean embedding strength $\gamma$: 0.1
        \end{itemize}
    \item AGLA
        \begin{itemize}
            \item Inner epoch : 1
            \item Outer epoch : 1
        \end{itemize}
    \item DANN
        \begin{itemize}
            \item adaptation parameter $\lambda$: 1.0
        \end{itemize}
\end{itemize}}

\begin{table}[htbp]\centering
\caption{\textcolor{black}{Shared hyper-parameters for compared methods.}}\label{tab: shared param}
\scriptsize
\resizebox{\linewidth}{!}{
\begin{tabular}{lcccc}\toprule
&MNIST $\leftrightarrow$ USPS &Office-31 &Office-Home &VisDA\\
\midrule
Learning rate & \{0.0001, \textbf{0.001}, 0.01\} &\{0.00001, \textit{0.0001(others)}, \textbf{0.001(CLAMP)}, 0.01, 0.1\} &\{0.00001, \textit{0.0001(others)}, \textbf{0.001(CLAMP)}, 0.01\} & 0.001 \\
Memory size &50 &10 &5 &10 \\
Epoch & 20 &\{10, \textbf{20}, 50\} &\{10, \textbf{20}\} &10 \\
\bottomrule
\end{tabular}
}
\end{table}

\section{Additional Experimental Results}

\begin{figure}[h!]
    \centering

    \begin{subfigure}[t]{0.45\textwidth}
        \includegraphics[width=\linewidth]{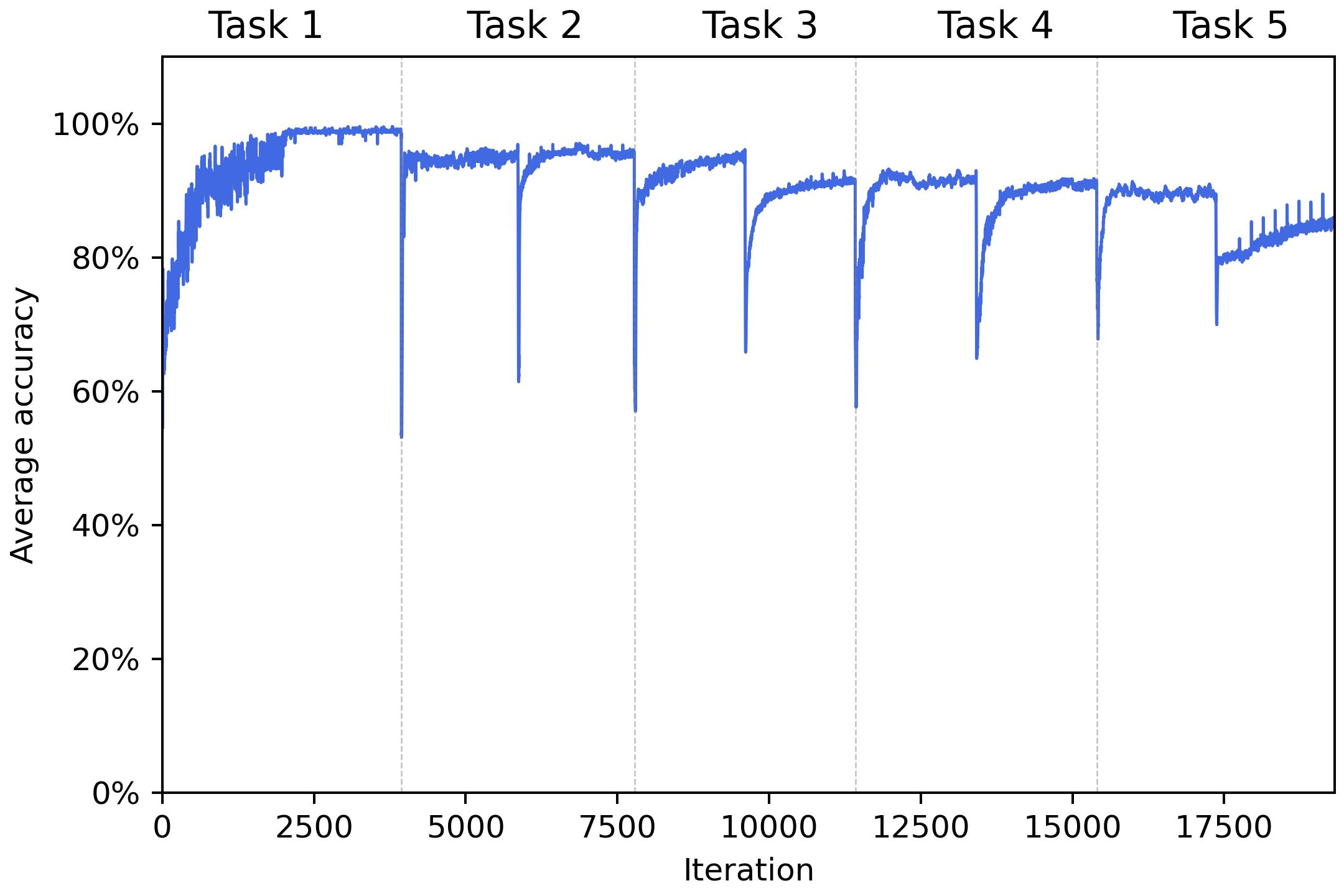}
        \caption{Average accuracy (USPS$\rightarrow$MNIST)}
        \label{fig:ave acc u2m}
    \end{subfigure}
    \hfill 
    \begin{subfigure}[t]{0.45\textwidth}
        \includegraphics[width=\linewidth]{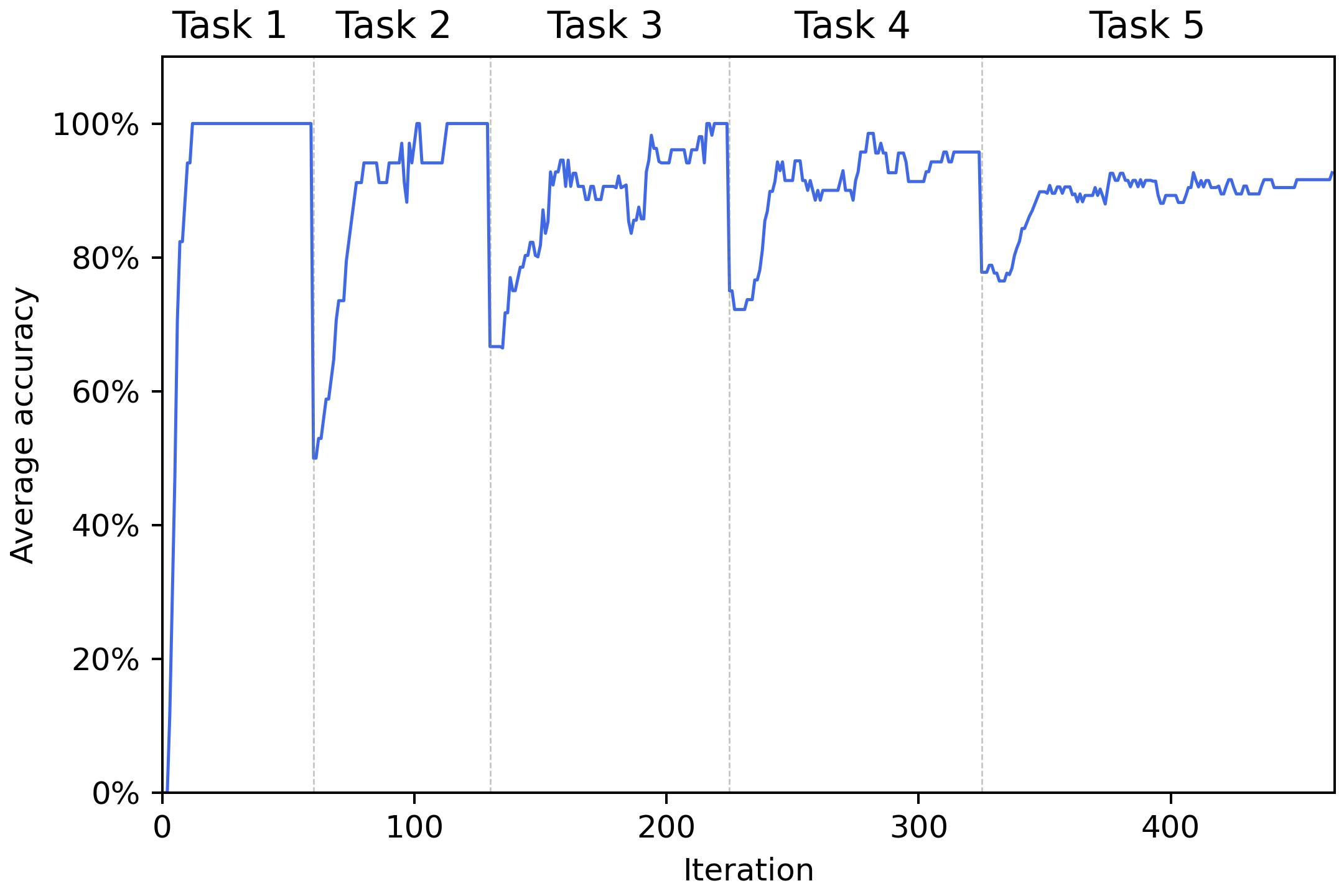}
        \caption{Average accuracy (Office-31: D$\rightarrow$W)}
        \label{fig:ave acc d2w}
    \end{subfigure}
    \newline 

    \begin{subfigure}[t]{0.45\textwidth}
        \includegraphics[width=\linewidth]{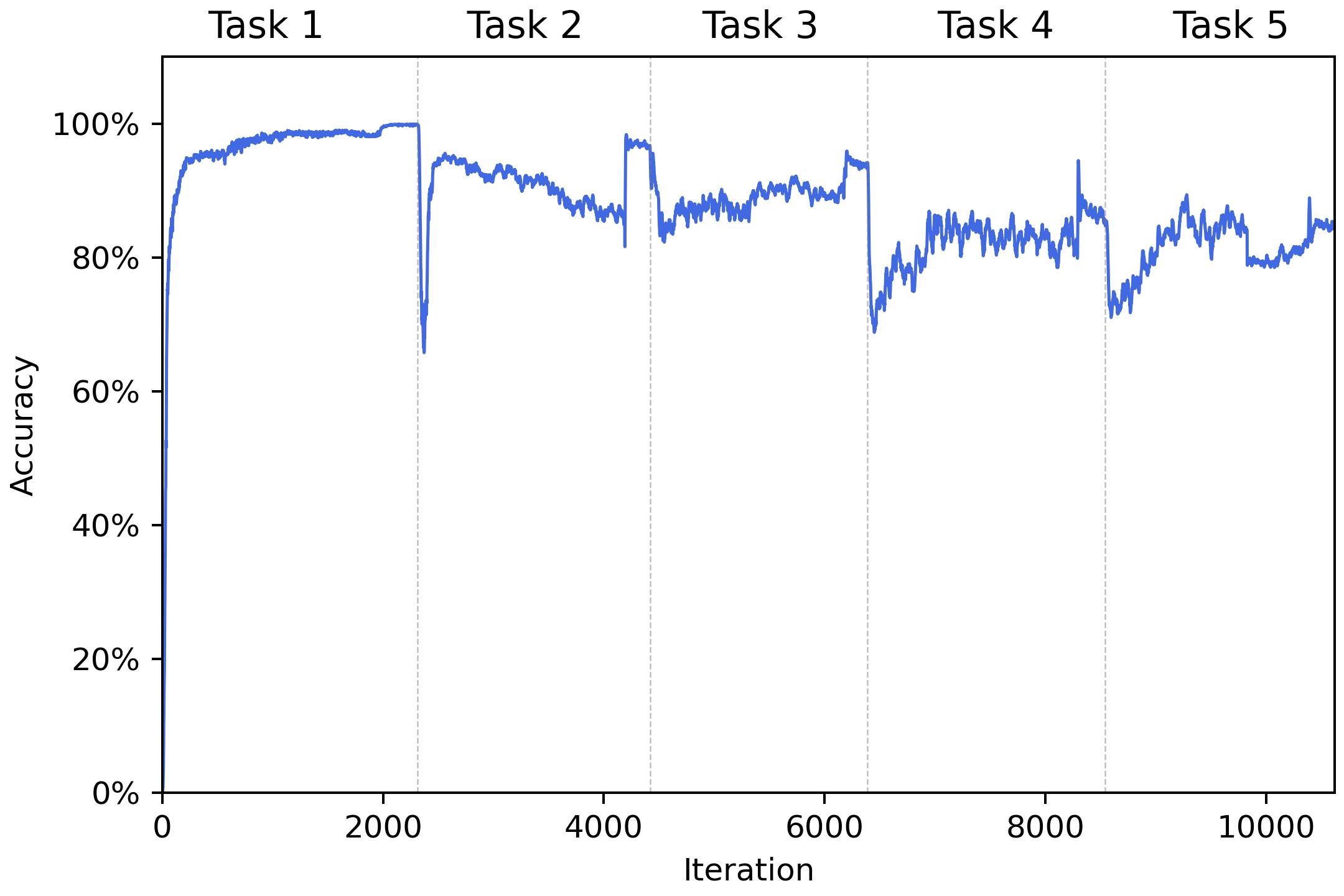}
        \caption{Accuracy on first task (USPS$\rightarrow$MNIST)}
        \label{fig:first acc u2m}
    \end{subfigure}
    \hfill 
    \begin{subfigure}[t]{0.45\textwidth}
        \includegraphics[width=\linewidth]{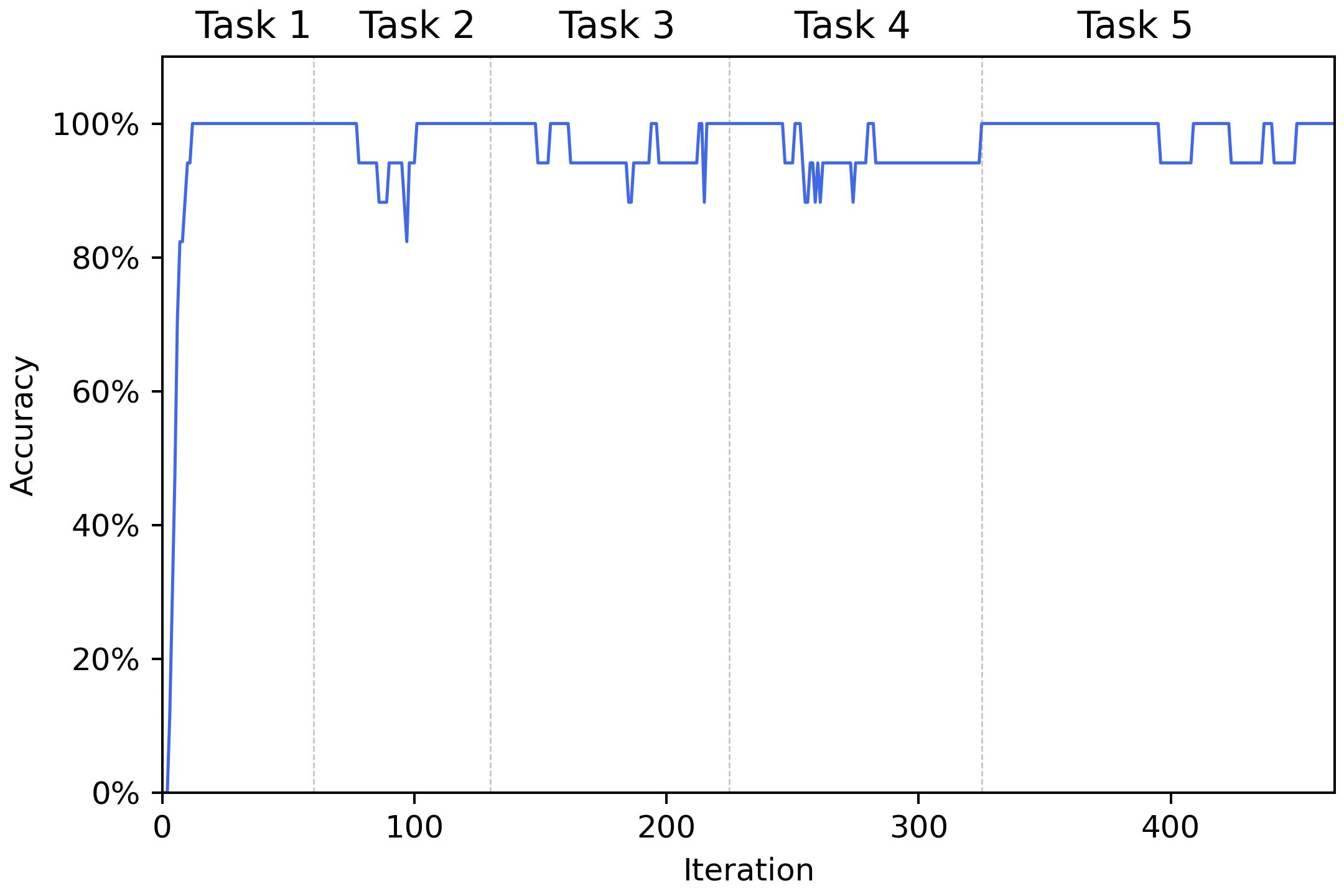}
        \caption{Accuracy on first task (Office-31: D$\rightarrow$W)}
        \label{fig:first acc d2w}
    \end{subfigure}

    \caption{(a) Average query accuracy for all learned tasks throughout the training phase in USPS$\rightarrow$MNIST; (b) Average query accuracy for all learned tasks throughout the training in Office-31: D$\rightarrow$W; (c) Query accuracy on the first task during the training phase in USPS$\rightarrow$MNIST; (d) Query accuracy on the first task during the training phase on Office-31: D$\rightarrow$W.}
    \label{fig: acc plot}
\end{figure}

In this section, we evaluate the ability of our method in continual learning manner. We report the query accuracy of all learned tasks as the training progresses across tasks from Task 1 to Task 5 on two cases, USPS$\rightarrow$MNIST (Fig. \ref{fig:ave acc u2m}) and Office-31 D$\rightarrow$W (Fig. \ref{fig:ave acc d2w}). Fig. \ref{fig:first acc u2m} and Fig. \ref{fig:first acc d2w} show the query accuracy on the first task along with the training progresses across tasks from Task 1 to Task 5 on USPS$\rightarrow$MNIST and Office-31 D$\rightarrow$W respectively. 

\begin{table}[h]\centering
\small
\caption{Accuracy (\%) of CLAMP with different exemplar sizes per class on MNIST$\leftrightarrow$USPS}\label{tab: buffer-mnistusps}
\begin{tabular}{ccc}\toprule
Exemplar per class &MNIST $\rightarrow$ USPS &USPS $\rightarrow$ MNIST \\\midrule
5 &71.93 $\pm$ 4.2 &84.40 $\pm$ 0.9 \\
25 &81.69 $\pm$ 1.1 &89.35 $\pm$ 1.1 \\
\textbf{50 (ours)} &\textbf{84.98 $\pm$ 1.3} &\textbf{89.63 $\pm$ 1.0} \\
100 &84.82 $\pm$ 1.5 &89.09 $\pm$ 0.6 \\
250 &85.02 $\pm$ 4.0 &88.33 $\pm$ 3.0 \\
\bottomrule
\end{tabular}
\end{table}

\section{Effect of the Episodic Memory Size}
We explore the effect of episodic memory size on the performance of CLAMP. Table \ref{tab: buffer-mnistusps} shows additional results on the digit recognition experiment. We can see that the accuracy increases along with the increase in the episodic memory size until it reaches 50 exemplars per class. After that, the performance drops when expanding the size from 50 to 100. For digit recognition (MNIST$\rightarrow$USPS), the accuracy approaches the highest of 85.02\% when the exemplar size is 250. 
Considering computational efficiency and optimal performance, we opt for 50 exemplars per class as the episodic memory size for the digit recognition case. This analysis underscores the delicate balance between memory size and performance, highlighting the need for episodic memory management in continual learning across many processes problem.



\bibliographystyle{model1-num-names}

\bibliography{cas-refs}

\begin{thebibliography}{10}
\expandafter\ifx\csname url\endcsname\relax
  \def\url#1{\texttt{#1}}\fi
\expandafter\ifx\csname urlprefix\endcsname\relax\def\urlprefix{URL }\fi
\expandafter\ifx\csname href\endcsname\relax
  \def\href#1#2{#2} \def\path#1{#1}\fi

\bibitem{chen2018lifelong}
Z.~Chen, B.~Liu, Lifelong machine learning, Vol.~1, Springer, 2018.

\bibitem{Parisi2018ContinualLL}
G.~I. Parisi, R.~Kemker, J.~L. Part, C.~Kanan, S.~Wermter, \href{https://api.semanticscholar.org/CorpusID:73497737}{Continual lifelong learning with neural networks: A review}, Neural networks : the official journal of the International Neural Network Society 113 (2018) 54--71.
\newline\urlprefix\url{https://api.semanticscholar.org/CorpusID:73497737}

\bibitem{Kirkpatrick2016OvercomingCF}
J.~Kirkpatrick, R.~Pascanu, N.~C. Rabinowitz, J.~Veness, G.~Desjardins, A.~A. Rusu, K.~Milan, J.~Quan, T.~Ramalho, A.~Grabska-Barwinska, D.~Hassabis, C.~Clopath, D.~Kumaran, R.~Hadsell, \href{https://api.semanticscholar.org/CorpusID:4704285}{Overcoming catastrophic forgetting in neural networks}, Proceedings of the National Academy of Sciences 114 (2016) 3521 -- 3526.
\newline\urlprefix\url{https://api.semanticscholar.org/CorpusID:4704285}

\bibitem{Mao2021ContinualLV}
F.~Mao, W.~Weng, M.~Pratama, E.~K.~Y. Yapp, \href{https://api.semanticscholar.org/CorpusID:233710658}{Continual learning via inter-task synaptic mapping}, ArXiv abs/2106.13954 (2021).
\newblock \href {https://doi.org/https://doi.org/10.1016/j.knosys.2021.106947} {\path{doi:https://doi.org/10.1016/j.knosys.2021.106947}}.
\newline\urlprefix\url{https://api.semanticscholar.org/CorpusID:233710658}

\bibitem{Cha2020CPRCR}
S.~Cha, H.~Hsu, F.~du~Pin~Calmon, T.~Moon, \href{https://api.semanticscholar.org/CorpusID:219636462}{Cpr: Classifier-projection regularization for continual learning}, ArXiv abs/2006.07326 (2020).
\newblock \href {https://doi.org/https://doi.org/10.48550/arXiv.2006.07326} {\path{doi:https://doi.org/10.48550/arXiv.2006.07326}}.
\newline\urlprefix\url{https://api.semanticscholar.org/CorpusID:219636462}

\bibitem{Rusu2016ProgressiveNN}
A.~A. Rusu, N.~C. Rabinowitz, G.~Desjardins, H.~Soyer, J.~Kirkpatrick, K.~Kavukcuoglu, R.~Pascanu, R.~Hadsell, \href{https://api.semanticscholar.org/CorpusID:15350923}{Progressive neural networks}, ArXiv abs/1606.04671 (2016).
\newblock \href {https://doi.org/https://doi.org/10.48550/arXiv.1606.04671} {\path{doi:https://doi.org/10.48550/arXiv.1606.04671}}.
\newline\urlprefix\url{https://api.semanticscholar.org/CorpusID:15350923}

\bibitem{Pratama2021UnsupervisedCL}
M.~Pratama, A.~Ashfahani, E.~D. Lughofer, \href{https://api.semanticscholar.org/CorpusID:235658071}{Unsupervised continual learning via self-adaptive deep clustering approach}, in: CSSL, 2021.
\newblock \href {https://doi.org/https://doi.org/10.48550/arXiv.2106.14563} {\path{doi:https://doi.org/10.48550/arXiv.2106.14563}}.
\newline\urlprefix\url{https://api.semanticscholar.org/CorpusID:235658071}

\bibitem{Yoon2017LifelongLW}
J.~Yoon, E.~Yang, J.~Lee, S.~J. Hwang, \href{https://api.semanticscholar.org/CorpusID:3693512}{Lifelong learning with dynamically expandable networks}, ArXiv abs/1708.01547 (2017).
\newblock \href {https://doi.org/https://doi.org/10.48550/arXiv.1708.01547} {\path{doi:https://doi.org/10.48550/arXiv.1708.01547}}.
\newline\urlprefix\url{https://api.semanticscholar.org/CorpusID:3693512}

\bibitem{lopez2017gradient}
D.~Lopez-Paz, M.~Ranzato, \href{https://api.semanticscholar.org/CorpusID:37308416}{Gradient episodic memory for continual learning}, Advances in neural information processing systems 30 (2017).
\newline\urlprefix\url{https://api.semanticscholar.org/CorpusID:37308416}

\bibitem{Shin2017ContinualLW}
H.~Shin, J.~K. Lee, J.~Kim, J.~Kim, \href{https://api.semanticscholar.org/CorpusID:1888776}{Continual learning with deep generative replay}, in: Neural Information Processing Systems, 2017.
\newblock \href {https://doi.org/https://doi.org/10.48550/arXiv.1705.08690} {\path{doi:https://doi.org/10.48550/arXiv.1705.08690}}.
\newline\urlprefix\url{https://api.semanticscholar.org/CorpusID:1888776}

\bibitem{Goodfellow2014GenerativeAN}
I.~J. Goodfellow, J.~Pouget-Abadie, M.~Mirza, B.~Xu, D.~Warde-Farley, S.~Ozair, A.~C. Courville, Y.~Bengio, \href{https://api.semanticscholar.org/CorpusID:261560300}{Generative adversarial nets}, in: Neural Information Processing Systems, 2014.
\newblock \href {https://doi.org/10.5555/2969033.2969125} {\path{doi:10.5555/2969033.2969125}}.
\newline\urlprefix\url{https://api.semanticscholar.org/CorpusID:261560300}

\bibitem{Kingma2013AutoEncodingVB}
D.~P. Kingma, M.~Welling, \href{https://api.semanticscholar.org/CorpusID:216078090}{Auto-encoding variational bayes}, CoRR abs/1312.6114 (2013).
\newline\urlprefix\url{https://api.semanticscholar.org/CorpusID:216078090}

\bibitem{Chandra2016AnAF}
S.~Chandra, A.~Haque, L.~Khan, C.~C. Aggarwal, \href{https://api.semanticscholar.org/CorpusID:16830931}{An adaptive framework for multistream classification}, Proceedings of the 25th ACM International on Conference on Information and Knowledge Management (2016).
\newblock \href {https://doi.org/https://doi.org/10.1145/2983323.2983842} {\path{doi:https://doi.org/10.1145/2983323.2983842}}.
\newline\urlprefix\url{https://api.semanticscholar.org/CorpusID:16830931}

\bibitem{Lao2021ATC}
Q.~Lao, X.~Jiang, M.~Havaei, Y.~Bengio, \href{https://api.semanticscholar.org/CorpusID:232113812}{A two-stream continual learning system with variational domain-agnostic feature replay}, IEEE Transactions on Neural Networks and Learning Systems 33 (2021) 4466--4478.
\newline\urlprefix\url{https://api.semanticscholar.org/CorpusID:232113812}

\bibitem{Lin2022PrototypeGuidedCA}
H.~Lin, Y.~Zhang, Z.~Qiu, S.~Niu, C.~Gan, Y.~Liu, M.~Tan, \href{https://api.semanticscholar.org/CorpusID:251018211}{Prototype-guided continual adaptation for class-incremental unsupervised domain adaptation}, ArXiv (2022).
\newblock \href {https://doi.org/https://doi.org/10.48550/arXiv.2207.10856} {\path{doi:https://doi.org/10.48550/arXiv.2207.10856}}.
\newline\urlprefix\url{https://api.semanticscholar.org/CorpusID:251018211}

\bibitem{Ganin2015DomainAdversarialTO}
Y.~Ganin, E.~Ustinova, H.~Ajakan, P.~Germain, H.~Larochelle, F.~Laviolette, M.~Marchand, V.~S. Lempitsky, \href{https://api.semanticscholar.org/CorpusID:2871880}{Domain-adversarial training of neural networks}, in: Journal of machine learning research, 2015.
\newblock \href {https://doi.org/https://doi.org/10.48550/arXiv.1505.07818} {\path{doi:https://doi.org/10.48550/arXiv.1505.07818}}.
\newline\urlprefix\url{https://api.semanticscholar.org/CorpusID:2871880}

\bibitem{Li2023MetaReweightedRF}
S.~Li, W.~Ma, J.~Zhang, C.~H. Liu, J.~Liang, G.~Wang, \href{https://api.semanticscholar.org/CorpusID:240536699}{Meta-reweighted regularization for unsupervised domain adaptation}, IEEE Transactions on Knowledge and Data Engineering 35 (2023) 2781--2795.
\newline\urlprefix\url{https://api.semanticscholar.org/CorpusID:240536699}

\bibitem{Zheng2020DeepML}
W.~Zheng, J.~Lu, J.~Zhou, \href{https://api.semanticscholar.org/CorpusID:219631345}{Deep metric learning via adaptive learnable assessment}, 2020 IEEE/CVF Conference on Computer Vision and Pattern Recognition (CVPR) (2020) 2957--2966.
\newline\urlprefix\url{https://api.semanticscholar.org/CorpusID:219631345}

\bibitem{VinciusdeCarvalho2024TowardsCC}
M.~V. de~Carvalho, M.~Pratama, J.~Zhang, C.~Haoyan, E.~K.~Y. Yapp, \href{https://api.semanticscholar.org/CorpusID:267759608}{Towards cross-domain continual learning}, 2024.
\newblock \href {https://doi.org/https://doi.org/10.48550/arXiv.2402.12490} {\path{doi:https://doi.org/10.48550/arXiv.2402.12490}}.
\newline\urlprefix\url{https://api.semanticscholar.org/CorpusID:267759608}

\bibitem{Masum2023AssessorGuidedLF}
M.~A. Ma'sum, M.~Pratama, E.~D. Lughofer, W.~Ding, W.~Jatmiko, \href{https://api.semanticscholar.org/CorpusID:257636622}{Assessor-guided learning for continual environments}, Inf. Sci. 640 (2023) 119088.
\newblock \href {https://doi.org/https://doi.org/10.48550/arXiv.2303.11624} {\path{doi:https://doi.org/10.48550/arXiv.2303.11624}}.
\newline\urlprefix\url{https://api.semanticscholar.org/CorpusID:257636622}

\bibitem{aljundi2018memory}
R.~Aljundi, F.~Babiloni, M.~Elhoseiny, M.~Rohrbach, T.~Tuytelaars, \href{https://api.semanticscholar.org/CorpusID:4254748}{Memory aware synapses: Learning what (not) to forget} (2018) 139--154\href {https://doi.org/https://doi.org/10.48550/arXiv.1711.09601} {\path{doi:https://doi.org/10.48550/arXiv.1711.09601}}.
\newline\urlprefix\url{https://api.semanticscholar.org/CorpusID:4254748}

\bibitem{Li2016LearningWF}
Z.~Li, D.~Hoiem, \href{https://api.semanticscholar.org/CorpusID:4853851}{Learning without forgetting}, IEEE Transactions on Pattern Analysis and Machine Intelligence 40 (2016) 2935--2947.
\newline\urlprefix\url{https://api.semanticscholar.org/CorpusID:4853851}

\bibitem{Schwarz2018ProgressC}
J.~Schwarz, W.~M. Czarnecki, J.~Luketina, A.~Grabska-Barwinska, Y.~W. Teh, R.~Pascanu, R.~Hadsell, \href{https://api.semanticscholar.org/CorpusID:21718339}{Progress \& compress: A scalable framework for continual learning}, ArXiv abs/1805.06370 (2018).
\newblock \href {https://doi.org/https://doi.org/10.48550/arXiv.1711.09601} {\path{doi:https://doi.org/10.48550/arXiv.1711.09601}}.
\newline\urlprefix\url{https://api.semanticscholar.org/CorpusID:21718339}

\bibitem{Paik2019OvercomingCF}
I.~Paik, S.~Oh, T.~Kwak, I.~Kim, \href{https://api.semanticscholar.org/CorpusID:199001153}{Overcoming catastrophic forgetting by neuron-level plasticity control}, in: AAAI Conference on Artificial Intelligence, 2019.
\newblock \href {https://doi.org/https://doi.org/10.48550/arXiv.1907.13322} {\path{doi:https://doi.org/10.48550/arXiv.1907.13322}}.
\newline\urlprefix\url{https://api.semanticscholar.org/CorpusID:199001153}

\bibitem{Li2019LearnTG}
X.~Li, Y.~Zhou, T.~Wu, R.~Socher, C.~Xiong, \href{https://api.semanticscholar.org/CorpusID:90259576}{Learn to grow: A continual structure learning framework for overcoming catastrophic forgetting}, ArXiv abs/1904.00310 (2019).
\newblock \href {https://doi.org/https://doi.org/10.48550/arXiv.1904.00310} {\path{doi:https://doi.org/10.48550/arXiv.1904.00310}}.
\newline\urlprefix\url{https://api.semanticscholar.org/CorpusID:90259576}

\bibitem{Xu2021AdaptivePC}
J.~Xu, J.~Ma, X.~Gao, Z.~Zhu, \href{https://api.semanticscholar.org/CorpusID:235767757}{Adaptive progressive continual learning}, IEEE Transactions on Pattern Analysis and Machine Intelligence 44 (2021) 6715--6728.
\newline\urlprefix\url{https://api.semanticscholar.org/CorpusID:235767757}

\bibitem{Ashfahani2021UnsupervisedCL}
A.~Ashfahani, M.~Pratama, \href{https://api.semanticscholar.org/CorpusID:237572299}{Unsupervised continual learning in streaming environments}, IEEE Transactions on Neural Networks and Learning Systems 34 (2021) 9992--10003.
\newline\urlprefix\url{https://api.semanticscholar.org/CorpusID:237572299}

\bibitem{Rakaraddi2022ReinforcedCL}
A.~Rakaraddi, S.-K. Lam, M.~Pratama, M.~V. de~Carvalho, \href{https://api.semanticscholar.org/CorpusID:252089650}{Reinforced continual learning for graphs}, Proceedings of the 31st ACM International Conference on Information \& Knowledge Management (2022).
\newblock \href {https://doi.org/https://doi.org/10.48550/arXiv.2209.01556} {\path{doi:https://doi.org/10.48550/arXiv.2209.01556}}.
\newline\urlprefix\url{https://api.semanticscholar.org/CorpusID:252089650}

\bibitem{Rebuffi2016iCaRLIC}
S.-A. Rebuffi, A.~Kolesnikov, G.~Sperl, C.~H. Lampert, \href{https://api.semanticscholar.org/CorpusID:206596260}{icarl: Incremental classifier and representation learning}, 2017 IEEE Conference on Computer Vision and Pattern Recognition (CVPR) (2016) 5533--5542\href {https://doi.org/https://doi.org/10.48550/arXiv.1611.07725} {\path{doi:https://doi.org/10.48550/arXiv.1611.07725}}.
\newline\urlprefix\url{https://api.semanticscholar.org/CorpusID:206596260}

\bibitem{Chaudhry2019OnTE}
A.~Chaudhry, M.~Rohrbach, M.~Elhoseiny, T.~Ajanthan, P.~K. Dokania, P.~H.~S. Torr, M.~Ranzato, \href{https://api.semanticscholar.org/CorpusID:173188188}{On tiny episodic memories in continual learning}, arXiv: Learning (2019).
\newblock \href {https://doi.org/https://doi.org/10.48550/arXiv.1902.10486} {\path{doi:https://doi.org/10.48550/arXiv.1902.10486}}.
\newline\urlprefix\url{https://api.semanticscholar.org/CorpusID:173188188}

\bibitem{Chaudhry2018EfficientLL}
A.~Chaudhry, M.~Ranzato, M.~Rohrbach, M.~Elhoseiny, \href{https://api.semanticscholar.org/CorpusID:54443381}{Efficient lifelong learning with a-gem}, ArXiv abs/1812.00420 (2018).
\newblock \href {https://doi.org/https://doi.org/10.48550/arXiv.1812.00420} {\path{doi:https://doi.org/10.48550/arXiv.1812.00420}}.
\newline\urlprefix\url{https://api.semanticscholar.org/CorpusID:54443381}

\bibitem{Chaudhry2019UsingHT}
A.~Chaudhry, A.~Gordo, P.~K. Dokania, P.~H.~S. Torr, D.~Lopez-Paz, \href{https://api.semanticscholar.org/CorpusID:210957697}{Using hindsight to anchor past knowledge in continual learning}, in: AAAI Conference on Artificial Intelligence, 2021.
\newblock \href {https://doi.org/https://doi.org/10.48550/arXiv.2002.08165} {\path{doi:https://doi.org/10.48550/arXiv.2002.08165}}.
\newline\urlprefix\url{https://api.semanticscholar.org/CorpusID:210957697}

\bibitem{VinciusdeCarvalho2022ClassIncrementalLV}
M.~V. de~Carvalho, M.~Pratama, J.~Zhang, Y.~San, \href{https://api.semanticscholar.org/CorpusID:252090326}{Class-incremental learning via knowledge amalgamation}, in: ECML/PKDD, 2022.
\newblock \href {https://doi.org/https://doi.org/10.48550/arXiv.2209.02112} {\path{doi:https://doi.org/10.48550/arXiv.2209.02112}}.
\newline\urlprefix\url{https://api.semanticscholar.org/CorpusID:252090326}

\bibitem{Dam2022ScalableAO}
T.~Dam, M.~Pratama, M.~M. Ferdaus, S.~G. Anavatti, H.~Abbas, \href{https://api.semanticscholar.org/CorpusID:252089827}{Scalable adversarial online continual learning}, in: ECML/PKDD, 2022.
\newblock \href {https://doi.org/https://doi.org/10.48550/arXiv.2209.01558} {\path{doi:https://doi.org/10.48550/arXiv.2209.01558}}.
\newline\urlprefix\url{https://api.semanticscholar.org/CorpusID:252089827}

\bibitem{Wang2021LearningTP}
Z.~Wang, Z.~Zhang, C.-Y. Lee, H.~Zhang, R.~Sun, X.~Ren, G.~Su, V.~Perot, J.~G. Dy, T.~Pfister, \href{https://api.semanticscholar.org/CorpusID:245218925}{Learning to prompt for continual learning}, 2022 IEEE/CVF Conference on Computer Vision and Pattern Recognition (CVPR) (2022) 139--149\href {https://doi.org/https://doi.org/10.48550/arXiv.2112.08654} {\path{doi:https://doi.org/10.48550/arXiv.2112.08654}}.
\newline\urlprefix\url{https://api.semanticscholar.org/CorpusID:245218925}

\bibitem{Wang2022DualPromptCP}
Z.~Wang, Z.~Zhang, S.~Ebrahimi, R.~Sun, H.~Zhang, C.-Y. Lee, X.~Ren, G.~Su, V.~Perot, J.~G. Dy, T.~Pfister, \href{https://api.semanticscholar.org/CorpusID:248085201}{Dualprompt: Complementary prompting for rehearsal-free continual learning}, ArXiv (2022) 631–648\href {https://doi.org/https://doi.org/10.1007/978-3-031-19809-0_36} {\path{doi:https://doi.org/10.1007/978-3-031-19809-0_36}}.
\newline\urlprefix\url{https://api.semanticscholar.org/CorpusID:248085201}

\bibitem{Finn2017ModelAgnosticMF}
C.~Finn, P.~Abbeel, S.~Levine, \href{https://api.semanticscholar.org/CorpusID:6719686}{Model-agnostic meta-learning for fast adaptation of deep networks}, in: International Conference on Machine Learning, 2017.
\newblock \href {https://doi.org/https://doi.org/10.48550/arXiv.1703.03400} {\path{doi:https://doi.org/10.48550/arXiv.1703.03400}}.
\newline\urlprefix\url{https://api.semanticscholar.org/CorpusID:6719686}

\bibitem{Schmidhuber1987EvolutionaryPI}
J.~Schmidhuber, \href{https://api.semanticscholar.org/CorpusID:264351059}{Evolutionary principles in self-referential learning, or on learning how to learn: The meta-meta-. hook}, 1987.
\newline\urlprefix\url{https://api.semanticscholar.org/CorpusID:264351059}

\bibitem{Javed2019MetaLearningRF}
K.~Javed, M.~White, \href{https://api.semanticscholar.org/CorpusID:168169908}{Meta-learning representations for continual learning}, arXiv e-prints abs/1905.12588 (2019).
\newblock \href {https://doi.org/https://doi.org/10.48550/arXiv.1905.12588} {\path{doi:https://doi.org/10.48550/arXiv.1905.12588}}.
\newline\urlprefix\url{https://api.semanticscholar.org/CorpusID:168169908}

\bibitem{Gupta2020LaMAMLLM}
G.~Gupta, K.~Yadav, L.~Paull, \href{https://api.semanticscholar.org/CorpusID:220831405}{La-maml: Look-ahead meta learning for continual learning}, Advances in Neural Information Processing Systems 33 (2020) 11588--11598.
\newline\urlprefix\url{https://api.semanticscholar.org/CorpusID:220831405}

\bibitem{Pham2021ContextualTN}
Q.~H. Pham, C.~Liu, D.~Sahoo, S.~C.~H. Hoi, \href{https://api.semanticscholar.org/CorpusID:235614405}{Contextual transformation networks for online continual learning}, in: International Conference on Learning Representations, 2021.
\newline\urlprefix\url{https://api.semanticscholar.org/CorpusID:235614405}

\bibitem{MSC}
S.~Chandra, A.~Haque, L.~Khan, C.~C. Aggarwal, \href{https://api.semanticscholar.org/CorpusID:16830931}{An adaptive framework for multistream classification}, Proceedings of the 25th ACM International on Conference on Information and Knowledge Management (2016).
\newblock \href {https://doi.org/https://doi.org/10.1145/2983323.2983842} {\path{doi:https://doi.org/10.1145/2983323.2983842}}.
\newline\urlprefix\url{https://api.semanticscholar.org/CorpusID:16830931}

\bibitem{FUSION}
A.~Haque, Z.~Wang, S.~Chandra, B.~Dong, L.~Khan, K.~W. Hamlen, \href{https://api.semanticscholar.org/CorpusID:25322234}{Fusion: An online method for multistream classification}, Proceedings of the 2017 ACM on Conference on Information and Knowledge Management (2017).
\newblock \href {https://doi.org/https://doi.org/10.1145/3132847.3132886} {\path{doi:https://doi.org/10.1145/3132847.3132886}}.
\newline\urlprefix\url{https://api.semanticscholar.org/CorpusID:25322234}

\bibitem{MSCRDR}
B.~Dong, Y.~Gao, S.~Chandra, L.~Khan, \href{https://api.semanticscholar.org/CorpusID:57888755}{Multistream classification with relative density ratio estimation}, in: AAAI Conference on Artificial Intelligence, 2019.
\newblock \href {https://doi.org/https://doi.org/10.1609/aaai.v33i01.33013478} {\path{doi:https://doi.org/10.1609/aaai.v33i01.33013478}}.
\newline\urlprefix\url{https://api.semanticscholar.org/CorpusID:57888755}

\bibitem{ATL}
M.~Pratama, M.~V. de~Carvalho, R.~Xie, E.~D. Lughofer, J.~Lu, \href{https://api.semanticscholar.org/CorpusID:203902308}{Atl: Autonomous knowledge transfer from many streaming processes}, Proceedings of the 28th ACM International Conference on Information and Knowledge Management (2019).
\newblock \href {https://doi.org/https://doi.org/10.48550/arXiv.1910.03434} {\path{doi:https://doi.org/10.48550/arXiv.1910.03434}}.
\newline\urlprefix\url{https://api.semanticscholar.org/CorpusID:203902308}

\bibitem{MELANIE}
H.~Du, L.~L. Minku, H.~Zhou, \href{https://api.semanticscholar.org/CorpusID:57721122}{Multi-source transfer learning for non-stationary environments}, 2019 International Joint Conference on Neural Networks (IJCNN) (2019) 1--8\href {https://doi.org/https://doi.org/10.1109/IJCNN.2019.8852024} {\path{doi:https://doi.org/10.1109/IJCNN.2019.8852024}}.
\newline\urlprefix\url{https://api.semanticscholar.org/CorpusID:57721122}

\bibitem{MARLINE}
H.~Du, L.~L. Minku, H.~Zhou, \href{https://api.semanticscholar.org/CorpusID:222262850}{Marline: Multi-source mapping transfer learning for non-stationary environments}, 2020 IEEE International Conference on Data Mining (ICDM) (2020) 122--131\href {https://doi.org/https://doi.org/10.1109/ICDM50108.2020.00021} {\path{doi:https://doi.org/10.1109/ICDM50108.2020.00021}}.
\newline\urlprefix\url{https://api.semanticscholar.org/CorpusID:222262850}

\bibitem{AOMSDA}
R.~Xie, M.~Pratama, \href{https://api.semanticscholar.org/CorpusID:237420507}{Automatic online multi-source domain adaptation}, ArXiv abs/2109.01996 (2021).
\newblock \href {https://doi.org/https://doi.org/10.48550/arXiv.2109.01996} {\path{doi:https://doi.org/10.48550/arXiv.2109.01996}}.
\newline\urlprefix\url{https://api.semanticscholar.org/CorpusID:237420507}

\bibitem{COMC}
H.~Tao, Z.~Wang, Y.~Li, M.~Zamani, L.~Khan, \href{https://api.semanticscholar.org/CorpusID:203605281}{Comc: A framework for online cross-domain multistream classification}, 2019 International Joint Conference on Neural Networks (IJCNN) (2019) 1--8\href {https://doi.org/https://doi.org/10.1109/IJCNN.2019.8851931} {\path{doi:https://doi.org/10.1109/IJCNN.2019.8851931}}.
\newline\urlprefix\url{https://api.semanticscholar.org/CorpusID:203605281}

\bibitem{ACDC}
M.~V. de~Carvalho, M.~Pratama, J.~Zhang, E.~K.~Y. Yapp, \href{https://api.semanticscholar.org/CorpusID:238259903}{Acdc: Online unsupervised cross-domain adaptation}, Knowl. Based Syst. 253 (2021) 109486.
\newline\urlprefix\url{https://api.semanticscholar.org/CorpusID:238259903}

\bibitem{Weng2022AutonomousCD}
W.~Weng, M.~Pratama, C.~Za'in, M.~V. de~Carvalho, A.~Rakaraddi, A.~Ashfahani, E.~K.~Y. Yapp, \href{https://api.semanticscholar.org/CorpusID:249955297}{Autonomous cross domain adaptation under extreme label scarcity}, IEEE Transactions on Neural Networks and Learning Systems 34 (2022) 6839--6850.
\newline\urlprefix\url{https://api.semanticscholar.org/CorpusID:249955297}

\bibitem{vandeVen2019ThreeSF}
G.~M. van~de Ven, A.~S. Tolias, \href{https://api.semanticscholar.org/CorpusID:119309522}{Three scenarios for continual learning}, ArXiv (2019).
\newblock \href {https://doi.org/https://doi.org/10.48550/arXiv.1904.07734} {\path{doi:https://doi.org/10.48550/arXiv.1904.07734}}.
\newline\urlprefix\url{https://api.semanticscholar.org/CorpusID:119309522}

\bibitem{Buzzega2020DarkEF}
P.~Buzzega, M.~Boschini, A.~Porrello, D.~Abati, S.~Calderara, \href{https://api.semanticscholar.org/CorpusID:215768806}{Dark experience for general continual learning: a strong, simple baseline}, ArXiv abs/2004.07211 (2020).
\newblock \href {https://doi.org/https://doi.org/10.48550/arXiv.2004.07211} {\path{doi:https://doi.org/10.48550/arXiv.2004.07211}}.
\newline\urlprefix\url{https://api.semanticscholar.org/CorpusID:215768806}

\bibitem{BenDavid2006AnalysisOR}
S.~Ben-David, J.~Blitzer, K.~Crammer, F.~C. Pereira, \href{https://api.semanticscholar.org/CorpusID:10908021}{Analysis of representations for domain adaptation}, in: Neural Information Processing Systems, 2006.
\newblock \href {https://doi.org/https://doi.org/10.7551/mitpress/7503.003.0022} {\path{doi:https://doi.org/10.7551/mitpress/7503.003.0022}}.
\newline\urlprefix\url{https://api.semanticscholar.org/CorpusID:10908021}

\bibitem{BenDavid2010ATO}
S.~Ben-David, J.~Blitzer, K.~Crammer, A.~Kulesza, F.~C. Pereira, J.~W. Vaughan, \href{https://api.semanticscholar.org/CorpusID:8577357}{A theory of learning from different domains}, Machine Learning 79 (2010) 151--175.
\newline\urlprefix\url{https://api.semanticscholar.org/CorpusID:8577357}

\bibitem{Volpi2020ContinualAO}
R.~Volpi, D.~Larlus, G.~Rogez, \href{https://api.semanticscholar.org/CorpusID:227745438}{Continual adaptation of visual representations via domain randomization and meta-learning}, 2021 IEEE/CVF Conference on Computer Vision and Pattern Recognition (CVPR) (2020) 4441--4451\href {https://doi.org/https://doi.org/10.48550/arXiv.2012.04324} {\path{doi:https://doi.org/10.48550/arXiv.2012.04324}}.
\newline\urlprefix\url{https://api.semanticscholar.org/CorpusID:227745438}

\bibitem{Saenko2010AdaptingVC}
K.~Saenko, B.~Kulis, M.~Fritz, T.~Darrell, \href{https://api.semanticscholar.org/CorpusID:7534823}{Adapting visual category models to new domains}, in: European Conference on Computer Vision, 2010.
\newblock \href {https://doi.org/https://doi.org/10.1007/978-3-642-15561-1_16} {\path{doi:https://doi.org/10.1007/978-3-642-15561-1_16}}.
\newline\urlprefix\url{https://api.semanticscholar.org/CorpusID:7534823}

\bibitem{Venkateswara2017DeepHN}
H.~Venkateswara, J.~Eus{\'e}bio, S.~Chakraborty, S.~Panchanathan, \href{https://api.semanticscholar.org/CorpusID:2928248}{Deep hashing network for unsupervised domain adaptation}, 2017 IEEE Conference on Computer Vision and Pattern Recognition (CVPR) (2017) 5385--5394\href {https://doi.org/https://doi.org/10.1109/CVPR.2017.572} {\path{doi:https://doi.org/10.1109/CVPR.2017.572}}.
\newline\urlprefix\url{https://api.semanticscholar.org/CorpusID:2928248}

\bibitem{Peng2017VisDATV}
X.~Peng, B.~Usman, N.~Kaushik, J.~Hoffman, D.~Wang, K.~Saenko, \href{https://api.semanticscholar.org/CorpusID:28698351}{Visda: The visual domain adaptation challenge}, arXiv preprint arXiv:1710.06924 abs/1710.06924 (2017).
\newblock \href {https://doi.org/https://doi.org/10.48550/arXiv.1710.06924} {\path{doi:https://doi.org/10.48550/arXiv.1710.06924}}.
\newline\urlprefix\url{https://api.semanticscholar.org/CorpusID:28698351}

\bibitem{Zenke2017ContinualLT}
F.~Zenke, B.~Poole, S.~Ganguli, \href{https://api.semanticscholar.org/CorpusID:10409742}{Continual learning through synaptic intelligence}, Proceedings of machine learning research 70 (2017) 3987--3995.
\newline\urlprefix\url{https://api.semanticscholar.org/CorpusID:10409742}

\bibitem{Chaudhry2018RiemannianWF}
A.~Chaudhry, P.~K. Dokania, T.~Ajanthan, P.~H.~S. Torr, \href{https://api.semanticscholar.org/CorpusID:4047127}{Riemannian walk for incremental learning: Understanding forgetting and intransigence}, ArXiv (2018).
\newblock \href {https://doi.org/https://doi.org/10.48550/arXiv.1801.10112} {\path{doi:https://doi.org/10.48550/arXiv.1801.10112}}.
\newline\urlprefix\url{https://api.semanticscholar.org/CorpusID:4047127}

\bibitem{Belouadah2019IL2MCI}
E.~Belouadah, A.~D. Popescu, \href{https://api.semanticscholar.org/CorpusID:204923710}{Il2m: Class incremental learning with dual memory}, 2019 IEEE/CVF International Conference on Computer Vision (ICCV) (2019) 583--592\href {https://doi.org/https://doi.org/10.1109/ICCV.2019.00067} {\path{doi:https://doi.org/10.1109/ICCV.2019.00067}}.
\newline\urlprefix\url{https://api.semanticscholar.org/CorpusID:204923710}

\bibitem{Castro2018EndtoEndIL}
F.~M. Castro, M.~J. Mar{\'i}n-Jim{\'e}nez, N.~G. Mata, C.~Schmid, A.~Karteek, \href{https://api.semanticscholar.org/CorpusID:50785377}{End-to-end incremental learning}, in: European Conference on Computer Vision, 2018.
\newblock \href {https://doi.org/https://doi.org/10.48550/arXiv.1807.09536} {\path{doi:https://doi.org/10.48550/arXiv.1807.09536}}.
\newline\urlprefix\url{https://api.semanticscholar.org/CorpusID:50785377}

\bibitem{Masana2020ClassincrementalLS}
M.~Masana, X.~Liu, B.~Twardowski, M.~Menta, A.~D. Bagdanov, J.~van~de Weijer, \href{https://api.semanticscholar.org/CorpusID:234353728}{Class-incremental learning: Survey and performance evaluation on image classification}, IEEE Transactions on Pattern Analysis and Machine Intelligence 45 (2020) 5513--5533.
\newline\urlprefix\url{https://api.semanticscholar.org/CorpusID:234353728}

\bibitem{He2015DeepRL}
K.~He, X.~Zhang, S.~Ren, J.~Sun, \href{https://api.semanticscholar.org/CorpusID:206594692}{Deep residual learning for image recognition}, 2016 IEEE Conference on Computer Vision and Pattern Recognition (CVPR) (2015) 770--778\href {https://doi.org/https://doi.org/10.1109/CVPR.2016.90} {\path{doi:https://doi.org/10.1109/CVPR.2016.90}}.
\newline\urlprefix\url{https://api.semanticscholar.org/CorpusID:206594692}

\bibitem{5206848}
J.~Deng, W.~Dong, R.~Socher, L.-J. Li, K.~Li, L.~Fei-Fei, Imagenet: A large-scale hierarchical image database (2009) 248--255\href {https://doi.org/https://doi.org/10.1109/CVPR.2009.5206848} {\path{doi:https://doi.org/10.1109/CVPR.2009.5206848}}.

\bibitem{Peng2018MomentMF}
X.~Peng, Q.~Bai, X.~Xia, Z.~Huang, K.~Saenko, B.~Wang, \href{https://api.semanticscholar.org/CorpusID:54458071}{Moment matching for multi-source domain adaptation}, 2019 IEEE/CVF International Conference on Computer Vision (ICCV) (2018) 1406--1415\href {https://doi.org/https://doi.org/10.48550/arXiv.1812.01754} {\path{doi:https://doi.org/10.48550/arXiv.1812.01754}}.
\newline\urlprefix\url{https://api.semanticscholar.org/CorpusID:54458071}

\bibitem{Geirhos2020ShortcutLI}
R.~Geirhos, J.-H. Jacobsen, C.~Michaelis, R.~S. Zemel, W.~Brendel, M.~Bethge, F.~Wichmann, \href{https://api.semanticscholar.org/CorpusID:215786368}{Shortcut learning in deep neural networks}, Nature Machine Intelligence 2 (2020) 665 -- 673.
\newline\urlprefix\url{https://api.semanticscholar.org/CorpusID:215786368}

\bibitem{Song2023SwitchingLikeES}
X.~Song, N.~Wu, S.~Song, V.~Stojanovic, \href{https://api.semanticscholar.org/CorpusID:257503064}{Switching-like event-triggered state estimation for reaction–diffusion neural networks against dos attacks}, Neural Processing Letters 55 (2023) 8997--9018.
\newline\urlprefix\url{https://api.semanticscholar.org/CorpusID:257503064}

\bibitem{Zhuang2023AnOI}
Z.~Zhuang, H.~Tao, Y.~Chen, V.~Stojanovic, W.~Paszke, \href{https://api.semanticscholar.org/CorpusID:254613055}{An optimal iterative learning control approach for linear systems with nonuniform trial lengths under input constraints}, IEEE Transactions on Systems, Man, and Cybernetics: Systems 53 (2023) 3461--3473.
\newline\urlprefix\url{https://api.semanticscholar.org/CorpusID:254613055}

\bibitem{Wang2023QlearningBF}
R.~Wang, Z.~Zhuang, H.~Tao, W.~Paszke, V.~Stojanovic, \href{https://www.sciencedirect.com/science/article/pii/S0019057823003506}{Q-learning based fault estimation and fault tolerant iterative learning control for mimo systems.}, ISA transactions 142 (2023) 123--135.
\newline\urlprefix\url{https://www.sciencedirect.com/science/article/pii/S0019057823003506}

\bibitem{LeCun1998GradientbasedLA}
Y.~LeCun, L.~Bottou, Y.~Bengio, P.~Haffner, \href{https://api.semanticscholar.org/CorpusID:14542261}{Gradient-based learning applied to document recognition}, Proc. IEEE 86 (1998) 2278--2324.
\newline\urlprefix\url{https://api.semanticscholar.org/CorpusID:14542261}

\bibitem{Hull1994ADF}
J.~J. Hull, \href{https://api.semanticscholar.org/CorpusID:8148915}{A database for handwritten text recognition research}, IEEE Trans. Pattern Anal. Mach. Intell. 16 (1994) 550--554.
\newblock \href {https://doi.org/https://doi.org/10.1109/34.291440} {\path{doi:https://doi.org/10.1109/34.291440}}.
\newline\urlprefix\url{https://api.semanticscholar.org/CorpusID:8148915}

\end{thebibliography}





\end{document}